\documentclass[journal]{IEEEtran}

\title{\looseness-1Constraint Learning in Multi-Agent Dynamic Games from Demonstrations of Local Nash Interactions
}

\ifCLASSINFOpdf
\else
\fi

\usepackage{cite}

\usepackage{amsmath,amssymb,amsfonts}
\usepackage{todonotes}

\usepackage{float}

\usepackage{booktabs}
\usepackage{adjustbox}
\usepackage{wrapfig}
\usepackage{lipsum}
\usepackage{amsthm}
\usepackage{amssymb}
\usepackage{amsmath}
\usepackage{siunitx}
\usepackage{mathtools}
\usepackage{hyperref}
\usepackage{multirow}

\let\labelindent\relax
\usepackage[inline]{enumitem}
\usepackage[ruled,vlined,linesnumbered,noend]{algorithm2e}
\usepackage[noend]{algpseudocode}
\usepackage{ifthen}
\usepackage{dsfont}
\usepackage{balance}
\usepackage{dirtytalk}
\usepackage{afterpage}
\usepackage{bm}
\usepackage{enumitem}

\usepackage{tabularx}

\allowdisplaybreaks

\newcommand{\Avoid}{\mathcal{A}}

\newcommand{\C}{\mathcal{C}}
\newcommand{\D}{\mathcal{D}}
\newcommand{\Demo}{\text{Demonstration}}

\newcommand{\F}{\mathcal{F}}
\newcommand{\G}{\mathcal{G}}

\newcommand{\MINLP}{\text{MINLP}}

\newcommand{\Q}{\mathcal{Q}}
\newcommand{\ra}{\rightarrow}
\newcommand{\relaxed}{\text{R}}
\newcommand{\R}{\mathbb{R}}

\newcommand{\Safe}{\mathcal{S}}

\newtheorem{remark}{Remark}

\newtheorem{theorem}{Theorem}

\newcommand{\urk}{{\urcorner k}}
\newcommand{\urs}{{\urcorner s}}

\newcommand{\stat}{\textrm{stat}}

\newcommand{\eq}{\textrm{eq}}
\newcommand{\ineq}{\textrm{ineq}}
\newcommand{\loc}{\textrm{loc}}
\newcommand{\KKT}{\text{KKT}}
\newcommand{\nom}{\textrm{nom}}
\newcommand{\boldlambda}{\boldsymbol{\lambda}}
\newcommand{\boldnu}{\boldsymbol{\nu}}

\newcommand{\bolda}{\boldsymbol{a}}
\newcommand{\boldA}{\textbf{A}}
\newcommand{\boldb}{\textbf{b}}
\newcommand{\boldq}{\boldsymbol{q}}
\newcommand{\boldr}{\boldsymbol{r}}
\newcommand{\boldz}{\boldsymbol{z}}
\newcommand{\boldell}{\boldsymbol{\ell}}
\newcommand{\cv}{\text{cv}}
\newcommand{\sample}{{(s)}}

\newcommand{\boldg}{\textbf{g}}
\newcommand{\boldh}{\textbf{h}}
\newcommand{\underM}{\underline{M}}
\newcommand{\overM}{\overline{M}}

\newcommand{\Eqn}{\text{Eqn. }}
\newcommand{\st}{\text{ s.t. }}
\newcommand{\paren}[1]{{({#1})}}

\newcommand{\revision}[1]{\textcolor[rgb]{0,0,0}{#1}}

\author{Zhouyu Zhang$^{\star}$ and Chih-Yuan Chiu$^{\star\dagger}$ and Glen Chou
\thanks{Manuscript received: August 26th, 2025; Revised January 30th, 2026; Accepted March 3rd, 2026.}
\thanks{This paper was recommended for publication by Editor M. Ani Hsieh upon evaluation of the Associate Editor and Reviewers' comments.}
\thanks{${}^\star$Equal contribution.}
\thanks{${}^\dagger$Corresponding author.}
\thanks{$^{1}$All authors are with the Georgia Institute of Technology; Zhouyu Zhang and Chih-Yuan Chiu are with the School of ECE; Glen Chou is with the Schools of AE and CSP (\texttt{
\{zzhang3097, cyc, chou\} at gatech dot edu}).}
\thanks{Digital Object Identifier (DOI): see top of this page.}
}

\begin{document}

\maketitle

\thispagestyle{plain}
\pagestyle{plain} 




\vspace{-6mm}
\begin{abstract}
\looseness-1
We present an inverse dynamic game-based algorithm to learn parametric constraints from a given dataset of local 
Nash equilibrium interactions between multiple agents. Specifically, we introduce mixed-integer linear programs (MILP) encoding the Karush-Kuhn-Tucker (KKT) conditions of the interacting agents, which recover constraints consistent with the local Nash stationarity of the interaction demonstrations. We establish theoretical guarantees that our method learns inner approximations of the true safe and unsafe sets.
We also use the interaction constraints recovered by our method to design motion plans that robustly satisfy the underlying constraints. 
Across simulations and hardware experiments, our methods accurately inferred constraints and designed safe interactive motion plans for various classes of constraints, both convex and non-convex, from interaction demonstrations of agents with nonlinear dynamics.

\end{abstract}
\vspace{-2pt}
\section{Introduction}
\label{sec: Introduction}
\vspace{-2pt}



\looseness-1
Learning from demonstrations (LfD) is a powerful paradigm for enabling robots to learn constraints in their workspace \cite{Chou2020LearningConstraintsFromLocallyOptimalDemonstrationsUnderCostFunctionUncertainty, Chou2020LearningParametricConstraintsinHighDimensions, 
McPherson2021MLConstraintInferenceFromStochasticDemonstrations
}. In particular, \cite{Chou2020LearningConstraintsFromLocallyOptimalDemonstrationsUnderCostFunctionUncertainty, 
Chou2020LearningParametricConstraintsinHighDimensions
} formulate constraint inference as identifying the parameters that best explain a set of approximately optimal demonstration trajectories. 
However, existing methods assume robots operate in isolation, and 
thus cannot infer coupled constraints that depend on the states or controls of multiple agents, e.g., collision avoidance, which cannot be easily encoded using cost function penalties.

To address this gap, we use tools from dynamic game theory and inverse optimal control (IOC) to learn constraints from the demonstrations of \textit{interactions} between multiple strategic agents. 
We recover unknown constraint parameters by posing an inverse optimization problem that uses Nash equilibrium constraints to encode steady-state agent interactions and show that the inferred constraints can be used for robust motion planning.
Although IOC and dynamic games have been applied for \textit{cost} inference in multi-agent settings \cite{Peters2021InferringObjectives, Li2023CostInferenceforFeedbackDynamicGames, Mehr2023MaximumEntropyMultiAgentDynamicGames, LeCleach2021LUCIDGames}, to our knowledge, our work designs the first game-theoretic algorithm for multi-agent \textit{constraint} inference that is \textit{guaranteed} to either recover or conservatively estimate the true constraint set. Our main contributions are:

\begin{figure}[ht]
    \centering
    \includegraphics[width=\linewidth]{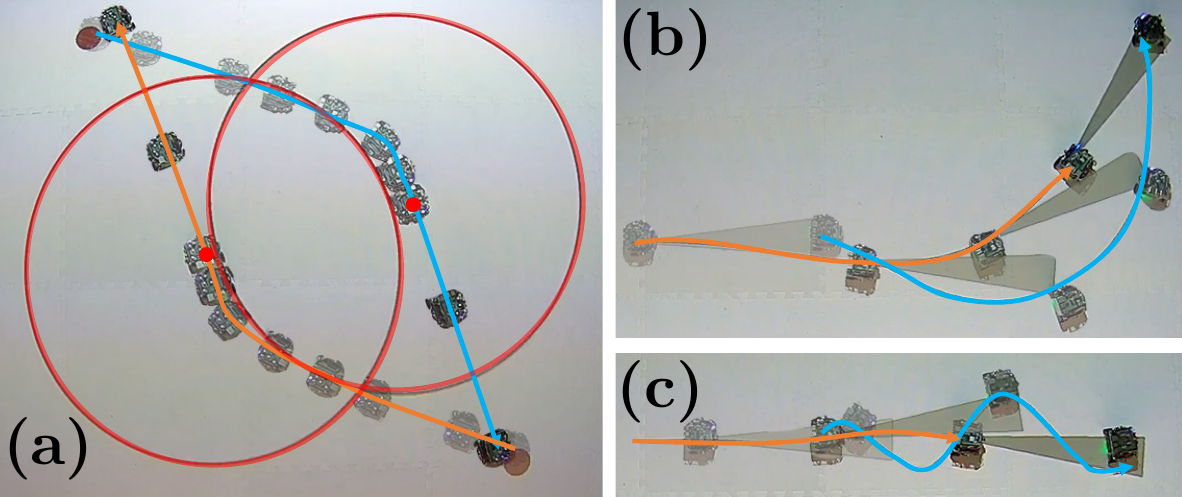}
    
    \vspace{-5pt}
    \caption{ 
    Time-lapse of safe plans computed for ground robots (Sec. \ref{subsec: Unicycle Simulations}) using (a) spherical collision-avoidance constraints (red circles) learned via our method; opacity increases over time;
    (b-c) learned line-of-sight constraints: the leading agent (blue) follows a (b) half-circle or (c) sinusoid; the pursuer (orange) maintains line-of-sight (shaded area).
\vspace{-10pt}
    }
    \label{fig:front_figure}
\end{figure}

\begin{enumerate}[leftmargin=*,align=left]
    \item We formulate a feasibility problem to learn parameterized constraints from multi-agent interaction demonstrations, generalizing the 
    method in \cite{Chou2020LearningConstraintsFromLocallyOptimalDemonstrationsUnderCostFunctionUncertainty} to the multi-agent setting. Under mild conditions, we prove that our method learns conservative estimates of the true safe and unsafe sets. 
    
    \item We use our framework to extract volumes of provably safe or 
    unsafe trajectories, or reject volumes of the constraint parameter space incompatible with the demonstrations.    
    This volume extraction approach enables safe motion planning that is robust to constraint ambiguity. 
    We also establish fundamental limitations of the learnability of interaction constraints from 
    demonstrations.
    
    
    \item We evaluate our method in simulation and hardware experiments where agents with nonlinear dynamics 
    interact under collision avoidance, line-of-sight, and proximity constraints. Our inverse learning and planning algorithms recovered unknown constraint parameters and generated safe motion plans robust to constraint uncertainty. 
    In contrast, naive applications of the baseline single-agent constraint inference method in \cite{Chou2020LearningConstraintsFromLocallyOptimalDemonstrationsUnderCostFunctionUncertainty} 
    failed to accurately recover unknown constraints with zero stationarity error (Sec. \ref{subsec: Quadcopter Simulations}).
    Moreover, cost-inference based methods which encode constraints as log barrier costs were unable to recover constraint information that could be used to generate safe motion plans downstream (Sec. \ref{subsec: Comparison Against Cost Inference Baseline}).
\end{enumerate}

\looseness-1Due to space, the appendix (with proofs, discussion, further experiments) is in an extended version of this paper
\cite{ArXivPaper}.

\vspace{-3pt}
\section{Related Work}
\label{sec: Related Work}
\vspace{-1pt}

\subsection{Constraint Learning via Inverse Optimal Control (IOC)}
\label{subsec: Constraint Learning via Inverse Optimal Control}

\looseness-1LfD via IOC has been applied to enable robots to learn new tasks \cite{Englert2017InverseKKT, Menner2021ConstrainedIOC}, infer the intent of other strategic agents \cite{Peters2021InferringObjectives, Li2023CostInferenceforFeedbackDynamicGames, Mehr2023MaximumEntropyMultiAgentDynamicGames}, and recover static environment constraints \cite{Chou2020LearningConstraintsFromLocallyOptimalDemonstrationsUnderCostFunctionUncertainty,
Chou2022GPConstraintLearning
}. In particular, \cite{Menner2021ConstrainedIOC} leverages IOC to infer convex constraints from a finite set of possible costs and constraints, 
while \cite{Mehr2016InferringAssistingConstraints, Li2016LearningObjectOrientationConstraints, Calinon2008ProbabilisticProgrammingbyDemonstrationFramework} use a single demonstration to infer local trajectory-level constraints. Our method is closest to \cite{Chou2020LearningConstraintsFromLocallyOptimalDemonstrationsUnderCostFunctionUncertainty}, which recovers unknown constraint parameters by enforcing the KKT conditions of a set of locally optimal demonstrations. However, these methods focus on single-agent scenarios.
In contrast, we use demonstrations of strategic multi-agent interactions to infer constraints coupled across agent states and controls.

\vspace{-2pt}
\subsection{Forward and Inverse Dynamic Games}
\label{subsec: Dynamic Games for Motion Planning, Cost Inference, and Constraint Inference}
\vspace{-1pt}

\looseness-1Motion planning and intent inference in multi-agent scenarios are naturally posed as dynamic games \cite{basar1998DynamicNoncooperativeGameTheory}, which provide a powerful theoretical framework for reasoning about strategic multi-robot interactions. Computationally tractable algorithms for approximating general-sum \textit{forward dynamic games} have been developed \cite{DiLamperski2019NewtonandDDPforDynamicGames, fridovich2020efficient, lecleach2020algames, Mehr2023MaximumEntropyMultiAgentDynamicGames, Kavuncu2021PotentialiLQR}, enabling efficient motion planning in interactive scenarios. Meanwhile, \cite{AwasthiLamperski2020InverseDifferentialGames, Peters2021InferringObjectives, Li2023CostInferenceforFeedbackDynamicGames, Mehr2023MaximumEntropyMultiAgentDynamicGames, LeCleach2021LUCIDGames} developed \textit{inverse dynamic game} algorithms that infer unknown agent costs from interaction demonstrations via KKT conditions encoding local Nash stationarity.
\revision{Other works focus on inferring information structure \cite{KhanDFK2024LeadershipInferenceForMultiAgentInteractions} or foresightedness \cite{Armstrong2025InferringForesightednessinDynamicNoncooperativeGames} from interactions at or near local Nash stationarity.
}
However, most existing inverse game-theoretic methods 
are not directly applicable to \textit{constraint} inference. Exceptions include \cite{Peters2024ContingencyGames}, which infers \textit{a priori} unknown parameters for agent costs and constraints from a finite set of hypotheses, and \cite{LiuPeters2024AutoEncodingBayesianInverseGames}, which produces posterior distributions over such parameters. Unlike \cite{Peters2024ContingencyGames, LiuPeters2024AutoEncodingBayesianInverseGames}, our method explicitly searches over a continuous parameter space, and is guaranteed to learn an inner approximation of the true constraint set.

\section{Dynamic Games and Problem Statement}
\label{sec: Forward Dynamic Game Formulation}



\looseness-1Consider an $N$-agent, $T$-stage discrete-time \textit{forward} dynamic game $\G$, in which $x_t^i \in \R^{n_i}$ is the state vector of each agent $i \in [N] := \{1, \cdots, N\}$ at each time $t \in [T] := \{1, \cdots, T\}$. We define $x_t := (x_t^1, \cdots, x_t^N) \in \R^n$ and $u_t := (u_t^1, \cdots, u_t^N) \in \R^m$ for each $t \in [T]$, where
$n := \sum_{i \in [N]} n_i$ and $m := \sum_{i \in [N]} m_i$, and $x := (x_1, \cdots, x_T) \in \R^{nT}$ and $u := (u_1, \cdots, u_T) \in \R^{mT}$.
Finally, we define $\xi := (x, u) \in \R^{(n+m)T}$ to be the system's state-control trajectory.


Each agent $i \in [N]$ aims to minimize its cost $J^i(\xi) 
$, whose value depends on the system trajectory $\xi$. \revision{Moreover, the trajectory of each agent $i \in [N]$ must ensure that
the scalar equality constraints $C^{\eq, i} := \{h_\beta^i(\xi) = 0: \beta \in [N^{\eq, i}]\}$, and scalar inequality constraints $C_k^{\ineq, i} := \{g_{\beta,k}^i(\xi) \leq 0: \beta \in [N_k^{\ineq, i}] \}$ and $C_\urk^{\ineq, i} := \{g_{\beta,\urk}^i(\xi, \theta^\star) \leq 0: \beta \in [N_{\urk}^{\ineq, i}] \}$ are satisfied,\footnote{\revision{We assume 
all unknown constraints are \textit{in}equality constraints. An unknown equality constraint $h_\beta^i(\xi, \theta) = 0$ can be inferred using $\D$ via regression analysis and root-finding methods, or expressed as two inequality constraints ($h_\beta^i(\xi, \theta) \leq 0$ and $- h_\beta^i(\xi, \theta) \leq 0$) and learned via our method.}}, where $N^{\eq,i}$, $N_k^{\ineq,i}$, and $N_\urk^{\ineq,i}$ are the number of scalar constraints in $C^{\eq,i}$, $C_k^{\ineq,i}$, and $C_\urk^{\ineq,i}$, respectively. 
Above, the constraint $g_{\beta,\urk}^i(\cdot, \theta^\star)$ is parameterized by the parameter $\theta^\star \in \Theta$,
while $\Theta$ is a compact set of all possible parameter values\footnote{Our methods readily extend to the case where each agent's constraint set is the \textit{union of intersections} of equality and inequality constraints. For simplicity, our formulation here considers only \textit{intersections}.
}.
We stack constraint functions $g_{\beta, k}^i$, $g_{\beta, \urk}^i$, and $h_\beta^i$ across all $\beta \in [N_c]$ to form $\boldg_k^i$, $\boldg_\urk^i$, and $\boldh^i$, respectively, for each agent $i \in [N]$. We further stack across all $i \in [N]$ to form $\boldg_k$, $\boldg_\urk$, and $\boldh$, respectively. Moreover, we write $\boldg(\xi, \theta) = (\boldg_k(\xi), \boldg_\urk(\xi, \theta))$.
%
}
\hspace{-1.5mm}The constraints $\boldh^i(\cdot)$ and $\boldg^i(\cdot)$ for each agent $i \in [N]$ encode the system dynamics $x_{t+1}^i = f_t(x_t^i, u_t^i)$ 
constraining the motion
of each agent, as well as obstacle avoidance, inter-agent collision avoidance, and other constraints. 
\revision{Finally, for any parameter value $\theta \in \Theta$, we denote by $\Safe(\theta)$ (resp., $\Avoid(\theta)$) the set of all trajectories satisfying (resp., violating) the inequality constraints $\boldg_\urk(\xi, \theta) \leq 0$, i.e., 
{
\setlength{\abovedisplayskip}{3pt}
\setlength{\belowdisplayskip}{3pt}
\begin{align} \label{Eqn: Safe(theta), def}
    \Safe(\theta) &:= \bigcap_{i \in [N]} \{\xi \in \R^{(n+m)T}: \boldg_\urk(\xi, \theta) \leq 0 \}, \\ \label{Eqn: Avoid(theta), def}
    \Avoid(\theta) &:= \Safe(\theta)^c.
\end{align}
}
}
\vspace{-5mm}

The objective of the \textit{forward} dynamic game is to compute the interactive system trajectory by solving the following coupled optimization problems for each agent $i \in [N]$:
{
\setlength{\abovedisplayskip}{3pt}
\setlength{\belowdisplayskip}{3pt}
\begin{subequations} \label{Eqn: Forward Game}
\begin{align} \label{Eqn: Forward Game, Costs}
    \min_{x^i, u^i} \hspace{5mm} &J^i(\xi) \\ \label{Eqn: Forward Game, Equality Constraint}
    \text{s.t.} \hspace{5mm} &h_\beta^i(\xi) = 0, \ \forall \beta \in [N^{\eq, i}], \\ \label{Eqn: Forward Game, Inequality Constraint, Known}
    &\revision{g_{\beta, k}^i(\xi) \leq 0, \ \forall \beta \in [N_k^{\ineq, i}]}, \\ \label{Eqn: Forward Game, Inequality Constraint, Unknown}
    &\revision{g_{\beta, \urk}^i(\xi, \theta^\star) \leq 0, \ \forall \beta \in [N_\urk^{\ineq, i}]}. 
\end{align}
\end{subequations}
}
\looseness-1
We call $u^\star$ a \textit{Nash equilibrium solution} to \eqref{Eqn: Forward Game} with state
trajectory $x^\star$ if the state-control trajectory $\xi^\star := (x^\star, u^\star)$ satisfies \eqref{Eqn: Forward Game}. We call $u^\star$ a \textit{local Nash equilibrium solution} to \eqref{Eqn: Forward Game} with trajectory $x^\star$ if $\xi^\star := (x^\star, u^\star) \in \R^{(n+m)T}$ satisfies the following:
There exists a neighborhood $\mathcal{N}(\xi^\star)$ of $\xi^\star$ such that for any agent $i \in [N]$ and any feasible state-control trajectory $\{(x_t^i, u_t^i): t \in [T]\}$ of agent $i$ in $\mathcal{N}(\xi^\star)$, we have $J^i(\xi^\star) = J^i(x^\star, u^\star) \leq J^i(x^{i\star}, u^{i\star}, x^{-i\star}, u^{-i\star}) := J^i(\xi^i, \xi^{-i\star})$, where $-i$ refers to all indices in $[N]$ apart from $i$.

\textbf{Problem Statement}:
\revision{Assuming 
$\{J^i(\cdot): i \in [N]\}$, $\boldg_k(\cdot)$, $\boldg_\urk(\cdot, \cdot)$, $\boldh_k(\cdot)$, and $\Theta$ are known}, 
we aim to infer $\theta^\star$
from a set of $D$ local Nash equilibrium demonstrations $\D := \{\xi_d^{\loc}: d \in [D] \}$,
and
leverage the learned constraint information to design safe interactive motion plans.


\section{
Multi-Agent Constraint Inference
and Robust Motion Planning}
\label{sec: Constraint Inference}


Below, we formulate our constraint inference problem using KKT conditions derived from the forward dynamic game \eqref{Eqn: KKT, Forward Game} (Sec. \ref{subsec: Inverse Games for Multi-Agent Constraint Inference}).
\revision{We illustrate that the problem of learning \textit{offset-parameterized interaction constraints} in dynamic games, e.g., various collision-avoidance constraints, can be reformulated as MILPs (Sec. \ref{subsec: MILP Reformulation for Unions of Offset-Parameterized Constraints}).
}
\hspace{-1mm}Then, we present methods for extracting 
guaranteed safe and unsafe trajectories, rejecting
parameter values inconsistent with the demonstrated interactions, and using the extracted knowledge to design safe plans under constraint uncertainty (Sec. \ref{subsec: Volume Extraction Over Trajectories and Parameters for Motion Planning}). 
\revision{Finally, we present fundamental limitations of learning interaction constraints from demonstrations (Sec. \ref{subsec: Theoretical Limitations of Learnability for Multi-Agent Constraints}).}

\vspace{-3pt}
\subsection{KKT Conditions for Multi-Agent Constraint Inference}
\label{subsec: Inverse Games for Multi-Agent Constraint Inference}
Since each interaction demonstration $\xi_d^\loc$ in $\D$ is at local Nash equilibrium, it must satisfy the KKT conditions associated with 
\eqref{Eqn: Forward Game}. Thus, for each $\xi_d^\loc \in \D$, there exist Lagrange multipliers $\boldlambda_{d, k}^i := \{\lambda_{d, \beta, k}^i, \forall \beta \in [N_k^{\ineq, i}] \}$, $\boldlambda_{d, \urk}^i := \{\lambda_{d, \beta, \urk}^i, \ \forall \beta \in [N_\urk^{\ineq, i}] \}$, and $\boldnu_d^i := \{\nu_{d, \beta}^i, \ \forall \beta \in [N^{\eq, i}] \}$ satisfying the following equations with $\theta = \theta^\star$, $\xi = \xi_d^\loc$:
{
\setlength{\abovedisplayskip}{3pt}
\setlength{\belowdisplayskip}{3pt}
\begin{subequations} \label{Eqn: KKT, Forward Game}
\begin{align} \label{Eqn: KKT, Forward, Primal Feasibility}
    &\boldh^i(\xi) = 0, \hspace{5mm} \boldg_k^i(\xi) \leq 0, \hspace{5mm} \boldg_\urk^i(\xi, \theta) \leq 0, \\ \label{Eqn: KKT, Forward, Lagrange Multiplier non-negativity}
    &\boldlambda_{d, k}^i, \boldlambda_{d, \urk}^i \geq 0, \\ \label{Eqn: KKT, Forward, Complementary Slackness}
    &\boldlambda_{d, k}^i \odot \boldg_k^i(\xi) = 0, \hspace{5mm} \boldlambda_{d, \urk}^i \odot \boldg_\urk^i(\xi, \theta) = 0, \\ \label{Eqn: KKT, Forward, Stationarity}
    &\nabla_{\xi^i} J^i(\xi) + (\boldlambda_{d, k}^i)^\top \nabla_{\xi^i} \boldg_k^i(\xi) \\ \nonumber
    & \hspace{2mm} + (\boldlambda_{d, \urk}^i)^\top \nabla_{\xi^i} \boldg_\urk^i(\xi, \theta) 
    + (\boldnu_d^i)^\top \nabla_{\xi^i} \boldh^i(\xi) = 0.
\end{align}
\end{subequations}
}

Here, $\odot$ denotes element-wise multiplication, and $\nabla$ defines the gradient of any differentiable function $\textbf{f}: \R^n \ra \R^m$, i.e., $[\nabla_x \textbf{f}]_{i'j'} := \frac{\partial f_{i'}}{\partial x_{j'}}$ for each $i' \in [m]$ and $j' \in [n]$. Above, \eqref{Eqn: KKT, Forward, Primal Feasibility} encodes primal feasibility, \eqref{Eqn: KKT, Forward, Lagrange Multiplier non-negativity} encodes dual feasibility,
\eqref{Eqn: KKT, Forward, Complementary Slackness} encodes complementary slackness, while \eqref{Eqn: KKT, Forward, Stationarity} encodes first-order KKT stationarity. 
We concatenate the dual variables $\boldlambda_{d,k}^i$, $\boldlambda_{d,\urk}^i$, and $\boldnu_d^i$ across $d \in [D]$ to form $\boldlambda_k^i$, $\boldlambda_\urk^i$, and $\boldnu^i$, respectively, and we further concatenate across $i \in [N]$ to form $\boldlambda_k$, $\boldlambda_\urk$, and $\boldnu$, respectively.
We also define:
\revision{
\setlength{\abovedisplayskip}{3pt}
\setlength{\belowdisplayskip}{3pt}
\begin{align} 
    \KKT^i(\xi_d^\loc) &:= \{(\theta, \boldlambda_k, \boldlambda_{\urk}, \boldnu): (\theta, \boldlambda_{d,k}^i, \boldlambda_{d,\urk}^i, \boldnu_d^i) \\ 
    \nonumber
    &\hspace{1cm} \text{ satisfies } \eqref{Eqn: KKT, Forward Game} \text{ with $\xi = \xi_d^\loc$} \}, \\ 
    \KKT(\D) &:= \bigcap\nolimits_{i \in [N]} \bigcap\nolimits_{d \in [D]} \KKT^i(\xi_d^\loc).
\end{align}
}
\hspace{-1mm}Conversely, we say that $\xi \in \R^{(n+m)T}$ is at \textit{local Nash stationarity} with respect to the underlying constraints $\boldh(\cdot)$, $\boldg_k(\cdot)$, and $\boldg_\urk(\cdot, \theta)$ if there exist $\boldnu$, $\boldlambda_k$, and $\boldlambda_\urk$ such that $(\theta, \boldlambda_k, \boldlambda_\urk, \boldnu) \in \KKT(\{\xi\})$.

Given the set $\D$ of local Nash equilibrium demonstrations, there must exist Lagrange multipliers $\{\boldlambda_{d, k}^{i^\star}, \boldlambda_{d, \urk}^{i^\star}, \boldnu_d^{i^\star}: i \in [N], d \in [D]\}$ such that $(\theta^\star, \boldlambda_k^{i^\star}, \boldlambda_\urk^{i^\star}, \boldnu^{i^\star})$ solves:
{
\setlength{\abovedisplayskip}{3pt}
\setlength{\belowdisplayskip}{3pt}
\begin{subequations}
\label{Eqn: KKT, Inverse, Optimal}
\begin{align} \label{Eqn: KKT, Inverse, Optimal, Objective}
    \text{find} \hspace{5mm} &\theta, \boldlambda_k, \boldlambda_\urk, \boldnu, \\ \label{Eqn: KKT, Inverse, Optimal, Constraints}
    \text{s.t.} \hspace{5mm} &(\theta, \boldlambda_k, \boldlambda_\urk, \boldnu) \in \KKT(\D).
\end{align}
\end{subequations}
}
\revision{
\looseness-1Conversely, the solution set of \eqref{Eqn: KKT, Inverse, Optimal} encodes parameter values $\theta$ consistent with the local Nash stationarity of $\D$, 
and contains the true parameter value $\theta^\star$.
We denote by $\F(\D)$ the set of parameter values compatible with $\KKT(\D)$, i.e., 
{
\setlength{\abovedisplayskip}{3pt}
    \setlength{\belowdisplayskip}{3pt}
\begin{align}
    \F(\D) &:= \{\theta \in \Theta: \exists \ \boldlambda_k, \boldlambda_\urk, \boldnu \st \\ \nonumber
    &\hspace{2cm} (\theta, \boldlambda_k, \boldlambda_\urk, \boldnu) \in \KKT(\D). \}
\end{align}}
}

\vspace{-5mm}
\revision{
Thm. \ref{Thm: Conservativeness of Safe and Unsafe Set Recovery from KKT, Inverse, Optimal} below establishes that the ground truth safe set $\Safe(\theta^\star)$ is guaranteed to be inner (i.e., conservatively) approximated by the set of all trajectories satisfying 
$\boldg(\xi, \theta) \leq 0$ for all $\theta \in \F(\D)$. Similarly, the ground truth avoid set $\Avoid(\theta^\star)$ is guaranteed to be inner approximated by the set of all trajectories violating 
$\boldg(\xi, \theta) \leq 0$ for all $\theta \in \F(\D)$.
}

\begin{theorem}(\textbf{\textit{Conservativeness of Safe and Unsafe Set Recovery from} \eqref{Eqn: KKT, Inverse, Optimal}}) 
\label{Thm: Conservativeness of Safe and Unsafe Set Recovery from KKT, Inverse, Optimal}
Define the learned set of \textit{guaranteed safe} (resp., \textit{unsafe}) trajectories, denoted by $\G_s(\D)$ (resp., $\G_\urs(\D)$), to be the set of trajectories that are safe (resp., unsafe) with respect to all parameters $\theta \in \F(\D)$, i.e.,:\footnote{Note that $\G_\urs(\D) \ne \G_s(\D)^c$, i.e., it is possible for a given trajectory to be neither guaranteed safe nor guaranteed unsafe.}
{
\setlength{\abovedisplayskip}{3pt}
\setlength{\belowdisplayskip}{3pt}
\begin{align} \label{Eqn: Gs, Def}
    \G_s(\D) &:= \bigcap_{\theta \in \F(\D)} \big\{ \xi \in \R^{(n+m)T}: \boldg(\xi, \theta) \leq 0 \big\}, \\ \label{Eqn: Gurs, Def}
    \G_\urs(\D) &:= \bigcap_{\theta \in \F(\D)} \big\{ \xi \in \R^{(n+m)T}: 
    \boldg(\xi, \theta) > 0 \big\}.
\end{align}
}
Then $\G_s(\D) \subseteq \Safe(\theta^\star)$ and $\G_\urs(\D) \subseteq \Avoid(\theta^\star)$.
\end{theorem}

\revision{
\hspace{-1mm}
\begin{proof}
By definition of $\G_s(\D)$ in \eqref{Eqn: Gs, Def} and $\Safe(\theta)$ in \eqref{Eqn: Safe(theta), def}, $\G_s(\D) = \bigcap_{\theta \in \F(\D)} \Safe(\theta)$.
Thus, to prove $\G_s(\D) \subseteq \Safe(\theta^\star)$, it suffices to verify that $\theta^\star \in \F(\D)$.
Since $\D$ contains local Nash equilibrium trajectories of the dynamic game with constraints parameterized by 
$\theta^\star$, for any $\xi_d \in \D$, there exist 
$\boldlambda_{d,k}, \boldlambda_{d,\urk}, \nu_d$ such that $(\theta^\star, \boldlambda_{d,k}, \boldlambda_{d,\urk}, \nu_d) \in \KKT(\{\xi_d\})$. Thus,
$\theta^\star \in \F(\D)$, 
so $\G_s(\D) \subseteq \Safe(\theta^\star)$.
%
Similarly, by definition of $\G_\urs(\D)$ in \eqref{Eqn: Gurs, Def} and $\Avoid(\theta)$ in \eqref{Eqn: Avoid(theta), def}, $\G_\urs(\D) = \bigcap_{\theta \in \F(\D)} \Avoid(\theta)$.
Thus, to prove $\G_s(\D) \subseteq \Safe(\theta^\star)$, it suffices to verify $\theta^\star \in \F(\D)$, which we have done above. 
\end{proof}
}

\looseness-1For many constraints (e.g., collision-avoidance), the primal feasibility conditions \eqref{Eqn: KKT, Forward, Primal Feasibility}
restrict the relative agent positions $\{x_t^j - x_t^i: \ \forall i, j \in [N], j \ne i, \forall t \in [T] \}$ 
to lie inside a constraint set characterized by a union of polytopes. In this case, 
\eqref{Eqn: KKT, Inverse, Optimal} can be reformulated as a mixed-integer linear program (MILP)
or a mixed-integer bilinear program (MIBLP), 
and solved via off-the-shelf solvers \cite{gurobi} (see Sec. \ref{subsec: MILP Reformulation for Unions of Offset-Parameterized Constraints} and App. \ref{subsec: App, MILPs for Inferring Polytopic Constraints}). The difference in the optimization problem class depends on the parameterization of the unknown constraint.


\begin{remark}
\label{Remark: Learning Constraints from Suboptimal Data}
\revision{
If the given demonstrations are not at local Nash stationarity,
we 
relax the stationarity condition \eqref{Eqn: KKT, Forward, Stationarity} in \eqref{Eqn: KKT, Inverse, Optimal, Constraints} into a cost defined using the following stationarity terms:
\begin{align}
    \stat^i(\xi_d^{\loc}, \theta) &:= \nabla_{\xi^i} J^i(\xi) + (\boldlambda_{d, k}^i)^\top \nabla_{\xi^i} \boldg_k^i(\xi) \\ \nonumber
    & \hspace{6mm} + (\boldlambda_{d, \urk}^i)^\top \nabla_{\xi^i} \boldg_\urk^i(\xi, \theta) 
    + (\boldnu_d^i)^\top \nabla_{\xi^i} \boldh^i(\xi)
\end{align}
across all $d \in [D]$ and $i \in [N]$. Concretely, we relax 
\eqref{Eqn: KKT, Inverse, Optimal} into the following optimization program:
}
\begin{subequations}
\label{Eqn: KKT, Inverse, Suboptimal}
\begin{align} \label{Eqn: KKT, Inverse, Suboptimal, Objective}
    \min_{\theta, \boldlambda_k, \boldlambda_\urk, \boldnu} \hspace{5mm} &\sum_{d \in [D]} \sum_{i \in [N]} \Vert \stat^i(\xi_d^\loc, \theta) \Vert_1, \\ \label{Eqn: KKT, Inverse, Suboptimal, Constraints}
    \text{s.t.} \hspace{1.2cm} &\eqref{Eqn: KKT, Forward, Primal Feasibility}-\eqref{Eqn: KKT, Forward, Complementary Slackness}, \hspace{5mm} i \in [N], d \in [D].
\end{align}
\end{subequations}
\end{remark}

\vspace{-3mm}
\revision{
\begin{remark} \label{Remark: Learning Constraints when Parameterization Type is Unknown}
Even with \textit{inaccurate} knowledge of the constraint parameterization, 
the constraint learner may 
still approximate constraints effectively by
selecting a base constraint shape (e.g., polytope) as a building block, and incrementally introducing more base shapes until the stationarity error \eqref{Eqn: KKT, Inverse, Suboptimal, Objective} is sufficiently small.
(see Sec. \ref{subsec: Constraint Inference Despite Misspecified Constraint Parameterizations} and Fig. \ref{fig:boxes_for_ellipse}).
\end{remark}
}

\vspace{-2pt}
\subsection{MILP Reformulation for Unions of Offset-Parameterized Constraints}
\label{subsec: MILP Reformulation for Unions of Offset-Parameterized Constraints}
\vspace{-3pt}

Suppose
all agents have equal state dimension (i.e., $n_i = n_j \ \forall i, j \in [N]$), and 
the unknown constraints $\boldg_\urk^i(\xi, \theta) \leq 0$ encode that $\forall t \in [T]$, the state of each other agent $j \in [N] \backslash \{i\}$ relative to agent $i$ lies outside a polytopic collision avoidance set with $N_c$ sides, each of which is represented as a scalar constraint.
Concretely, feasible system-level trajectories must lie outside avoid sets:
\revision{
{
\setlength{\abovedisplayskip}{3pt}
\setlength{\belowdisplayskip}{3pt}
\begin{align}
    \Avoid^{i,j,t}(\theta^\star) := \bigcap_{\beta=1}^{N_c} \{ \xi \in \R^{(n+m)T}: \bolda_\beta^{i,j,t\top} \xi < b_\beta^i(\theta^\star) \}
\end{align}
}
}
\hspace{-1mm}for all $i, j \in [N]$, $j \ne i$, $t \in [T]$, where $\bolda_\beta^{i,j,t} \in \R^{(n+m)T}$, $b_\beta^i(\theta^\star) \in \R$, $\forall i, j \in [N]$, $j \ne i$, $t \in [T]$, and each $b_\beta^i(\theta^\star)$ is specified by an unknown parameter $\theta^\star$.
Thus, trajectories $\xi$ must evade the avoid set: 
\vspace{-1mm}
\revision{
\begin{align} \label{Eqn: Avoid set, as union of agent and time specific avoid sets}
    \Avoid(\theta^\star) := 
    \bigcup_{i, j \in [N], j \ne i, t \in [T]} \Avoid^{i,j,t}(\theta^\star),
\end{align}
}
\hspace{-1mm}or equivalently, stay within the safe set $\Safe(\theta^\star) := (\Avoid(\theta^\star))^c$:
\begin{align} \label{Eqn: Safe Set, 2}
    \hspace{-5pt}
    \Safe(\theta^\star) &:= 
    \bigcap_{i, j \in [N], j \ne i, t \in [T]} \bigcup_{\beta \in [N_c]} \big\{ 
    \xi: 
    \bolda_\beta^{i,j,t\top} \xi \geq b_\beta^i(\theta^\star) \big\}
\end{align}
To recover $\theta^\star$,
we now reformulate the KKT conditions \eqref{Eqn: KKT, Forward Game} as an MILP by extending the approach of \cite[Sec. IV-B]{Chou2020LearningConstraintsFromLocallyOptimalDemonstrationsUnderCostFunctionUncertainty} to accommodate our multi-agent constraint learning problem.
\revision{
For notational ease, in the rest of Sec. \ref{subsec: MILP Reformulation for Unions of Offset-Parameterized Constraints},
we focus on learning interaction constraints between $N = 2$ agents from a single demonstration $\xi^\loc$ (i.e., $D = 1$). 
We set $\bolda_\beta^{i,t} := \bolda_\beta^{i,3-i,t}$ and $\Avoid^{i,t}(\theta^\star) := \Avoid^{i,3-i,t}(\theta^\star)$ $\forall i \in [2], t \in [T]$.
For a treatment without the above restrictions, see App. \ref{subsec: App, MILPs for Inferring Polytopic Constraints}.
}

\begin{enumerate}
    \item \textbf{Reformulating primal feasibility \eqref{Eqn: KKT, Forward, Primal Feasibility}}:

    \revision{
    $\hspace{5mm}$ First, we rewrite \eqref{Eqn: KKT, Forward, Primal Feasibility} via the big-M formulation \cite{Bertsimas1998IntroductiontoLinearOptimization, Chou2020LearningConstraintsFromLocallyOptimalDemonstrationsUnderCostFunctionUncertainty}, by introducing binary vectors $\boldz^{i,t} = (z_\beta^{i,t} \in \{0, 1\} : \beta \in [N_c]) \in \{0, 1\}^{N_c}$, for each $i \in [2]$, $t \in [T]$, as follows
    ---$\forall i \in [2], t \in [T]$:
    \begin{subequations} \label{Eqn: Primal Feasibility, Reformulation}
    \begin{align} 
        &(\boldA^{i,t})^\top \xi^\loc \geq \boldb^{i,t}(\theta) - M(\textbf{1}_{N_c} - \boldz^{i,t}), \\
        &\sum\nolimits_{\beta = 1}^{N_c} z_\beta^{i,t} \geq 1, 
    \end{align}
    \end{subequations}
    Above, $M \gg 0$, 
    $\textbf{1}_{N_c} := (1, \cdots, 1) \in \R^{N_c}$,
    while $\boldA^{i,t}$ $\in \R^{(n+m)T \times N_c}$ and $\boldb^{i,t}(\theta) \in \R^{N_c}$ are the concatenations of $\bolda_\beta^{i,t}$ and $b_\beta^{i,t}(\theta)$, respectively, across all $\beta \in [N_c]$.
    }

    \item \textbf{Reformulating complementary slackness \eqref{Eqn: KKT, Forward, Complementary Slackness}}:

    \revision{
    $\hspace{5mm}$
    Let $\boldlambda_{\urk}^{i,t} = (\lambda_{\beta, \urk}^{i,t} \geq 0: \beta \in [N_c]) \in \R^{N_c}$ denote the dual variables associated with the inequalities $(\bolda_\beta^{i,t})^\top \xi \geq b_\beta^{i,t}(\theta)$, across $\beta \in [N_c]$.  
    Since the safe set definition \eqref{Eqn: Safe Set, 2} involves a union over $\beta \in [N_c]$ $\forall i \in [2]$, $t \in [T]$, the constraint $\{ \xi \in \R^{(n+m)T}: (\bolda_\beta^{i,t})^\top \xi^\loc \geq b_\beta^{i,t}(\theta) \}$, for each $i \in [2]$, $t \in [T]$, might only be enforced for a strict subset of the indices $\beta \in [N_c]$, which we denote by $\mathcal{B}^{i,t} \subseteq [N_c]$. Likewise, $\forall i \in [2]$, $t \in [T]$, complementary slackness \eqref{Eqn: KKT, Forward, Complementary Slackness} should only be enforced for indices in $\mathcal{B}^{i,t}$, and in turn, terms of the form $\lambda_{\beta, \urk}^{i,t} \nabla_{\xi^{i,t}} g_{\beta, \urk} (\xi, \theta)$ are only included in the stationarity condition \eqref{Eqn: KKT, Forward, Stationarity} if $\beta \in \mathcal{B}^{i,t}$. Thus, we use a big-M formulation, in which we introduce binary vectors 
    $\hat \boldz_{1}^{i,t} = (\hat z_{1, \beta}^{i,t}: \beta \in [N_c]) \in \{0, 1\}^{N_c}$ and $\hat \boldz_{2}^{i,t} = (\hat z_{2,\beta}^{i,t}: \beta \in [N_c]) \in \{0, 1\}^{N_c}$
    to \textit{encode} 
    \eqref{Eqn: KKT, Forward, Complementary Slackness}, and binary vectors
    $\boldq^{i,t} = (q_\beta^{i,t}: \beta \in [N_c]) \in \{0, 1\}^{N_c}$ 
    to \textit{enforce} 
    \eqref{Eqn: KKT, Forward, Complementary Slackness}. Concretely, we rewrite the complementary slackness conditions \eqref{Eqn: KKT, Forward, Complementary Slackness} as follows---$\forall i \in [2]$, $t \in [T]$:    
    \vspace{-4mm}
    \begin{subequations} \label{Eqn: Complementary Slackness, Reformulation}
    \begin{align}
        &\begin{bmatrix}
            \boldlambda_{\urk}^{i,t} \\
            \boldA^{i,t\top} \xi^\loc - \boldb^{i,t}(\theta)
        \end{bmatrix} 
        \leq M \begin{bmatrix}
            \hat \boldz_{1}^{i,t} \\ \hat \boldz_{2}^{i,t}
        \end{bmatrix}, \\
        &\hat z_{1, \beta}^{i,t} + \hat z_{2, \beta}^{i,t} \leq 2 - q_\beta^{i,t}, \ \ \forall \ \beta \in [N_c], \\ 
        &\sum\nolimits_{\beta' = 1}^{N_c} q_{\beta'}^{i,t} \geq 1.
    \end{align}
    \end{subequations}
    }
    \vspace{-4mm}

    \item \textbf{Reformulating stationarity \eqref{Eqn: KKT, Forward, Stationarity}}:

    \revision{
    $\hspace{5mm}$
    As noted above, the stationarity condition \eqref{Eqn: KKT, Forward, Stationarity} includes the term $\lambda_{\beta, \urk}^{i,t} \nabla_{\xi^{i,t}} g_{\beta, \urk} (\xi, \theta)$ \textit{only if} $\beta \in \mathcal{B}^{i,t}$. To enforce this condition, $\forall i \in [2]$, $t \in [T]$, we define $L^{i,t} \in$ $\R^{N_c \times (n_i + m_i)T}$ as the matrix whose $\beta$-th row equals $\lambda_{\beta, \urk}^{i,t} \nabla_{\xi^{i,t}} (b_\beta^{i,t}(\theta) - \bolda_\beta^{i,t\top} \xi) |_{\xi = \xi^\loc}$.
    We then rewrite the term ${\boldlambda_{\urk}^{i,t}}^\top \nabla_{\xi^{i,t}} \boldg_\urk^{i,t}(\xi, \theta)$ in \eqref{Eqn: KKT, Forward, Stationarity} as follows---$\forall i \in [2]$:
    {
    \setlength{\abovedisplayskip}{3pt}
    \setlength{\belowdisplayskip}{3pt}
    \begin{align} \label{Eqn: L, Bilinear Terms}
        &\sum_{t \in [T]} \boldlambda_{\urk}^{i,t\top} \nabla_{\xi^{i,t}} \big( b^{i,t}(\theta) - \boldA^{i,t\top} \xi \big)^\top |_{\xi = \xi^\loc} \\ \nonumber
        = \ &\sum_{t \in [T]} \boldq^{i,t\top} L^{i,t}.
    \end{align}
    } 
    }
    \revision{
    By replacing the bilinear terms $\boldq^{i,t\top} L^{i,t}$ in \eqref{Eqn: L, Bilinear Terms} with slack variables and adding constraints, we can reformulate \eqref{Eqn: L, Bilinear Terms} in a linear manner. 
    Concretely, $\forall i \in [2]$, $t \in [T]$, we define $\tilde \boldq^{i,t} := 1 - \boldq^{i,t}$ and introduce slack variables $R^{i,t} \in \R^{N_c \times (n_i + m_i)}$.
    Let $R_{\beta,\ell}^{i,t}$ and $L_{\beta,\ell}^{i,t}$ denote the ($\beta$-th row, $\ell$-th column) entry of $R^{i,t}$ and $L^{i,t}$, respectively, $\forall \beta \in [N_c], \ell \in [n]$.
    We assume each
    $L_{\beta,\ell}^{i,t}$ is component-wise bounded below and above by some $\underM$ and $\overM$, respectively.
    We then introduce the constraints below, $\forall \beta \in [N_c]$, $\ell \in [n]$, to linearize $\boldq^{i,t\top} L^{i,t}$:
    {
    \setlength{\abovedisplayskip}{3pt}
    \setlength{\belowdisplayskip}{4pt}
    \begin{subequations} 
    \begin{align}
        \min\{0, \underM \} &\leq R_{\beta,\ell}^{i,t} \leq \overM, \\
        \underM q_{\beta}^{i,t} &\leq R_{\beta,\ell}^{i,t} \leq \overM q_{\beta}^{i,t}, \\
        L_{\beta,\ell}^{i,t} - \tilde q_{\beta}^{i,t} \overM &\leq R_{\beta,\ell}^{i,t} \leq L_{\beta,\ell}^{i,t} - \tilde q_{\beta}^{i,t} \underM \\
        R_{\beta,\ell}^{i,t} &\leq L_{\beta,\ell}^{i,t} + \tilde q_{\beta}^{i,t} \overM
    \end{align}
    \end{subequations}
    }
    }
    \vspace{-6.5mm}
\end{enumerate}

\subsection{Volume Extraction Over Trajectories and Parameters for Motion Planning}
\label{subsec: Volume Extraction Over Trajectories and Parameters for Motion Planning}

When the given demonstrations are insufficiently informative to allow the unambiguous recovery of the true constraint parameter $\theta^\star$, an incorrect point estimate of $\theta^\star$ may cause motion plans generated downstream to be unsafe
(Sec. \ref{sec: Experiments}, Fig. \ref{fig:rebuttal___volume_extraction_robustness} in App. \ref{subsubsec: App, Double Integrator Simulations}
\cite{ArXivPaper}).
To address this issue,
we describe methods to recover inner approximations of the guaranteed safe set $\G_s(\D)$, by 
\revision{extracting \textit{volumes} of guaranteed safe trajectories from a set of queried trajectories (Thm. \ref{Thm: Volume Extraction Over Trajectories}) or parameter values (Thm. \ref{Thm: Volume Extraction, Theta, Generates Conservative Safe Set}). 
Strategies for querying trajectories or constraint parameters are provided in Remark \ref{Remark: Strategies for Querying Trajectories or Constraint Parameters}.}
\begin{theorem}[\textbf{Volume Extraction Over Trajectories}] \label{Thm: Volume Extraction Over Trajectories}
\revision{
\begin{subequations} \label{Eqn: Volume Extraction for Safe Set, Trajectories}
\hspace{-1mm}Given a set of queried trajectories $\Q_\xi := \{\xi_q \in \R^{(n+m)T}: q \in [Q_\xi] \}$, for each $\xi_q \in \Q_\xi$, define:
\begin{align}
    \epsilon_q := \min_{\theta \in \F(\D), \xi}. \hspace{5mm} &\Vert \xi - \xi_q \Vert_\infty \\
    \label{Eqn: Volume Extraction for Safe Set, Trajectory xi is unsafe}
    \text{s.t.} \hspace{5mm} &\boldg(\xi, \theta) > 0.
\end{align}
\end{subequations}
For each $\xi_q \in \Q_\xi$,
let
$B_{\epsilon_{q}}^\infty(\xi_q)$ be the open $\infty$-norm 
neighborhood 
of radius $\epsilon_q$ centered at $\xi_q$. Then: 
\begin{align} \label{Eqn: Volume Extraction, Trajectory xi, inner approx for Safe set}
    \bigcup\nolimits_{\xi_q \in \Q_\xi} B_{\epsilon_q}^\infty(\xi_q) \subseteq \G_s(\D).
\end{align}
}
\end{theorem}
\vspace{-3mm}

\revision{
In words, if $\xi_q \in \R^{(n+m)T}$
were safe, \eqref{Eqn: Volume Extraction for Safe Set, Trajectories} computes the largest hypercube in $\R^{(n+m)T}$ that is centered at $\xi_q$ and contained in $\G_s(\D)$; if $\xi_q$ were unsafe, then \eqref{Eqn: Volume Extraction for Safe Set, Trajectories} returns zero. 
Moreover, one can inner approximate the guaranteed safe set $\G_s(\D)$ by taking the union of the $B_{\epsilon_{q}}^\infty(\xi_q)$ hypercubes. For the proof of Thm. \ref{Thm: Volume Extraction Over Trajectories}, see App. \ref{subsubsec: App, Volume Extraction of Safe Trajectories for Motion Planning}.
}

To inner approximate the learned \textit{unsafe} set $\G_\urs(\D)$, we replace \eqref{Eqn: Volume Extraction for Safe Set, Trajectory xi is unsafe} with the condition that the trajectory $\xi$ is safe, i.e., $\textbf{g}_k(\xi, \theta) \leq 0$, and proceed as described in Thm. \ref{Thm: Volume Extraction Over Trajectories} above.

Alternatively, under some constraint parameterizations, the volume extraction problem is more naturally posed and easily solved in parameter ($\theta$) space than in trajectory ($\xi$) space, as is the case for the line-of-sight constraint
(see Sec. \ref{subsec: Unicycle Simulations}). In this alternative approach, \revision{formalized as Thm. \ref{Thm: Volume Extraction, Theta, Generates Conservative Safe Set} and detailed below}, we iteratively reject parameter values in $\Theta$ that are incompatible with the interaction demonstrations in $\D$.



\begin{theorem}[\textbf{Volume Extraction Over Parameter Space}] 
\label{Thm: Volume Extraction, Theta, Generates Conservative Safe Set}
\revision{
Given a set of queried parameters $\Q_\theta := \{\theta_q \in \Theta: q \in [Q_\theta] \}$, for each $\theta_q \in \Q_\theta$,
define:
\begin{align} \label{Eqn: Volume Extraction for Safe Set, Theta}
    r_q := \min_{\theta \in \F(\D)} &\Vert \theta - \theta_q \Vert_\infty.
\end{align}
Then:
{
\small
\begin{align} \label{Eqn: Volume Extraction, theta, generates Safe set}
    \bigcap_{\theta \in \F(\D) \backslash \bigcup_{q \in [Q_\theta]} B_{r_q}^\infty(\theta_q)} \big\{ \xi \in \R^{(n+m)T}: \boldg(\xi, \theta) \leq 0 \big\} = \G_s(\D).
\end{align}
}
}
\end{theorem}
\vspace{-3mm}
\revision{
In words, if a parameter value $\theta_q \in \Q_\theta$ were incompatible with the local Nash stationarity of 
$\D$ (i.e., if $\theta_q \not\in \F(\D)$), then \eqref{Eqn: Volume Extraction for Safe Set, Theta} characterizes the largest hypercube $B_{r_q}^\infty(\theta_q)$ in $\Theta$, centered at $\theta_q$, 
such that all 
$\theta \in B_{r_q}^\infty(\theta_q)$  are incompatible with the local Nash stationarity of $\D$. 
Otherwise,
if $\theta_q \in \F(\D)$, then \eqref{Eqn: Volume Extraction for Safe Set, Theta} returns zero. 
Thus, 
the true parameter $\theta^\star$ must lie outside of all hypercubes of side length $2r_q$ centered at $\theta_q$, i.e., outside $\bigcup_{q \in [Q_\theta]} B_{r_q}^\infty(\theta_q)$.
Finally, Thm. \ref{Thm: Volume Extraction, Theta, Generates Conservative Safe Set} states that the guaranteed safe set $\G_s(\D)$ can be computed as the set of all trajectories marked safe by feasible parameters outside $\bigcup_{q \in [Q_\theta]} B_{r_q}^\infty(\theta_q)$.
For the proof of Thm. \ref{Thm: Volume Extraction, Theta, Generates Conservative Safe Set}, see App. \ref{subsubsec: App, Volume Extraction of Safe Trajectories for Motion Planning}
\cite{ArXivPaper}.
}
\vspace{-3mm}

\vspace{-4mm}
\revision{
\begin{remark} \label{Remark: Strategies for Querying Trajectories or Constraint Parameters}
In practice, to form $\Q_\xi$, one can sample safe trajectories from the vicinity of each demonstration trajectory in $\D$. To form $\Q_\theta$, one can sample parameter values $\theta$ that characterize constraints under which some demonstration trajectory in $\D$ would be unsafe.
\end{remark}
}


The guaranteed unsafe set $\G_\urs(\D)$ can be similarly characterized via volume extraction over $\Theta$. 
Since the extracted estimates of $\G_s(\D)$ and $\G_\urs(\D)$ (termed $\hat \G_s(\D)$ and $\hat \G_\urs(\D)$ below, respectively) 
either equal or 
inner approximate
$\G_s(\D)$ and $\G_\urs(\D)$, respectively, they can be \textit{directly} used by a motion planner downstream to verify constraint satisfaction or violation. 
Concretely, any generated trajectory in $\hat \G_s(\D)$ is guaranteed safe, while any trajectory outside $\hat \G_\urs(\D)$ is at least not guaranteed to be unsafe.
Alternatively, a motion planner can design safe trajectories 
via \textit{implicit} constraint checking, 
e.g., via Model Predictive Path Integral (MPPI) control \cite{
Williams2016AggressiveDrivingwithMPPI
}, summarized as thus: 
We sample i.i.d. control sequences near a given nominal control sequence, 
and update our nominal controls to be a convex combination of the sampled controls, 
with weights computed using the cost and degree of constraint violation incurred along each corresponding state trajectory.
For details, see App. \ref{subsec: App, Robust Motion Planning via Implicit Constraint Checking}
\cite{ArXivPaper}.

\subsection{Theoretical Limitations of Constraint Learnability}
\label{subsec: Theoretical Limitations of Learnability for Multi-Agent Constraints}

Here, we present conditions under which the constraint parameter $\theta$ is provably irrecoverable, thus establishing theoretical limitations on constraint learnability (Thm. \ref{Thm: Limitations of Learnability}). 

\begin{theorem}[\textbf{Limitations of Learnability}]
\label{Thm: Limitations of Learnability}
\revision{
For any $\theta \in \Theta$, let $\D(\theta)$ denote the set of local Nash stationary trajectories corresponding to the constraints $\boldh(\xi) = 0$, $\boldg_k(\xi) \leq 0$, and $\boldg_\urk(\xi, \theta) \leq 0$, i.e.,:
{
\setlength{\abovedisplayskip}{3pt}
\setlength{\belowdisplayskip}{3pt}
\begin{align} \label{Eqn: D(theta), definition}
    \D(\theta) &:= \{\xi \in \R^{(n+m)T}: \ \exists \ \boldlambda_k, \boldlambda_\urk, \boldnu \\ \nonumber
    &\hspace{1cm} \st (\theta, \boldlambda_k, \boldlambda_\urk, \boldnu) \in \KKT(\{\xi\}) \}.
\end{align}
}
\hspace{-2mm}Suppose the constraint vector $\boldg_\urk(\cdot, \cdot)$ can be partitioned into two components, $\boldg_\urk^{\paren{1}}(\cdot, \cdot)$ and $\boldg_\urk^\paren{2}(\cdot, \cdot)$, such that 
the constraint $\boldg_\urk^\paren{2}(\xi, \theta^\star) \leq 0$ is strictly looser than the remaining constraints $\boldh(\xi) = 0$, $\boldg_k(\xi) \leq 0$, $\boldg_\urk^\paren{1}(\xi, \theta^\star) \leq 0$, 
i.e.,:
{
\setlength{\abovedisplayskip}{4pt}
\setlength{\belowdisplayskip}{4pt}
\begin{align} \nonumber
    &\{\xi \in \R^{(n+m)T}: \boldh(\xi) = 0, \boldg_k(\xi) \leq 0, \boldg_\urk^\paren{1}(\xi, \theta^\star) \leq 0 \} \\ \label{Eqn: g2 subsumed by else, theta star}
    \subseteq \ &\{\xi \in \R^{(n+m)T}: \boldg_\urk^\paren{2}(\xi, \theta^\star) < 0 \}.
\end{align}
}
\hspace{-1mm}Then, for any parameter value $\theta$ satisfying:
{
\setlength{\abovedisplayskip}{4pt}
\setlength{\belowdisplayskip}{4pt}
\begin{align} \label{Eqn: g1 same across theta, theta star}
    &\boldg_\urk^\paren{1}(\cdot, \theta^\star) = \boldg_\urk^\paren{1}(\cdot, \theta), \\ \nonumber
    &\{\xi \in \R^{(n+m)T}: \boldh(\xi) = 0, \boldg_k(\xi) \leq 0, \boldg_\urk^\paren{1}(\xi, \theta^\star) \leq 0 \} \\ \label{Eqn: g2 subsumed by else, theta}
    &\hspace{5mm} \subseteq \{\xi \in \R^{(n+m)T}: \boldg_\urk^\paren{2}(\xi, \theta) < 0 \},
\end{align}
}
we have $\D(\theta^\star) = \D(\theta)$.
}
\end{theorem}

In words, each local Nash stationary demonstration corresponding to
$\theta^\star$ is also at local Nash stationarity under
$\theta$, and vice versa. Thus, when inferring constraints from demonstrations, 
$\theta$ will always remain as a valid alternative to 
$\theta^\star$.
The proof of Thm. \ref{Thm: Limitations of Learnability}, as well as a motivating example, are presented in App. \ref{subsubsec: App, Theoretical Limitations of Learnability}
\cite{ArXivPaper}.


\section{Experiments}
\label{sec: Experiments}

\begin{figure}
    \centering
    \includegraphics[width=\linewidth]{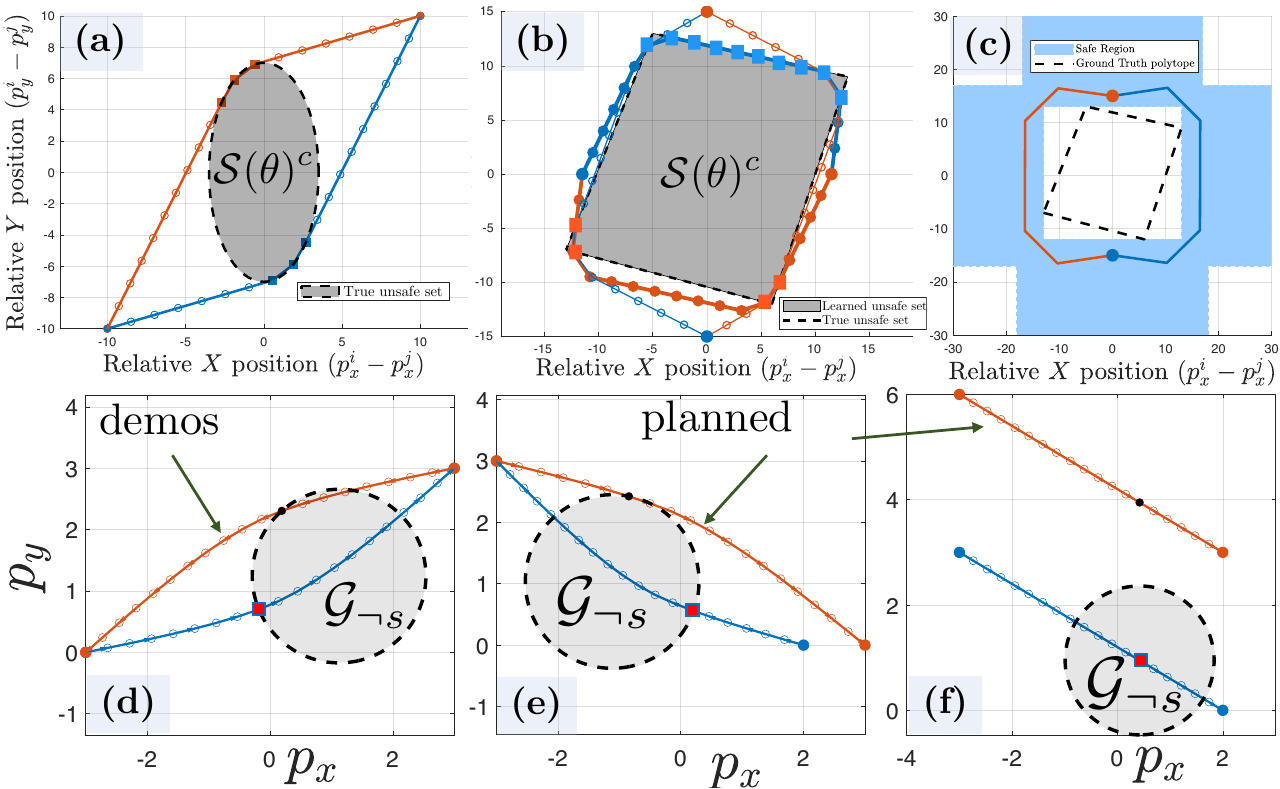}
    \caption{ 
    Learned constraint sets of Agent 1 (blue) and 2 (orange) with double integrator dynamics 
    and (a) ellipsoidal, (b) polytopic, or (d) velocity-dependent spherical collision avoidance constraints. (c) Inner approximation of the safe set (from (b)) via volume extraction, and corresponding safe motion plans. (e, f) Motion plans designed using learned constraints from (d).
    In all subplots, solid circles at the ends of the trajectories indicate start and goal positions. In (a) and (b), solid squares indicate tight (i.e., activated) constraints. In (d), (e), and (f), red (resp., black) squares indicate Agent 1 (resp., 2) states corresponding to the gray velocity-dependent spherical constraints depicted.
    \vspace{-15pt}
    }
    \label{fig:DI_poly_ell}
\end{figure}

To validate our methods, we perform constraint learning and volume extraction-based safe motion planning under elliptical and polytopic constraint parameterizations (Sec. \ref{subsec: Constraint Parameterizations and Agent Costs}), via simulations on double integrator (Sec. \ref{subsec: Double Integrator Simulations})
and quadcopter (Sec. \ref{subsec: Quadcopter Simulations}) dynamics,
and
hardware experiments on ground robots with unicycle dynamics (Sec. \ref{subsec: Hardware Experiments}). 
\revision{For each agent dynamics and constraint type, we solve \eqref{Eqn: Forward Game} using a given set of costs $J^i$ and ground truth constraints (via the KKT conditions \eqref{Eqn: KKT, Forward Game}) to generate demonstrations $\D$
that activate the unknown constraint.
We solve \eqref{Eqn: KKT, Inverse, Optimal, Objective} to learn constraints in simulation and \eqref{Eqn: KKT, Inverse, Suboptimal} to learn constraints from hardware demonstrations that may not satisfy the KKT conditions \eqref{Eqn: KKT, Forward Game} perfectly.
}
We also use the recovered constraints to generate safe motion plans; in contrast, a cost inference-based baseline method generates unsafe trajectories (Sec. \ref{subsec: Comparison Against Cost Inference Baseline}).
Our experiments are implemented with YALMIP \cite{Lofberg2004}, Gurobi \cite{gurobi}, CasADi \cite{Andersson2019CasADI}, and IPOPT \cite{wachter2006implementation}. For additional experiments and details, 
see App. \ref{subsec: App, Experiment Details}
\cite{ArXivPaper}.

\subsection{Constraint Parameterizations and Agent Costs}
\label{subsec: Constraint Parameterizations and Agent Costs}

In our experiments, we consider the following constraint parameterizations over agent states $\{x_t^i: i \in [N], t \in [T] \}$. We denote by $p_t^i$ and $v_t^i$ the components of $x_t^i$ corresponding to position and velocity, respectively, $\forall i \in [N], t \in [T]$
\footnote{We show in App. \ref{subsec: App, Experiment Details} 
\cite{ArXivPaper} 
that many dynamics models which do not explicitly encode the system's velocity in the state, e.g., 2D unicycles, can be transformed into equivalent dynamics models that do.
}.

\paragraph{Elliptical Constraints} We encode elliptical collision avoidance constraints for 2D systems via the parameterization $g_{t, \urk}^i(\xi, \theta^i) = - (p_t^i - p_t^j)^\top \text{diag}\{\theta_2^i, \theta_3^i\} (p_t^i - p_t^j) + (\theta_1^i)^2 \leq 0$ with $\theta_k^i > 0, \forall \ k \in [3]$, where $\text{diag}\{\cdot\}$ describes a diagonal matrix with given entries as diagonal values\footnote{For 3D systems, we replace $\text{diag}\{\theta_2^i, \theta_3^i \}$ with $\text{diag}\{\theta_2^i, \theta_3^i, \theta_4^i \}$ 
}. As a special case, this parameterization encodes spherical collision avoidance constraints when $\theta_2^i = \theta_3^i = 1, \forall \ i \in [N]$ are known. Moreover, by replacing $(p_t^i - p_t^j)$ with $(p_t^i - p_t^j, v_t^i - v_t^j)$ and increasing the dimension of 
$\text{diag}\{\theta_2^i, \theta_3^i\}$, we 
encode \textit{velocity-dependent} collision avoidance, 
which allows agents to modulate their collision avoidance radii depending on whether other agents are approaching or moving away.

\paragraph{Polytopic Constraints} We use intersections of unions of half-spaces to encode polytopic constraints for each agent $i \in [N]$ via the parameterization:
{
\setlength{\abovedisplayskip}{3pt}
\setlength{\belowdisplayskip}{3pt}
\begin{align*}
    C_\urk^{\ineq, i} &= \bigwedge_{t \in [T]} \bigwedge_{j \in [N] \backslash \{i\}} \\
    &\hspace{1cm} \bigwedge_{\alpha \in [\overline N_c]} \bigvee_{\beta \in [N_c]} \big\{ A_{\alpha, \beta}(\theta) (p_t^j - p_t^i) \leq b_{\alpha, \beta}(\theta) \big\}.
\end{align*}
}
\hspace{-1mm}where $A_{\alpha,  \beta}(\theta)$ and $b_{\alpha, \beta}(\theta)$ are components of $\theta$ across indices $\alpha, \beta$. As special cases, the above parameterization can encode polytopic collision avoidance and proximity constraints (which bound agents' relative positions). By replacing $A_{\alpha,  \beta}(\theta) (p_t^j - p_t^i)$ with $A_{\alpha, \beta}(\theta) (p_t^j - p_t^i) + \bar A_{\alpha, \beta}(\theta) (v_t^j - v_t^i)$ above, we can encode
line-of-sight constraints which compel agents to keep other agents in sight, an important requirement for pursuit-evasion or herding applications \cite{Zhou2016CooperativePursuitwithVoronoiPartitions}.

\paragraph{Origin and Goal Constraints}
In all experiments, the trajectory of each agent $i \in [N]$ is constrained by a prescribed set of origin ($\bar p_o^i$) and goal ($\bar p_d^i$) positions, enforced via the (known) constraint $h_t^i(\xi) := (p_0^i - \bar p_o^i, p_T^i - \bar p_d^i ) = 0$.

\paragraph{Agent Costs} 
Unless otherwise specified, 
each agent's cost is the individual smoothness cost $J^i := \sum_{t=1}^{T-1} \Vert p_{t+1}^i - p_t^i \Vert_2^2$, or shared smoothness cost $J^i := \sum_{t=1}^{T-1} \sum_{j=1}^N \Vert p_{t+1}^j - p_t^j \Vert_2^2$. 

\subsection{Double Integrator Simulations}
\label{subsec: Double Integrator Simulations}

\paragraph{Constraint Inference}


We validate our method by recovering
polytopic, elliptic, and velocity-dependent spherical collision avoidance constraints. 
Fig. \ref{fig:DI_poly_ell}a-\ref{fig:DI_poly_ell}b
plot demonstrations generated via solving \eqref{Eqn: Forward Game} across all $i \in [N]$, and mark timesteps when constraints are active. By solving \eqref{Eqn: KKT, Inverse, Optimal}, we were able to recover the true constraint parameter $\theta^\star$.

\paragraph{Motion Planning}
To evaluate our volume extraction approach for planning
(Sec. \ref{subsec: Volume Extraction Over Trajectories and Parameters for Motion Planning} and App. \ref{subsec: App, Robust Motion Planning via Implicit Constraint Checking})
\cite{ArXivPaper},
we generate safe trajectories which satisfy a given set of polytopic collision-avoidance constraints. Concretely, using a set of queried trajectories $\Q_\xi$, we extract hypercubes of safe trajectories which were then used for constraint checking during trajectory generation. Our method generates constraint-satisfying trajectories for each agent (Fig. \ref{fig:DI_poly_ell}d).
Moreover, Fig. \ref{fig:rebuttal___volume_extraction_robustness} in App. \ref{subsubsec: App, Double Integrator Simulations} 
\cite{ArXivPaper} 
presents 
MPPI-generated robust motion plans, informed by safe trajectories queried via volume extraction methods, that are safe with respect to an underlying constraint
that
is conservatively but not perfectly recovered. In contrast, naive motion plans designed using a point estimate of the constraint parameter are unsafe. 


\begin{figure}[ht]
    \centering
    \includegraphics[width=0.97\linewidth]{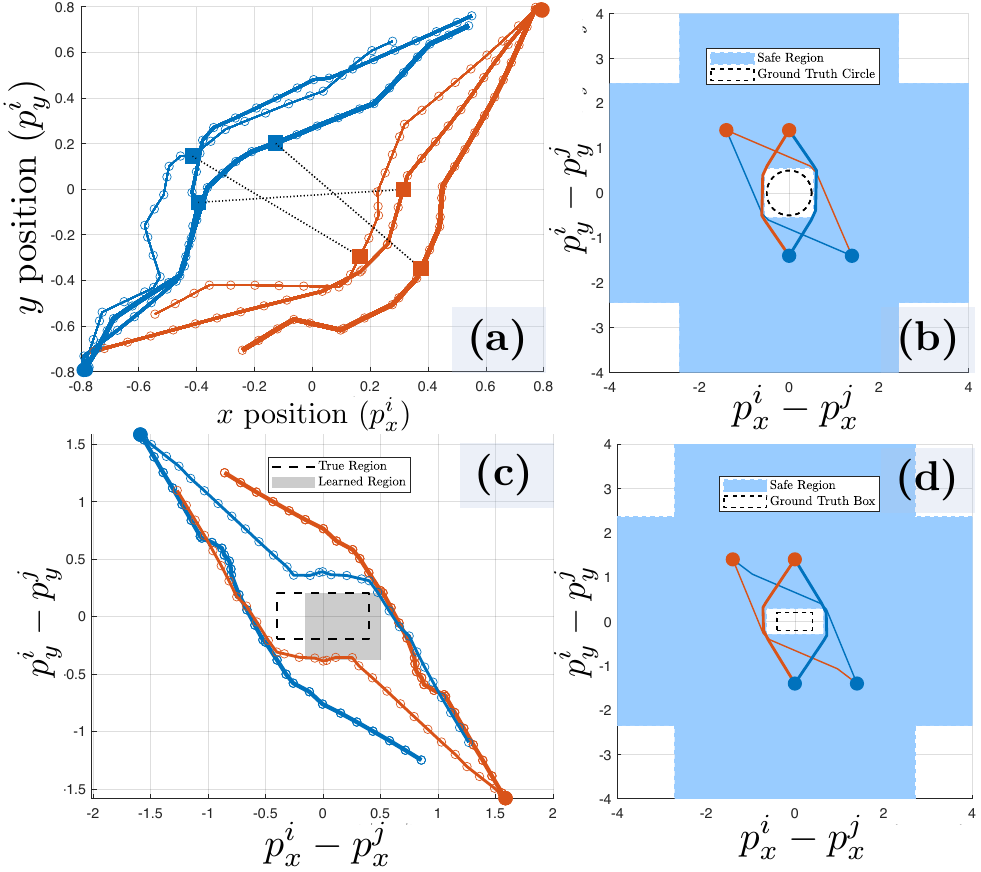}
    \vspace{-5pt}
    \caption{(a) Demonstrations (delinated by linewidth) and (b) safe planning via volume extraction for hardware unicycle agents 1 (blue) and 2 (orange) with spherical collision-avoidance constraints. 
    (c)-(d) present the box-constraint case; despite reduced recovery accuracy from demonstrator suboptimality, the volume-extraction planner still produces safe trajectories. 
    In (a), squares on and dashed lines between trajectories indicate constraint activation (i.e., tightness).
    \vspace{-8pt}
    }
    \label{fig:hardware}
\end{figure}

\paragraph{Compute Time vs. Number of Agents}

\revision{For the inference of spherical collision avoidance constraints between agents with double integrator dynamics, 
we record Gurobi solve times for problem instances with different numbers of agents $N$.
The $N = 2, 10, 20, 30, 100$ settings
respectively required 0.03 s, 0.16 s, 0.22 s, 2.87 s, and 6.24 s to solve,
indicating that our method remains tractable on large-scale problems.
Similarly, the $N = 2, 4, 10$ instances for box collision avoidance constraints required 0.06 s, 0.59 s, and 5.37 s to solve, respectively.
}
For details, see App. \ref{subsec: App, Experiment Details} \cite{ArXivPaper}.


\subsection{Unicycle Simulations}
\label{subsec: Unicycle Simulations}

\begin{figure}[ht]
    \centering
    \includegraphics[width=0.90\linewidth]{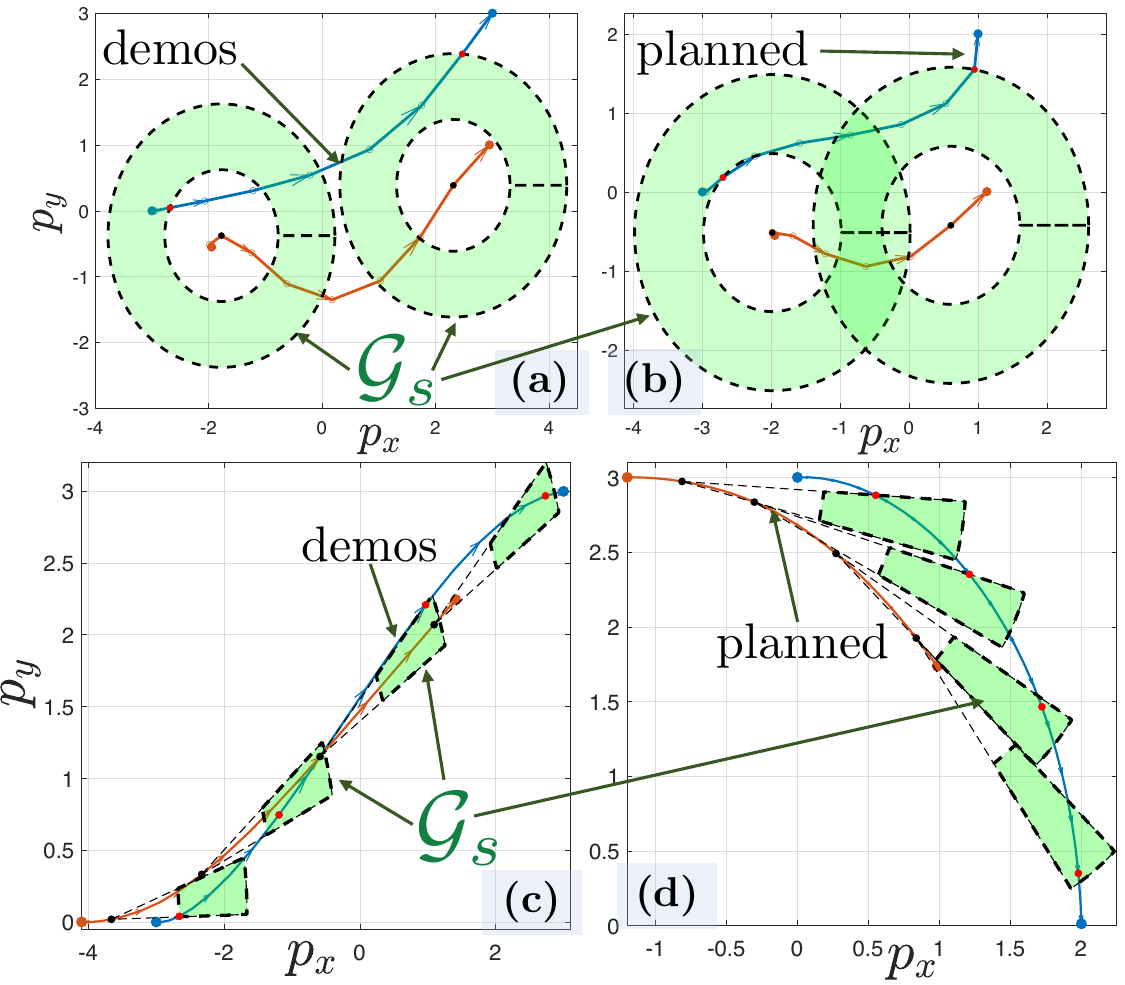}
    \caption{
    Constraint learning and planning for Agent 1 (blue) and 2 (orange) with unicycle dynamics satisfying (a, b) proximity or (c, d) line of sight. Learned constraints (shaded green) coincide with the true constraints (dashed black lines).
    }
    \vspace{-12pt}
    \label{fig:unicycle}
\end{figure}

In simulations of interacting agents with unicycle dynamics, our method recovers unknown line-of-sight and spherical proximity and collision avoidance constraints from local Nash interactions (Fig. \ref{fig:unicycle}a and \ref{fig:unicycle}c).
We then used volume extraction (Sec. \ref{subsec: Volume Extraction Over Trajectories and Parameters for Motion Planning}) to compute safe interactive motion plans (Fig. \ref{fig:unicycle}b and \ref{fig:unicycle}d). For details, see
App. \ref{subsubsec: App, Unicycle Dynamics Simulations}
\cite{ArXivPaper}.

\subsection{Quadcopter Simulations}
\label{subsec: Quadcopter Simulations}

\begin{figure}[ht]
    \centering
    \includegraphics[width=0.90\linewidth]{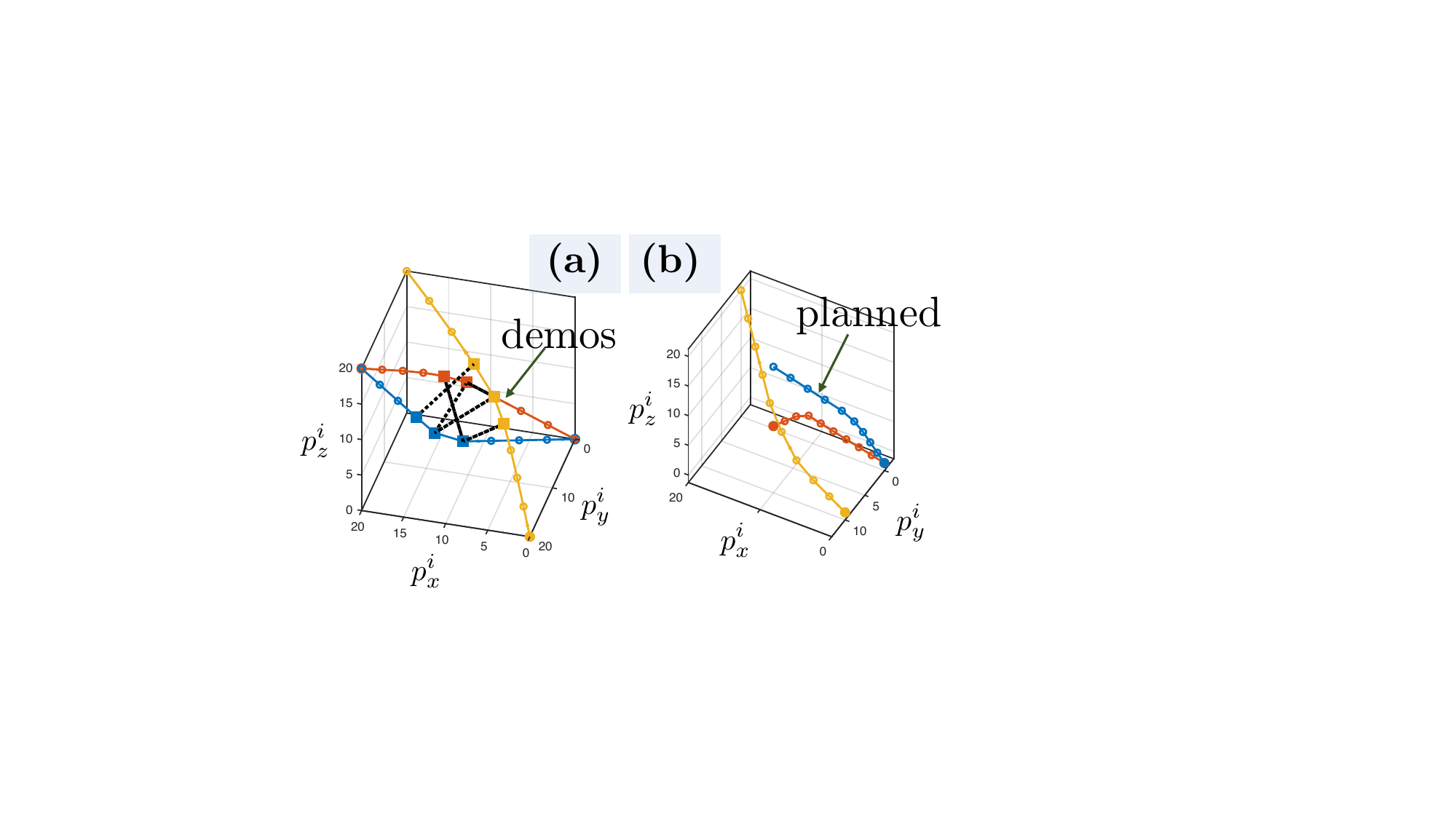}\vspace{-5pt}
    \caption{Constraint learning and planning for Agents 1 (blue), 2 (orange), and 3 (yellow) with quadcopter dynamics satisfying spherical collision-avoidance constraints.
    (a) A demonstration of all agents, in absolute coordinates, interacting while satisfying the constraints. Our method exactly learns 
    all
    constraint parameters.
    Filled squares on, and dashed lines between, trajectories indicate tight constraints.
    (b) Using our learned constraints, we generate safe motion plans via volume extraction over the trajectory space.
    }
    \label{fig:quad_sphere}
    \vspace{-10pt}
\end{figure}

We accurately recover 
spherical (Fig. \ref{fig:quad_sphere}) and 
box-shaped (App. \ref{subsubsec: App, Quadcopter Simulations}
\cite{ArXivPaper}, 
Fig. \ref{fig:quad_box}) collision avoidance constraints from interaction demonstrations at local Nash stationarity between 3 agents with 12-D quadcopter dynamics. We then compute safe motion plans via volume extraction.
In contrast, the baseline single-agent constraint learning method in \cite{Chou2020LearningConstraintsFromLocallyOptimalDemonstrationsUnderCostFunctionUncertainty}, which treats Agents 1 and 3 as unconstrained moving obstacles, failed to recover the constraint of Agent 2 and incorrectly perceived the provided demonstrations as suboptimal, with a stationarity error of 1.8.
For details, see App. \ref{subsubsec: App, Quadcopter Simulations} 
\cite{ArXivPaper}.



\subsection{Nonlinear Constraint Recovery under Cost Uncertainty}
\label{subsec: Single Integrator Simulations for Nonlinear Constraint Recovery under Cost Uncertainty}

Our method can learn constraints and generate safe plans even when the agents' objectives are also \textit{a priori} unknown, or when the constraint parameterization is nonlinear in agent states, leading to nonconvex unsafe sets. For illustration, we simulate two interacting agents with single integrator dynamics. 
Relevant figures are in App. \ref{subsubsec: App, Single Integrator Simulations for Nonlinear Constraint Recovery}
\cite{ArXivPaper}.

\subsection{Learning with Misspecified Constraint Parameterizations}
\label{subsec: Constraint Inference Despite Misspecified Constraint Parameterizations}

\begin{figure}[ht]
  \begin{minipage}[c]{0.53\linewidth}
    \includegraphics[width=0.99\linewidth]{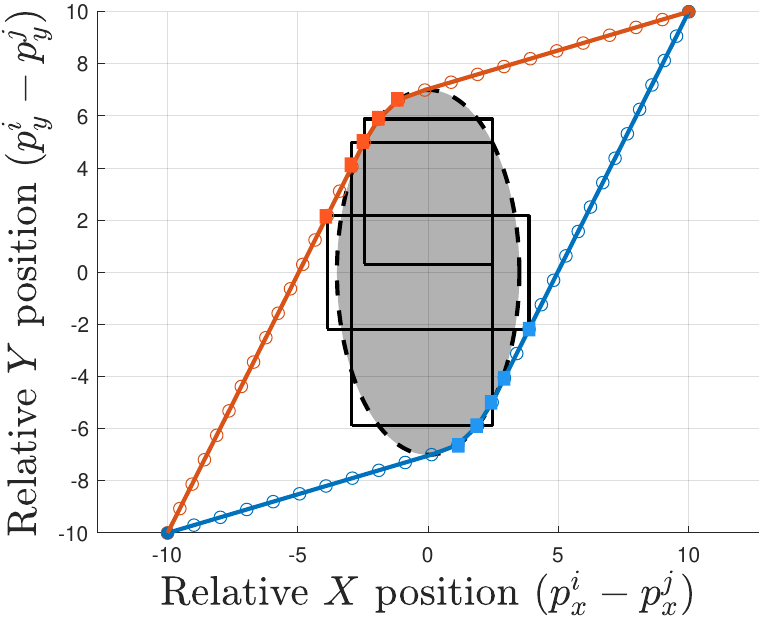}
  \end{minipage}\begin{minipage}[c]{0.47\linewidth}
    \caption{Our method well-approximates an ellipsoidal collision-avoidance set as the union of three boxes from interaction demonstrations with bicycle dynamics, even when constraints are mistakenly assumed to be box-parameterized.}\label{fig:boxes_for_ellipse}
  \end{minipage}
  \vspace{-20pt}
\end{figure}

We 
simulate a constraint learner who mistakenly believes that the ellipse-shaped collision-avoidance constraint underlying an interaction demonstration is a union of box-shaped constraints (Fig. \ref{fig:boxes_for_ellipse}). Despite this incorrect belief, 
our method still reasonably approximated the ellipse-shaped collision avoidance constraint as the union of three boxes with zero stationarity.
This result illustrates
that given a constraint parameterization, the number of constraint sets recovered can be flexibly adjusted to explain the demonstration's optimality.

\subsection{Hardware Experiments}
\label{subsec: Hardware Experiments}



We demonstrate constraint inference and safe motion planning on hardware experiments involving two agents with unicycle dynamics whose trajectories satisfy spherical or box-shaped collision-avoidance 
or line-of-sight (Fig. \ref{fig:hardware} and Fig. \ref{fig:front_figure} time-lapses) constraints. Despite suboptimality in the hardware demonstrations, our volume extraction-based motion planning method inner approximated the guaranteed safe set, and generated safe trajectories for both agents. 

\begin{figure}[!ht]
  \begin{minipage}[c]{0.53\linewidth}
    \includegraphics[width=0.99\linewidth]{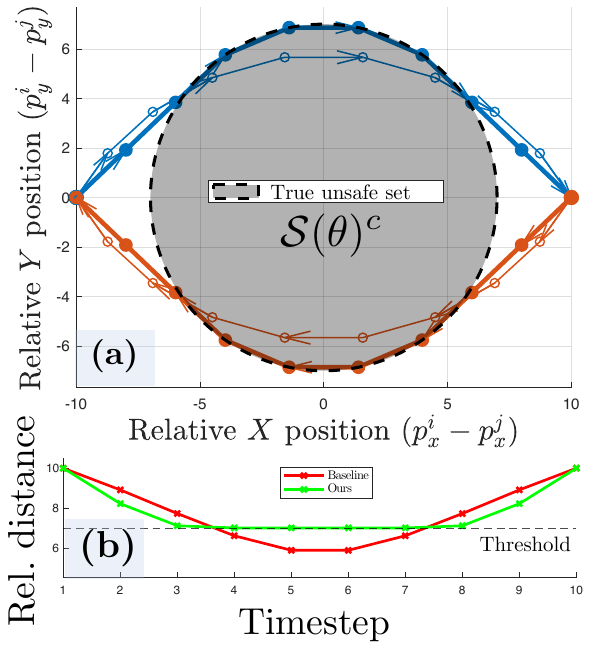}
  \end{minipage}\begin{minipage}[c]{0.47\linewidth}
    \caption{(a) Local Nash trajectories of Agents 1 (blue) and 2 (orange) in relative coordinates, using spherical constraints learned via our method (thick) and the baseline \cite{Peters2021InferringObjectives} (thin, Sec. \ref{subsec: Comparison Against Cost Inference Baseline}). Baseline trajectories overlap the circular unsafe set, with violations further shown in the inter-agent distance-time plot in (b).
    }\label{fig:baseline}
  \end{minipage}
  \vspace{-20pt}
\end{figure}


\subsection{Comparison Against Cost Inference Baseline}
\label{subsec: Comparison Against Cost Inference Baseline}



Finally, we compare the efficacy of our constraint learning-based method for safe motion planning against the cost inference-based approach in \cite{Peters2021InferringObjectives}. Specifically, we simulate interacting agents with double integrator dynamics who abide by \textit{a priori} unknown spherical collision avoidance constraints. Our method recovers the constraints, which are then used for motion planning via solving the KKT conditions \eqref{Eqn: KKT, Forward Game} for the forward game \eqref{Eqn: Forward Game}. Meanwhile, the approach in \cite{Peters2021InferringObjectives} assumes each agent $i$'s collision avoidance intent is embedded in its cost  $\tilde J^i(\tilde \theta)$ as a log barrier with an unknown weight $\tilde \theta$, i.e., $\tilde J^i(\tilde \theta) := J^i
- \tilde \theta \cdot \sum_{t=1}^T \log(\Vert p_t^i - p_t^{2-i} \Vert_2^2)$. Then, the method in \cite{Peters2021InferringObjectives} infers $\tilde \theta$ from data, and uses $\tilde J^i(\tilde \theta)$ to design motion plans by solving the forward dynamic game \eqref{Eqn: Forward Game} \textit{without explicit constraints.} Fig. \ref{fig:baseline} shows that, while our method generates safe plans, the approach in \cite{Peters2021InferringObjectives} produces constraint-violating plans.
Across 100 demonstrations with distinct initial and goal positions, our method required an average runtime of $99.4 \pm 11.2$ ms and always generated constraint-abiding interactions, while the method in \cite{Peters2021InferringObjectives} required an average runtime of $68.7 \pm 39.8$ ms\footnote{Since the approach in \cite{Peters2021InferringObjectives} can be posed as an optimization problem with no binary variable, it is, unsurprisingly, slightly faster than our method.} but generated constraint-violating plans 36\% of the time.

\vspace{-2pt}
\section{Conclusion and Future Work}
\label{sec: Conclusion and Future Work}

We presented an inverse dynamic games-based constraint learning framework that infers strategic agents' coupled constraints from demonstrations of their equilibrium interactions, and uses the inferred constraints for robust, safe motion planning.
In simulation and hardware experiments,
our method accurately infers agent constraints and generates robust motion plans,
even 
when using demonstrations away from local Nash equilibria.
Future work will extend our 
approach to infer temporally-extended interaction constraints
\cite{VazquezChanlatte2018LearningTaskSpecificationsfromDemonstrations},
and infer interaction constraints with unknown parameterization
using Gaussian processes
\cite{Chou2022GPConstraintLearning}.



\bibliography{references}
\bibliographystyle{IEEEtran}





\clearpage

\appendix

\revision{Below, App. \ref{subsec: App, Table of Notation} presents a table of commonly used notation in our work. 
App. \ref{subsec: App, MILPs for Inferring Polytopic Constraints} then illustrates how the constraint recovery problem \eqref{Eqn: KKT, Inverse, Optimal} can be recast as a mixed-integer linear program (MILP)  for offset-parameterized constraints.
Next, App. \ref{subsec: App, Theoretical Analysis and Proofs} presents proofs of theoretical results presented in the main paper and establishes additional theoretical results beyond those given in the main paper.
Then, App. \ref{subsec: App, Robust Motion Planning via Implicit Constraint Checking} describes the implicit constraint checking-based safe trajectory design first described at the end of Sec. \ref{subsec: Volume Extraction Over Trajectories and Parameters for Motion Planning}.
Finally, App. \ref{subsec: App, Experiment Details} provide additional details for the experiments whose results are described in Sec. \ref{sec: Experiments}.
}

\subsection{Table of Notation}
\label{subsec: App, Table of Notation}

\begin{table}[ht]
\caption{Notation commonly used throughout the work.}
\renewcommand{\arraystretch}{1.15}
\footnotesize
\centering

\begin{tabularx}{\columnwidth}{@{}lX@{}}
\toprule
\textbf{Symbol} & \textbf{Definition} \\
\midrule

$N$ & Number of agents \\
$T$  & Time horizon length \\
$x_t^i$ & State of agent $i \in [N]$ at time $t \in [T]$ \\
$u_t^i$ & Control of agent $i \in [N]$ at time $t \in [T]$ \\

$x_t$ & State at time $t \in [T]$, i.e., $x_t := (x_t^1, \cdots, x_t^N)$
\\
$u_t$ & Control at time $t \in [T]$, i.e., $u_t := (u_t^1, \cdots, u_t^N)$
\\

$x$ & System-level state trajectory, i.e., $x := (x_1, \cdots, x_T)$  \\
$u$ & System-level control trajectory, i.e., $u := (u_1, \cdots, u_T)$ \\

$J^i(\xi)$ & Cost function of agent $i \in [N]$ \\

$h_\beta^i(\xi)=0$ & $\beta$-th ($\beta\in[N^{\eq,i}]$) equality constraint of agent $i \in [N]$ across all times \\
$g_{\beta,k}^i(\xi)\le0$ & $\beta$-th ($\beta\in[N_k^{\ineq,i}]$) known inequality constraint of agent $i \in [N]$ across all times\\
$g_{\beta,\urk}^i(\xi,\theta)\le0$ & $\beta$-th ($\beta\in[N_\urk^{\ineq,i}]$) unknown inequality constraint of agent $i \in [N]$ across all times \\

$\theta^\star$ & True unknown constraint parameter \\
$\Theta$ & Parameter space, i.e., the set of all possible parameter values (Note: $\theta^\star \in \Theta$) \\

$\boldh^i,\; \boldg_k^i,\; \boldg_\urk^i$ & Stacked equality, known inequality, and unknown inequality constraints for agent $i \in [N]$ \\
$\boldh,\; \boldg_k,\; \boldg_\urk$ & Stacked equality, known inequality, and unknown inequality constraints across all agents \\

$\Safe(\theta)$ & Safe set, i.e., the set of all safe trajectories $\{\xi:\boldg_\urk(\xi,\theta)\le0\}$ \\
$\Avoid(\theta)$ & Unsafe set, given by $\Avoid(\theta) = \Safe(\theta)^c$ \\

$\mathcal{D}=\{\xi_d^{\loc}\}_{d=1}^D$ & Dataset of locally Nash-stationary demonstrations \\

$\boldlambda_{d,k}^i$ & Lagrange multipliers for known inequalities for agent $i \in [N]$ in demonstration $d \in [D]$ \\
$\boldlambda_{d,\urk}^i$ & Lagrange multipliers for unknown inequalities for agent $i \in [N]$ in demonstration $d \in [D]$ \\
$\boldnu_d^i$ & Lagrange multipliers for equalities for agent $i \in [N]$ w.r.t. the demonstration $d \in [D]$ \\

$\boldlambda_k,\;\boldlambda_\urk,\;\boldnu$ & 
$\boldlambda_{d,k}^i$, $\boldlambda_{d,\urk}^i$, and $\boldnu_d^i$ stacked across all agents and demonstrations, respectively
\\

$\nabla_x \mathbf{f}$ & Jacobian of $\mathbf{f}$ \\

$\stat^i(\xi_d^{\loc})$ & KKT stationarity residual term corresponding to agent $i \in [N]$ and demonstration $d \in [D]$ \\
$\KKT^i(\xi_d^{\loc})$ & KKT-feasible set of parameters $\theta$ and Lagrange multipliers $(\boldlambda_k, \boldlambda_\urk, \boldnu)$ w.r.t. agent $i \in [N]$ and the demonstration $\xi_d^{\loc}$ \\
$\KKT(\mathcal{D})$ & 
Intersection of $\KKT^i(\xi_d^{\loc})$ across all agents and demonstrations in $\D$
\\

$\mathcal{F}(\mathcal{D})$ & Parameters in $\Theta$ consistent with demonstrations in $\D$\\
$\mathcal{G}_s(\mathcal{D})$ & Guaranteed safe trajectories w.r.t. all $\theta \in \mathcal{F}(\mathcal{D})$ \\
$\mathcal{G}_{\urs}(\mathcal{D})$ & Guaranteed unsafe trajectories w.r.t. all $\theta \in \mathcal{F}(\mathcal{D})$\\

\bottomrule
\end{tabularx}
\end{table}

\subsection{MILPs for Inferring Polytopic Constraints}
\label{subsec: App, MILPs for Inferring Polytopic Constraints}

In this subsection, we elaborate on how the constraint recovery problem \eqref{Eqn: KKT, Inverse, Optimal} can be formulated as a mixed-integer linear program (MILP) \ref{subsubsec: App, Reformulation of the Constraint Inference Problem for Unions of Offset-Parameterized Constraints} for offset-parameterized constraints. We also discuss how \eqref{Eqn: KKT, Inverse, Optimal} can be formulated as a mixed-integer bilinear program (MIBLP) in \ref{subsubsec: App, Reformulation of the Constraint Inference Problem for Unions of Affine-Parameterized Constraints} for affine-parameterized constraints.

In Sections \ref{subsubsec: App, Reformulation of the Constraint Inference Problem for Unions of Offset-Parameterized Constraints} and \ref{subsubsec: App, Reformulation of the Constraint Inference Problem for Unions of Affine-Parameterized Constraints} below, we use the following notation: Given any trajectory $\xi \in \R^{(n+m)T}$, let $x_t^i(\xi) \in \R^{n_i}$ denote the value of agent $i$'s state at time $t$, as prescribed by the trajectory $\xi$. We also define $u_t^i(\xi) \in \R^{m_i}$, $x_t(\xi) \in \R^n$, and $u_t(\xi) \in \R^m$ similarly.

\subsubsection{Reformulation of the Constraint Inference Problem for Unions of Offset-Parameterized Constraints}
\label{subsubsec: App, Reformulation of the Constraint Inference Problem for Unions of Offset-Parameterized Constraints}
Below, we present a more detailed explanation of the fact, first introduced in Sec. \ref{subsec: MILP Reformulation for Unions of Offset-Parameterized Constraints}, the problem of learning \textit{offset-parameterized interaction constraints} in the context of dynamic games can be reformulated as MILPs. We note that offset-parameterized interaction constraints often arise in dynamic games, e.g., in the context of spherical or box-shaped collision-avoidance constraints with uncertain dimensions. In particular, consider the specific setting in which the states of each agent have equal dimension (i.e., $n_i = n_j$ for each $i, j \in [N]$). 
the unknown constraints $\boldg_\urk^i(\xi, \theta) \leq 0$ encode that for all $t \in [T]$, the state of each other agent $j \in [N] \backslash \{i\}$ relative to agent $i$ lies outside a polytopic collision avoidance set with $N_c$ sides, each of which can be represented as a scalar constraint.
Concretely, feasible system-level trajectories must lie outside the avoid sets:
\begin{align} \label{App, Eqn: A ij, Offset-Parameterized}
    \Avoid^{i,j,t}(\theta) := \bigcap_{\beta=1}^{N_c} \{ \xi \in \R^{(n+m)T}: \bolda_\beta^{i,j,t\top} \xi < b_\beta^i(\theta) \}
\end{align}
for all $i, j \in [N]$, $j \ne i$ and $t \in [T]$, where, for each $\beta \in [N_c]$, the vector $\bolda_\beta^{i,j,t} \in \R^{(n+m)T}$ and the scalar $b_\beta^i(\theta) \in \R$ for each $i, j \in [N]$, $j \ne i$ $t \in [T]$ together represent one of the $N_c$ sides of the polytope-shaped collision-avoidance constraint set. Moreover, each $b_\beta^i(\theta) \in \R$ is specified by an unknown parameter $\theta$.
Thus, the trajectory $\xi$ must evade the avoid set $\Avoid(\theta) \subset \R^{(n+m)T}$, given by:
\begin{align} \label{App, Eqn: A(theta), Offset-Parameterized}
    \Avoid(\theta) := 
    \bigcup_{i, j \in [N], j \ne i, t \in [T]} \Avoid^{i,j,t}(\theta),
\end{align}
or equivalently, stay within the safe set $\Safe(\theta) := (\Avoid(\theta))^c \subset \R^{(n+m)T}$ given by:
\begin{align} \nonumber
    \Safe(\theta) &:= 
    \bigcap_{i, j \in [N], j \ne i, t \in [T]} \bigcup_{\beta \in [N_c]} \big\{ 
    \xi: 
    \bolda_\beta^{i,j,t\top} \xi \geq b_\beta^i(\theta) \big\}
\end{align}

We now present a transformation of the primal feasibility conditions \eqref{Eqn: KKT, Forward, Primal Feasibility}, complementary slackness conditions \eqref{Eqn: KKT, Forward, Complementary Slackness}-\eqref{Eqn: KKT, Forward, Complementary Slackness}, and stationarity conditions \eqref{Eqn: KKT, Forward, Stationarity} that allow us to reformulate the KKT conditions \eqref{Eqn: KKT, Inverse, Optimal} into an MILP. Our method extends the approach presented in Sections IV-B of \cite{Chou2020LearningConstraintsFromLocallyOptimalDemonstrationsUnderCostFunctionUncertainty} to accommodate the multi-agent constraint learning problem presented in this work.

\begin{enumerate}
    \item \textbf{Reformulating the primal feasibility conditions \eqref{Eqn: KKT, Forward, Primal Feasibility}}:

    $\hspace{3mm}$ First, we rewrite the primal feasibility conditions \eqref{Eqn: KKT, Forward, Primal Feasibility} via the big-M formulation \cite{Bertsimas1998IntroductiontoLinearOptimization, Chou2020LearningConstraintsFromLocallyOptimalDemonstrationsUnderCostFunctionUncertainty} by introducing binary vectors $\boldz_d^{i,j,t} = (z_{d,\beta}^{i,j,t} \in \{0, 1\}) \in \{0, 1\}^{N_c}$ as shown below---$\forall d \in [D]$, $i, j \in [N]$, $t \in [T]$:
    \begin{subequations} \label{App, Eqn: Primal Feasibility, Reformulation}
    \begin{align} 
        &(\boldA^{i,j,t})^\top \xi_d^\loc \geq \boldb^i(\theta) - M(\textbf{1}_{N_c} - \boldz_d^{i,j,t}), \\ \nonumber
        &\sum\nolimits_{\beta = 1}^{N_c} z_{d,\beta}^{i,j,t} \geq 1, 
    \end{align}
    \end{subequations}
    Above, $M \gg 0$, 
    $\textbf{1}_{N_c} := (1, \cdots, 1) \in \R^{N_c}$,
    while $\boldA^{i,j,t}$ $\in \R^{(n+m)T \times N_c}$ and $\boldb^i(\theta) \in \R^{N_c}$ are the concatenations of $\bolda_\beta^{i,j,t}$ and $b_\beta^i(\theta)$, respectively, across all $\beta \in [N_c]$.

    \item \textbf{Reformulating the complementary slackness conditions \eqref{Eqn: KKT, Forward, Complementary Slackness}}:

    $\hspace{5mm}$ Let $\boldlambda_{d, \urk}^{i,j,t} = (\lambda_{d,\beta, \urk}^{i,j,t} \geq 0: \beta \in [N_c]) \in \R^{N_c}$ denote the dual variables associated with the scalar inequality constraints $(\bolda_\beta^{i,j,t})^\top \xi \geq b_\beta^i(\theta)$, across $\beta \in [N_c]$.    
    Since the safe set definition \eqref{Eqn: Safe Set, 2} involves a union over $\beta \in [N_c]$ for each $i, j \in [N]$, $j \ne i$, $t \in [T]$, the constraint $\{ \xi \in \R^{(n+m)T}: (\bolda_\beta^{i,j,t})^\top \xi_d^\loc \geq b_\beta^i(\theta) \}$ might only be enforced for a strict (but non-empty) subset of the indices $\beta \in [N_c]$, which we denote by $B^{i,j,t} \subseteq [N_c]$. Likewise, complementary slackness \eqref{Eqn: KKT, Forward, Complementary Slackness} should only be enforced for indices $\beta \in B^{i,j,t}$, for each $i, j \in [N], j \ne i$, $t \in [T]$, and in turn, terms of the form $\lambda_{d, \beta, \urk}^{i,j,t} \nabla_{\xi^i} g_{\beta, \urk} (\xi, \theta)$ are only included in the stationarity condition \eqref{Eqn: KKT, Forward, Stationarity} if $\beta \in B^{i,j,t}$. Thus, we use a big-M formulation with binary vectors 
    $\hat \boldz_{d,1}^{i,j,t} = (\hat z_{d,1, \beta}^{i,j,t}: \beta \in [N_c]) \in \{0, 1\}^{N_c}$ and $\hat \boldz_{d,2}^{i,j,t} = (\hat z_{d,2,\beta}^{i,j,t}: \beta \in [N_c]) \in \{0, 1\}^{N_c}$
    to encode 
    \eqref{Eqn: KKT, Forward, Complementary Slackness},
    and the
    binary vectors
    $\boldq_d^{i,j,t} = (q_{d,\beta}^{i,j,t}: \beta \in [N_c]) \in \{0, 1\}^{N_c}$ 
    to encode enforcement of 
    \eqref{Eqn: KKT, Forward, Complementary Slackness}. Then, we rewrite the complementary slackness conditions \eqref{Eqn: KKT, Forward, Complementary Slackness} as follows---$\forall d \in [D]$, $i, j \in [N], i \ne j$, $t \in [T]$:    
    \begin{subequations} \label{App, Eqn: Complementary Slackness, Reformulation}
    \begin{align}
        &\begin{bmatrix}
            \boldlambda_{d,\urk}^{i,j,t} \\
            \boldA^{i,j,t\top} \xi_d^\loc - \boldb^i(\theta)
        \end{bmatrix} 
        \leq M \begin{bmatrix}
            \hat \boldz_{d,1}^{i,j,t} \\ \hat \boldz_{d,2}^{i,j,t}
        \end{bmatrix}, \\
        &\hat z_{d,1, \beta}^{i,j,t} + \hat z_{d,2, \beta}^{i,j,t} \leq 2 - q_{d,\beta}^{i,j,t}, \ \ \forall \ \beta \in [N_c], \\ \nonumber 
        &\sum\nolimits_{\beta' = 1}^{N_c} q_{d, \beta'}^{i,j,t} \geq 1.
    \end{align}
    \end{subequations}
    
    \item \textbf{Reformulating the stationarity condition \eqref{Eqn: KKT, Forward, Stationarity}}:

    $\hspace{3mm}$ As noted above, the stationarity condition \eqref{Eqn: KKT, Forward, Stationarity} includes the term $\lambda_{d, \beta, \urk}^{i,j,t} \nabla_{\xi^i} g_{\beta, \urk} (\xi, \theta)$ \textit{only if} $\beta \in B^{i,j,t}$. To enforce this condition, for each $d \in [D]$, $t \in [T]$, $i, j \in [N], j \ne i$, we define $L_d^{i,j,t} \in$ $\R^{N_c \times (n_i + m_i)T}$ as the matrix whose $\beta$-th row equals:
    \begin{align} \nonumber
        \lambda_{d, \beta, \urk}^{i,j,t} \nabla_{\xi^i} (b_\beta^i(\theta) - \bolda_\beta^{i,j,t\top} \xi)^\top |_{\xi = \xi_d^\loc}.
    \end{align}
    We then rewrite the term ${\boldlambda_{d, \urk}^i}^\top \nabla_{\xi^i} \boldg_\urk^i(\xi, \theta)$ in \eqref{Eqn: KKT, Forward, Stationarity} as follows---$\forall i \in [N], d \in [D], t \in [T]$:
    \begin{align} \nonumber
        &\sum_{j \in [N]}\sum_{t \in [T]} \boldlambda_{d, \urk}^{i,j,t\top} \nabla_{\xi^i} \big( b^i(\theta) - \boldA^{i,j,t\top} \xi \big) |_{\xi = \xi_d^\loc} 
        \\ \nonumber
        = \ &\sum_{j \in [N]}\sum_{t \in [T]} \boldq_d^{i,j,t\top} L_d^{i,j,t}.
    \end{align}
    By replacing the bilinear terms $\boldq_d^{i,j,t\top} L_d^{i,j,t}$ in \eqref{Eqn: L, Bilinear Terms} with slack variables and adding constraints, we can reformulate \eqref{Eqn: L, Bilinear Terms} in a linear manner as well.   
    Specifically, we define $\tilde \boldq_d^{i,j,t} := 1 - \boldq_d^{i,j,t}$ for each $d \in [D]$, $i, j \in [N]$, $j \ne i$, $t \in [T]$, and we introduce slack variables $R_d^{i,j,t} \in \R^{N_c \times (n_i + m_i)}$, for each $d \in [D]$, $i,j \in [N]$, $j \ne i$, $t \in [T]$.
    Let $R_{d,\beta,\ell}^{i,j,t}$ and $L_{d,\beta,\ell}^{i,j,t}$ respectively denote the $\beta$-th row and $\ell$-th column entry of $R_d^{i,j,t}$ and $L_d^{i,j,t}$, for each $\beta \in [N_c], \ell \in [n]$.
    We assume that the slack variables $L_{d,\beta,\ell}^{i,j,t}$ are uniformly bounded below and above by $\underM$ and $\overM$, respectively, across all indices $i, j \in [N]$, $i \ne j$, $t \in [T]$, $d \in [D]$, $\beta \in [N_c]$, $\ell \in [n_i + m_i]$.
    We then introduce the additional constraints below, for each $\beta \in [N_c]$ and $\ell \in [n]$, to linearize the bilinear terms of the form $\boldq_d^{i,j,t\top} L_d^{i,j,t}$:
    \begin{subequations} \label{App, Eqn: Stationarity Term, Reformulation via Linearization, with q}
    {
    \begin{align}
        \min\{0, \underM \} &\leq R_{d,\beta,\ell}^{i,j,t} \leq \overM, \\
        \underM q_{d,\beta}^{i,j,t} &\leq R_{d,\beta,\ell}^{i,j,t} \leq \overM q_{d,\beta}^{i,j,t}, \\
        L_{d,\beta,\ell}^{i,j,t} - \tilde q_{d,\beta}^{i,j,t} \overM &\leq R_{d,\beta,\ell}^{i,j,t} \\ \nonumber
        &\leq L_{d,\beta,\ell}^{i,j,t} - \tilde q_{d,\beta}^{i,j,t} \underM \\
        R_{d,\beta,\ell}^{i,j,t} &\leq L_{d,\beta,\ell}^{i,j,t} + \tilde q_{d,\beta}^{i,j,t} \overM
    \end{align}
    }
    \end{subequations}
\end{enumerate}

We now use \eqref{App, Eqn: Primal Feasibility, Reformulation}, \eqref{App, Eqn: Complementary Slackness, Reformulation}, and \eqref{App, Eqn: Stationarity Term, Reformulation via Linearization, with q} to reformulate the constraint recovery problem \eqref{Eqn: KKT, Inverse, Optimal} as follows:
\begin{subequations} 
\label{App, Eqn: KKT, Inverse, for Unions, Offset-Parameterized}
\begin{align} 
\label{App, Eqn: KKT, Inverse, for Unions, Offset-Parameterized, Objective}
    \text{find} \hspace{5mm} &\theta, \boldlambda_{d,k}^{i,j,t}, \boldlambda_{d,\urk}^{i,j,t}, \boldnu_d^i, R_d^{i,j,t}, L_d^{i,j,t}, \boldq_d^{i,j,t}, \tilde \boldq_d^{i,j,t}, \\ \nonumber
    &\hspace{5mm} \boldz_{d,1}^{i,j,t}, \boldz_{d,2}^{i,j,t}, \hspace{5mm} \forall \ d, i, j, t \\ \label{App, Eqn: KKT, Inverse, for Unions, Offset-Parameterized, Primal Feasibility}
    \text{s.t.} \hspace{5mm} &\Eqn \eqref{App, Eqn: Primal Feasibility, Reformulation}, \hspace{5mm} \forall \ d \in [D], i, j \in [N] \\ \nonumber
    &\hspace{3cm} \st j \ne i, t \in [T], \\ 
    \label{App, Eqn: KKT, Inverse, for Unions, Offset-Parameterized, Lagrange Multiplier Non-negativity}
    \hspace{5mm} &\boldlambda_{d,k}^{i,j,t}, \boldlambda_{d,\urk}^{i,j,t} \geq 0, \hspace{5mm} \forall \ d \in [D], i, j \in [N] \\ \nonumber
    &\hspace{3cm} \st j \ne i, t \in [T], \\ 
    \label{App, Eqn: KKT, Inverse, for Unions, Offset-Parameterized, Complementary Slackness, 1}
    \hspace{5mm} &\Eqn \eqref{App, Eqn: Complementary Slackness, Reformulation}, \hspace{5mm} \forall \ d \in [D], i, j \in [N] \\ \nonumber
    &\hspace{3cm} \st j \ne i, t \in [T], \\ 
    \label{App, Eqn: KKT, Inverse, for Unions, Offset-Parameterized, Complementary Slackness, 2}
    \hspace{5mm} &\nabla_{\xi^i} J^i(\xi_d^\loc) + (\boldlambda_{d, k}^i)^\top \nabla_{\xi^i} \boldg_k^i(\xi_d^\loc) \\ \nonumber
    &\hspace{5mm} + \sum_{j \in [N]}\sum_{t \in [T]} \boldsymbol{1}_{N_c}^\top R_d^{i,j,t} \\
    \nonumber
    &\hspace{1cm} + (\boldnu_d^i)^\top \nabla_{\xi^i} \boldh^i(\xi_d^\loc) = 0, \\ \nonumber
    &\hspace{5mm} \forall \ d \in [D], i, j \in [N] \st j \ne i, t \in [T], \\ 
    \label{App, Eqn: KKT, Inverse, for Unions, Offset-Parameterized, Stationarity}
    &\Eqn \eqref{App, Eqn: Stationarity Term, Reformulation via Linearization, with q}, \hspace{5mm} \forall \ d \in [D], \beta \in [N_c], i, j \in [N] \\ \nonumber
    &\hspace{3cm} \st j \ne i, t \in [T],
\end{align}
\end{subequations}
where \eqref{App, Eqn: KKT, Inverse, for Unions, Offset-Parameterized, Primal Feasibility} encodes primal feasibility, \eqref{App, Eqn: KKT, Inverse, for Unions, Offset-Parameterized, Lagrange Multiplier Non-negativity} encodes the non-negativity of all Lagrange multiplier terms associated with inequality constraints, \eqref{App, Eqn: KKT, Inverse, for Unions, Offset-Parameterized, Complementary Slackness, 1}-\eqref{App, Eqn: KKT, Inverse, for Unions, Offset-Parameterized, Complementary Slackness, 2} encode complementary slackness, and \eqref{App, Eqn: KKT, Inverse, for Unions, Offset-Parameterized, Stationarity} encodes stationarity. We observe that \eqref{App, Eqn: KKT, Inverse, for Unions, Offset-Parameterized} is indeed an MILP.

\subsubsection{Reformulation of the Constraint Inference Problem for Unions of Affine-Parameterized Constraints}
\label{subsubsec: App, Reformulation of the Constraint Inference Problem for Unions of Affine-Parameterized Constraints}

Below, we generalize the constraints \eqref{App, Eqn: A ij, Offset-Parameterized} and \eqref{App, Eqn: A(theta), Offset-Parameterized} to accommodate the setting of \textit{affine-parameterized} constraints, i.e., the setting in which the vectors $\bolda_\beta^{i,j,t} \in \R^n$, which are multiplied to the trajectories $\xi \in \R^{(n+m)T}$, are also \textit{a priori} unknown and parameterized by the (unknown) constraint parameter $\theta$. Concretely, we assume that each agent $i \in [N]$ wishes to stay outside the following avoid sets, for each $j \in [N] \backslash \{i\}$, $t \in [T]$:
\begin{align} \label{Eqn: A ij, Affine-Parameterized}
    \Avoid^{i,j,t}(\theta^\star) &:= \bigcap_{\beta = 1}^{N_c} \big\{ \xi \in \R^{(n+m)T}: \bolda_\beta^{i,j,t}(\theta^\star)^\top x < b_\beta^i(\theta^\star) \big\}.
\end{align}
In other words, across all agents $i \in [N]$, the constraints $\boldg_\urk^i(\xi, \theta^\star) \leq 0$ together enforce that the trajectory $\xi$ must avoid the following unsafe set:
\begin{align} \label{Eqn: A(theta), Affine-Parameterized}
    \Avoid(\theta^\star) &:= \bigcup_{i,j \in [N], j \ne i, t \in [T]} \Avoid^{i,j,t}(\theta^\star).
\end{align}
If $N_c > 1$, one can rewrite the multiple constraints over which we are taking the intersection with a single constraint using the max operator. Thus, for ease of exposition, we will assume below that $N_c = 1$, and thus drop the indices $\beta$ in the terms $a_\beta^{i,j,t}(\theta)$ and $b_\beta^i$, i.e.,:
\begin{align} \label{Eqn: A(theta), Affine-Parameterized, no beta}
    \Avoid(\theta) &= \bigcup_{i,j \in [N], j \ne i, t \in [T]} \big\{ x \in \R^n: a_\beta^{i,j,t}(\theta)^\top x < b_\beta^i(\theta) \big\}.
\end{align}
We begin by first analyzing the hypothetical setting where the avoid set consists of a single affine-parameterized scalar constraint, i.e.,
\begin{align} \label{Eqn: A(theta), Affine-Parameterized, scalar}
    \Avoid(\theta) &= \big\{ x \in \R^n: a(\theta)^\top x < b(\theta) \big\},
\end{align}
before restoring the unions over indices $t \in [T]$, and $i, j \in [N] \st j \ne i$ in \eqref{Eqn: A(theta), Affine-Parameterized, no beta}. In short, we temporarily ignore the variables $i, j \in [N]$ and $t \in [T]$. Note that, as a byproduct of the above simplification, all binary variables $q$ introduced in \eqref{App, Eqn: Stationarity Term, Reformulation via Linearization, with q} (for linearization purposes) can be set to 1, and hence are dropped from our notation throughout our discussion of the scalar constraint setting.

Unlike the setting of offset-parameterized avoid sets (Section \ref{subsubsec: App, Reformulation of the Constraint Inference Problem for Unions of Offset-Parameterized Constraints}), the stationarity conditions formulated using the avoid set \eqref{Eqn: A(theta), Affine-Parameterized, scalar} will exhibit bilinearity, since they will contain terms of the form $\lambda_{d, \urk} \nabla_{\xi^i} g_\urk(\xi, \theta)$, which in turn are products of the primal variable $\theta$ and the dual variable $\lambda$ (corresponding to the inequality constraints).
Thus, a mixed-integer \textit{bilinear} program (MIBLP) must be solved to infer $\theta$, since products of continuous variables cannot be linearized exactly. However, we can still relax the resulting MIBLP into an MILP. Concretely, we introduce a new vector variable $\boldell := (\ell_d \in \R: d \in [D]) \in \R^D$, where each $\ell_d$ represents the term $\lambda_{d, \urk} \nabla_{\xi^i} g_\urk(\xi, \theta)$, and a new vector variable $\boldz_1 := (\boldz_{d,1} \in \R: d \in [D]) \in \R^D$, which (as in the offset-parameterized case) forces $\lambda_{d, \urk}$ to be 0 when $z_{d,1} = 0$. We then replace each bilinear term of the form $\lambda_{d, \urk} \nabla_{\xi^i} g_\urk(\xi, \theta)$ with the bilinear term $\ell_d z_{d,1}$.
Note that the relaxation gap introduced by the above process is non-zero only when the Lagrange multipliers $\lambda_{d, \urk}$ (and hence the binary variables $z_{d,1}$) are non-zero, i.e., when the interaction demonstration $\xi_d^\loc$ is active; otherwise, the bilinear term $\lambda_{d, \urk} \nabla_{\xi^i} g_\urk(\xi, \theta)$ vanishes from the stationarity conditions.

Effectively, the relaxation process introduced above replaces a product of two continuous variables, i.e., $\lambda_{d, \urk} \nabla_{\xi^i} g_\urk(\xi, \theta)$, with a product of a continuous variable and a binary variable, i.e., $\ell_d z_{d,1}$, with nonzero relaxation gap only when the associated constraints are active. We can now proceed to linearize $\ell_d z_{d,1}$ in the manner described in Section \ref{subsubsec: App, Reformulation of the Constraint Inference Problem for Unions of Offset-Parameterized Constraints}. Specifically, we introduce slack variables $\boldr := (r_d \in \R: d \in [D]) \in \R^D$ and constrain them using an appropriate analog of \eqref{App, Eqn: Stationarity Term, Reformulation via Linearization, with q}, for some positive integers $M, \underM, \overM > 0$. 
In summary,
by concatenating $r_d$, $z_{d,1}$, and $\ell_d$ horizontally across indices $d \in [D]$ to form the vectors $\boldr$, $\boldz_1$, and $\boldell$ respectively, 
we obtain the following \textit{relaxed} constraint inference problem for the setting of a single affine-parameterized scalar constraint:
\begin{subequations} \label{Eqn: KKT, Inverse, for Unions, Affine-Parameterized}
\begin{align} \label{Eqn: KKT, Inverse, for Unions, Affine-Parameterized, Objective}
    \text{find} \hspace{5mm} &\theta, \boldlambda_k, \boldlambda_{\urk}, \boldnu, \boldr, \boldz_1, \boldz_2, \boldell \\
    \st \hspace{5mm} &g_\urk(\xi_d^\loc, \theta) \leq 0, \hspace{5mm} \forall \ d \in [D], \\
    &\boldlambda_k, \boldlambda_\urk \geq 0, \\
    &\begin{bmatrix}
        \lambda_{d, \urk} \\ - g_\urk(\xi_d^\loc, \theta)
    \end{bmatrix} \leq 
    M \begin{bmatrix}
        z_{d,1} \\ z_{d,2}
    \end{bmatrix}, \hspace{5mm} \forall \ d \in [D], \\
    &z_{d,1} + z_{d,2} \leq 1,  \hspace{5mm} \forall \ d \in [D], \\
    &\nabla_{\xi^i} J^i(\xi_d^\loc) + (\boldlambda_{d,k})^\top \nabla_{\xi^i} \boldg_k^i(\xi_d^\loc) + r_d \\ \nonumber
    &\hspace{1cm} + \boldnu_d \nabla_{\xi^i} \boldh(\xi_d^\loc) = 0, \\ \nonumber
    &\hspace{5mm} \forall \ i \in [N], d \in [D], \\
    &\min\{0, \underM\} \textbf{1} \leq r_d \leq \overM \textbf{1}, \hspace{5mm} \forall \ d \in [D], \\
    &\underM z_{d,1} \leq r_d \leq \overM z_{d,1}, \hspace{5mm} \forall \ d \in [D], \\
    &\ell_d - (\textbf{1} - z_{d,1}) \overM \leq r_d \leq \ell_d - (\textbf{1} - z_d) \underM, \\ \nonumber
    &\hspace{5mm} \forall \ d \in [D], \\
    &r_{d,1} \leq \ell_d + (\textbf{1} - z_{d,1}), \hspace{5mm} \forall \ d \in [D],
\end{align}
\end{subequations}


\subsection{Theoretical Analysis and Proofs}
\label{subsec: App, Theoretical Analysis and Proofs}

In this appendix, we will first present theoretical results showing that our algorithm provides an 1) inner approximation of the true safe set and 2) outer approximation of the true unsafe set (App. \ref{subsubsec: App, Theoretical Guarantees of Learnability}), which enables robust safety for downstream planning. Then, we will present theoretical results describing some of the limits of constraint learnability in the multi-agent setting (App. \ref{subsubsec: App, Theoretical Limitations of Learnability}).

\subsubsection{Theoretical Guarantees for Inner- (Outer-) Approximating Safe (Unsafe) Sets}
\label{subsubsec: App, Theoretical Guarantees of Learnability}

The following results in this subsection (i.e., Section \ref{subsubsec: App, Theoretical Guarantees of Learnability}) are direct analogs of the theoretical analysis in Section V-A of \cite{Chou2020LearningConstraintsFromLocallyOptimalDemonstrationsUnderCostFunctionUncertainty} for the multi-agent setting, and are included for completeness. Specifically, we prove that the methods presented in \eqref{Eqn: KKT, Inverse, Optimal} and \eqref{Eqn: KKT, Inverse, for Unions, Affine-Parameterized} are guaranteed to recover inner approximations (i.e., conservative estimates) of the set of trajectories $\Safe(\theta^\star)$ which obey the true, unknown constraints $\boldg_\urk(\xi, \theta^\star) \leq 0$, as well as inner approximations of the set of trajectories
$\Avoid(\theta^\star)$ which violate $\boldg_\urk(\xi, \theta^\star) \leq 0$. 

Next, we show that in the setting of affine-parameterized constraint sets, the solution to \eqref{Eqn: KKT, Inverse, for Unions, Affine-Parameterized} likewise provides an inner approximation to the true feasible set. Below, to disambiguate between the solutions of the MINLP originally associated with the affine-parameterized constraints in Section \ref{subsubsec: App, Reformulation of the Constraint Inference Problem for Unions of Affine-Parameterized Constraints} (i.e., without the relaxation involving the new variables $r$, $z$, and $\ell$), we will denote the subset of parameters consistent with the demonstrations $\D$ under the \textit{original MINLP problem} by $\F_\MINLP(\D)$, and the subset of parameters consistent with the demonstrations $\D$ under the \textit{relaxed problem} by $\F_\relaxed(\D)$. Similarly, we will denote the guaranteed safe set learned by solving the original MINLP problem by $\G_{s, \MINLP}(\D)$, and the guaranteed safe set learned by solving the relaxed problem by $\G_{s, \relaxed}(\D)$.

\begin{theorem}(\textbf{\textit{Conservativeness of Safe and Unsafe Set Recovery from \eqref{Eqn: KKT, Inverse, for Unions, Affine-Parameterized}}})
Given any set $\D$ of interaction demonstrations at local Nash stationarity, the solution to \eqref{Eqn: KKT, Inverse, for Unions, Affine-Parameterized} generates a learned guaranteed safe set $\G_{s, \relaxed}(\D)$ and and a learned guaranteed unsafe set $\G_{\urs, \relaxed}(\D)$ satisfying $\G_{s, \relaxed}(\D) \subseteq \Safe(\theta^\star)$ and $\G_{\urs, \relaxed}(\D) \subseteq \Avoid(\theta^\star)$.
\end{theorem}

\begin{proof}
Since $\F_\MINLP$ denotes the set of parameters consistent with the original problem (pre-relaxation), Theorem \ref{Thm: Conservativeness of Safe and Unsafe Set Recovery from KKT, Inverse, Optimal} implies that $\G_{s, \MINLP}(\D) \subseteq \Safe(\theta^\star)$ and $\G_{\urs, \MINLP}(\D) \subseteq \Avoid(\theta^\star)$. Thus, it suffices to show that $\G_{s, \relaxed}(\D) \subseteq \G_{s, \MINLP}(\D)$ and $\G_{\urs, \relaxed}(\D) \subseteq \G_{\urs, \MINLP}(\D)$. By definition:
\begin{align} \nonumber
    \G_{s, \relaxed}(\D) &:= \bigcap_{\theta \in \F_\relaxed(\D)} \big\{ \xi \in \R^{(n+m)T}: \boldg(\xi, \theta) > 0 \big\} \\ \nonumber
    \G_{s, \MINLP}(\D) &:= \bigcap_{\theta \in \F_\MINLP(\D)} \big\{ \xi \in \R^{(n+m)T}: \boldg(\xi, \theta) > 0 \big\}.
\end{align}
Thus, to show that $\G_{s, \relaxed}(\D) \subseteq \G_{s, \MINLP}(\D)$, it suffices to show that $\F_\MINLP(\D) \subseteq \F_\relaxed(\D)$, which in turn holds true, since relaxing the original problem enlarges the set of feasible parameters. We therefore have $\G_{s, \MINLP}(\D) \subseteq \Safe(\theta^\star)$, and through a similar chain of logic, $\G_{\urs, \MINLP}(\D) \subseteq \Avoid(\theta^\star)$.
\end{proof}

\subsubsection{Volume Extraction of Safe Trajectories for Motion Planning}
\label{subsubsec: App, Volume Extraction of Safe Trajectories for Motion Planning}

First, we prove that the volume extraction methods formulated in Section \ref{subsec: Volume Extraction Over Trajectories and Parameters for Motion Planning} yield inner approximations of the true, unknown inequality constraint set $\Safe(\theta^\star) := \{\xi \in \R^{(n+m)T}: \boldg(\xi, \theta^\star) \leq 0 \}$.

\begin{theorem}(\textit{\textbf{Volume Extraction over Trajectories Generates Conservative Safe Set}}) \label{Thm: Volume Extraction, Trajectories, Generates Conservative Safe Set}
Let $\Q_\xi := \{\xi_q: q \in [Q_\xi] \}$ denote a set of query trajectories, and let $\{\epsilon_q: q \in [Q_\xi] \}$ denote the minimum objective values obtained by solving \eqref{Eqn: Volume Extraction for Safe Set, Trajectories} using each queried trajectory. Then \eqref{Eqn: Volume Extraction, Trajectory xi, inner approx for Safe set} holds.
\end{theorem}

\begin{proof}
Suppose by contradiction that \eqref{Eqn: Volume Extraction, Trajectory xi, inner approx for Safe set} is false, i.e., there exists some $q \in [Q_\xi]$ such that $\epsilon_q > 0$, and some $\xi' \in B_{\epsilon_q}^\infty(\xi_q)$ such that $\xi' \not\in \G_s(\D)$. By definition of $\G_s(\D)$ (as given by \eqref{Eqn: Gs, Def}) and $\F(\D)$, there exists some $\theta, \boldlambda_k, \boldlambda_\urk$, and $\boldnu$ such that $(\theta, \boldlambda_k, \boldlambda_\urk, \boldnu) \in \KKT(\D)$ but $\boldg(\xi', \theta) > 0$. Thus, $(\xi', \theta, \boldlambda_k, \boldlambda_\urk, \boldnu)$ is feasible for \eqref{Eqn: Volume Extraction for Safe Set, Trajectories}, and so the solution for \eqref{Eqn: Volume Extraction for Safe Set, Trajectories}, given by $\epsilon_q$, is less than or equal to $\Vert \xi' - \xi_q \Vert_\infty$, contradicting the fact that $\xi' \in B_{\epsilon_q}^\infty (\xi_q)$.
\end{proof}

Next, we prove that the volume extraction methods formulated in Sec. \ref{subsec: Volume Extraction Over Trajectories and Parameters for Motion Planning} yield inner approximations of the true, unknown inequality constraint set $S(\theta^\star) := \{\xi \in \R^{(n+m)T}: \boldg(\xi, \theta^\star) \leq 0 \}.$

\begin{theorem}(\textbf{Volume Extraction over Parameters Generates Conservative Safe Set})
Let $\{\theta_q: q \in [Q_\theta] \}$ denote a set of queried parameters, and let $\{r_q: q \in [Q_\theta] \}$ denote the minimum objective values obtained by solving \eqref{Eqn: Volume Extraction for Safe Set, Theta} using each queried trajectory. 
Then \eqref{Eqn: Volume Extraction, theta, generates Safe set} holds.
\end{theorem}

\begin{proof}
By definition of $\G_s(\D)$ in \eqref{Eqn: Gs, Def}, it suffices to prove that:
\begin{align*}
    \F(\D) \Big\backslash \bigcup_{q \in [Q_\theta]} B_{r_q}^\infty(\theta_q) = \F(\D),
\end{align*}
or equivalently, that:
\begin{align} \label{Eqn: F(D) has empty intersection with neighborhoods of queried thetas}
    \F(\D) \bigcap \Big( \bigcup_{q \in [Q_\theta]} B_{r_q}^\infty(\theta_q) \Big) = \emptyset.
\end{align}
Suppose by contradiction that \eqref{Eqn: F(D) has empty intersection with neighborhoods of queried thetas}, i.e., there exists some $q \in [Q_\theta]$, $\theta' \in \R^{(n+m)T}$ such that $r_q > 0$ and $\theta' \in B_{r_q}^\infty(\theta_q) \bigcap \F(\D)$. By definition of $\F(\D)$, there exist $\boldlambda_k, \boldlambda_\urk, \boldnu$ such that $(\theta', \boldlambda_k, \boldlambda_\urk, \boldnu) \in \KKT(\D)$, i.e., such that $(\theta', \boldlambda_k, \boldlambda_\urk, \boldnu)$ lies in the feasible set of \eqref{Eqn: Volume Extraction for Safe Set, Theta}. Thus, the solution for \eqref{Eqn: Volume Extraction for Safe Set, Trajectories}, as given by $r_q$, is less than or equal to $\Vert \theta' - \theta_q \Vert_\infty$, contradicting the fact that $\theta' \in B_{r_q}^\infty (\theta_q)$.


\end{proof}

The guaranteed unsafe set $\G_\urs$ can be similarly characterized via a volume extraction procedure over the parameter space. 

Above, we have established guarantees that our volume extraction method will always recover the safe set $\G_s$ and the avoid set $\G_\urs$ either accurately or conservatively, as well as conditions under which the constraint parameter $\theta$ can never be fully recovered, thus establishing theoretical limitations on learnability.
Regardless, since the extracted estimates of $\G_s$ and $\G_\urs$ (termed $\hat \G_s$ and $\hat \G_\urs$ below, respectively) either equal or are inner approximations of $\G_s$ and $\G_\urs$, respectively, they can be directly used by a motion planner downstream to verify constraint satisfaction or violation. 
Alternatively, a motion planner can design safe trajectories with \textit{implicit} constraint checking, e.g., via Model Predictive Path Integral (MPPI) control \cite{
Williams2016AggressiveDrivingwithMPPI
}. More details are provided in App. \ref{subsec: App, Robust Motion Planning via Implicit Constraint Checking}. In particular, any generated trajectory in $\hat \G_s$ is guaranteed to be safe, while any trajectory outside $\hat G_\urs$ is at least not guaranteed to be unsafe.



\subsubsection{Theoretical Limitations of Learnability}
\label{subsubsec: App, Theoretical Limitations of Learnability}



Finally, we analyze the limitations of multi-agent constraint learning in settings in which some agents have constraints that are strictly more lax than the constraints of other agents, and are thus undetectable from interaction demonstrations at local Nash stationarity. As an example, consider interactions between two agents (Agents 1 and 2), in which Agent 1 is constrained to maintain a distance of $\theta_1 = 1$ meter away from Agent 2 at all times, while Agent 2 is constrained to maintain a distance of $\theta_2 = 2$ meters away from Agent 1 at all times. Since $\theta_2 > \theta_1$, the two agents will maintain a distance of $\theta_2 = \theta_2$ meters apart at all times in any local Nash stationary interaction. Thus, from demonstrations of local Nash stationary interactions, it would be impossible to disambiguate Agent 1's constraint parameter ($\theta_1 = 1$) from any other possible constraint parameter for Agent 1 (e.g., $\theta_1 = 1.5$) that is also consistent with all interactions at local Nash stationarity between the two agents. Therefore, $\theta_1$ can never be precisely recovered from local Nash interactions. 

We introduce the following theorem to concretely characterize this intuition. Let $\D(\theta)$ denote the set of all local Nash stationary equilibrium trajectories corresponding to the constraints $\boldh(\xi) = 0$, $\boldg_k(\xi) \leq 0$, and $\boldg_\urk(\xi, \theta) \leq 0$, i.e.,:
\begin{align} 
    \D(\theta) &:= \{\xi \in \R^{(n+m)T}: \ \exists \ \boldlambda_k, \boldlambda_\urk, \boldnu \\ \nonumber
    &\hspace{1cm} \st (\theta, \boldlambda_k, \boldlambda_\urk, \boldnu) \in \KKT(\{\xi\}) \}.
\end{align}
Suppose the true, unknown constraint vector $\boldg_\urk(\xi, \theta^\star)$ can be partitioned into two components, $\boldg_\urk^{\paren{1}}(\xi, \theta^\star)$ and $\boldg_\urk^\paren{2}(\xi, \theta^\star)$, where the constraint $\boldg_\urk^\paren{2}(\xi, \theta^\star) \leq 0$ is strictly looser than the remaining constraints $\boldh(\xi) = 0$, $\boldg_k(\xi) \leq 0$, $\boldg_\urk^\paren{1}(\xi, \theta^\star) \leq 0$, in the sense described below:
\begin{align} 
    &\{\xi \in \R^{(n+m)T}: \boldh(\xi) = 0, \boldg_k(\xi) \leq 0, \\ \nonumber
    &\hspace{6mm} \boldg_\urk^\paren{1}(\xi, \theta^\star) \leq 0 \} \subseteq \{\xi \in \R^{(n+m)T}: \boldg_\urk^\paren{2}(\xi, \theta^\star) < 0 \}.
\end{align}
Then, for any parameter value $\theta$ satisfying:
\begin{align} 
    &\boldg_\urk^\paren{1}(\cdot, \theta^\star) = \boldg_\urk^\paren{1}(\cdot, \theta), \\ 
    &\{\xi \in \R^{(n+m)T}: \boldh(\xi) = 0, \boldg_k(\xi) \leq 0, \boldg_\urk^\paren{1}(\xi, \theta^\star) \leq 0 \} \\ \nonumber
    &\hspace{1cm} \subseteq \{\xi \in \R^{(n+m)T}: \boldg_\urk^\paren{2}(\xi, \theta) < 0 \},
\end{align}
we will prove that $\D(\theta^\star) = \D(\theta)$.
In words, each local Nash stationary demonstration corresponding to the constraint parameter $\theta^\star$ is also at local Nash stationarity under the constraint parameter $\theta$, and vice versa. Thus, from the perspective of constraint learning from demonstrations, the parameter value $\theta$ will always remain as a valid alternative to the true parameter value $\theta^\star$.


Below, we repeat the statement of Thm. \ref{Thm: Limitations of Learnability} for convenience and present its proof.

\begin{theorem}[\textbf{Limitations of Learnability}] 
Let $\boldg_\urk^\paren{1}$ and $\boldg_\urk^\paren{2}$ denote a partition of the components of the unknown constraint $\boldg_\urk$ (i.e., $(\boldg_\urk^\paren{1}, \boldg_\urk^\paren{2})$ is a re-ordering of $\boldg_\urk$) such that \eqref{Eqn: g2 subsumed by else, theta star} holds, i.e.,:
\begin{align} \nonumber
    &\{\xi \in \R^{(n+m)T}: \boldh(\xi) = 0, \boldg_k(\xi) \leq 0, \boldg_\urk^\paren{1}(\xi, \theta^\star) \leq 0 \} \\ \nonumber
    &\hspace{1cm} \subseteq \{\xi \in \R^{(n+m)T}: \boldg_\urk^\paren{2}(\xi, \theta^\star) < 0 \}.
\end{align}
Then, for any $\theta \in \Theta$ satisfying \eqref{Eqn: g1 same across theta, theta star} and \eqref{Eqn: g2 subsumed by else, theta}, i.e.,:
\begin{align} \nonumber
    &\boldg_\urk^\paren{1}(\cdot, \theta^\star) = \boldg_\urk^\paren{1}(\cdot, \theta), \\ \nonumber
    &\{\xi \in \R^{(n+m)T}: \boldh(\xi) = 0, \boldg_k(\xi) \leq 0, \boldg_\urk^\paren{1}(\xi, \theta^\star) \leq 0 \} \\ \nonumber
    &\hspace{1cm} \{\xi \in \R^{(n+m)T}: \boldg_\urk^\paren{2}(\xi, \theta) < 0 \}.
\end{align}
we have $\D(\theta^\star) = \D(\theta)$.
\end{theorem}

\begin{proof}
Fix $\xi \in \D(\theta)$ arbitrarily; we aim to show that $\xi \in \D(\theta^\star)$. Since $\xi \in \D(\theta)$, there exist $\boldlambda_k$, $\boldlambda_\urk^\paren{1}$, $\boldlambda_\urk^\paren{2}$, and $\boldnu$ such that $(\theta, \boldlambda_k$, $\boldlambda_\urk^\paren{1}$, $\boldlambda_\urk^\paren{2}) \in \KKT(\{\xi\})$, i.e., for each $i \in [N]$:
\begin{subequations} \label{Eqn: KKT, Limitations of Learnability, theta}
\begin{align} \label{Eqn: KKT, Limitations of Learnability, theta, Primal Feasibility}
    &\boldh^i(\xi) = 0, \ \boldg_k^i(\xi) \leq 0, \ \boldg_\urk^{i, \paren{1}}(\xi, \theta) \leq 0, \ \boldg_\urk^{i, \paren{2}}(\xi, \theta) \leq 0, \\ 
    \label{Eqn: KKT, Limitations of Learnability, theta, Lagrange Multiplier non-negativity}
    &\boldlambda_k^i, \boldlambda_\urk^{i, \paren{1}}, \boldlambda_\urk^{i, \paren{2}} \geq 0, \\ 
    \label{Eqn: KKT, Limitations of Learnability, theta, Complementary Slackness}
    &\boldlambda_k^i \odot \boldg_k^i(\xi) = 0, \ \boldlambda_\urk^i \odot \boldg_\urk^{i, \paren{1}}(\xi, \theta) = 0, \\ \nonumber
    &\hspace{5mm} \boldlambda_\urk^{i, \paren{2}} \odot \boldg_\urk^{i, \paren{2}}(\xi, \theta) = 0, \\ 
    \label{Eqn: KKT, Limitations of Learnability, theta, Stationarity}
    &\nabla_{\xi^i} J^i(\xi) + (\boldlambda_k^i)^\top \nabla_{\xi^i} \boldg_k^i(\xi) + (\boldlambda_\urk^{i, \paren{1}})^\top \nabla_{\xi^i} \boldg_\urk^{i, \paren{1}}(\xi, \theta) \\ \nonumber
    &\hspace{1cm} + (\boldlambda_\urk^{i, \paren{2}})^\top \nabla_{\xi^i} \boldg_\urk^{i, \paren{2}}(\xi, \theta) \\ \nonumber
    &\hspace{1cm} + (\boldnu^i)^\top \nabla_{\xi^i} \boldh^i(\xi) = 0.
\end{align}
\end{subequations}
By \eqref{Eqn: KKT, Limitations of Learnability, theta, Primal Feasibility}, \eqref{Eqn: g1 same across theta, theta star}, and \eqref{Eqn: g2 subsumed by else, theta}, we have $\boldg_\urk^\paren{2}(\xi, \theta) < 0$. Substituting $\boldg_\urk^\paren{2}(\xi, \theta) < 0$ into \eqref{Eqn: KKT, Limitations of Learnability, theta, Complementary Slackness} then yields $\boldlambda_\urk^\paren{2} = 0$, and thus, \eqref{Eqn: KKT, Limitations of Learnability, theta, Stationarity} becomes:
\begin{align} \label{Eqn: KKT, Limitations of Learnability, theta, Stationarity, lambda 2 is 0}
    &\nabla_{\xi^i} J^i(\xi) + (\boldlambda_k^i)^\top \nabla_{\xi^i} \boldg_k^i(\xi) + (\boldlambda_\urk^{i, \paren{1}})^\top \nabla_{\xi^i} \boldg_\urk^{i, \paren{1}}(\xi, \theta) \\ \nonumber
    &\hspace{1cm} + (\boldnu^i)^\top \nabla_{\xi^i} \boldh^i(\xi) = 0.
\end{align}
We now claim that, for each $i \in [N]$:
\begin{subequations} \label{Eqn: KKT, Limitations of Learnability, theta star}
    \begin{align} \label{Eqn: KKT, Limitations of Learnability, theta star, Primal Feasibility}
    &\boldh^i(\xi) = 0, \ \boldg_k^i(\xi) \leq 0, \ \boldg_\urk^{i, \paren{1}}(\xi, \theta^\star) \leq 0, \ \boldg_\urk^{i, \paren{2}}(\xi, \theta^\star) \leq 0, \\ 
    \label{Eqn: KKT, Limitations of Learnability, theta star, Lagrange Multiplier non-negativity}
    &\boldlambda_k^i, \boldlambda_\urk^{i, \paren{1}}, \boldlambda_\urk^{i, \paren{2}} \geq 0, \\ 
    \label{Eqn: KKT, Limitations of Learnability, theta star, Complementary Slackness}
    &\boldlambda_k^i \odot \boldg_k^i(\xi) = 0, \ \boldlambda_\urk^i \odot \boldg_\urk^{i, \paren{1}}(\xi, \theta^\star) = 0, \\ \nonumber
    &\hspace{1cm} \boldlambda_\urk^{i, \paren{2}} \odot \boldg_\urk^{i, \paren{2}}(\xi, \theta^\star) = 0, \\ 
    \label{Eqn: KKT, Limitations of Learnability, theta star, Stationarity}
    &\nabla_{\xi^i} J^i(\xi) + (\boldlambda_k^i)^\top \nabla_{\xi^i} \boldg_k^i(\xi) + (\boldlambda_\urk^{i, \paren{1}})^\top \nabla_{\xi^i} \boldg_\urk^{i, \paren{1}}(\xi, \theta^\star) \\ \nonumber
    &\hspace{1cm} + (\boldlambda_\urk^{i, \paren{2}})^\top \nabla_{\xi^i} \boldg_\urk^{i, \paren{2}}(\xi, \theta^\star) \\ \nonumber
    &\hspace{1cm} + (\boldnu^i)^\top \nabla_{\xi^i} \boldh^i(\xi) = 0,
\end{align}
\end{subequations}
Below, we prove \eqref{Eqn: KKT, Limitations of Learnability, theta star, Primal Feasibility}-\eqref{Eqn: KKT, Limitations of Learnability, theta star, Stationarity}. To prove that \eqref{Eqn: KKT, Limitations of Learnability, theta star, Primal Feasibility} holds, we first note that \eqref{Eqn: KKT, Limitations of Learnability, theta, Primal Feasibility} and \eqref{Eqn: g1 same across theta, theta star} imply that $\boldh(\xi) = 0$, $\boldg_k(\xi) \leq 0$, and $\boldg_\urk^\paren{1}(\xi, \theta^\star) \leq 0$, which, in conjunction with \eqref{Eqn: g2 subsumed by else, theta star}, implies that $\boldg_\urk^\paren{2}(\xi, \theta^\star) < 0$. Meanwhile, \eqref{Eqn: KKT, Limitations of Learnability, theta star, Lagrange Multiplier non-negativity} is identical to \eqref{Eqn: KKT, Limitations of Learnability, theta, Lagrange Multiplier non-negativity}, and thus holds. Next, to show \eqref{Eqn: KKT, Limitations of Learnability, theta star, Complementary Slackness}, note that \eqref{Eqn: KKT, Limitations of Learnability, theta, Complementary Slackness} implies $\boldlambda_k^i \odot \boldg_k^i(\xi) = 0$ for each $i \in [N]$, \eqref{Eqn: KKT, Limitations of Learnability, theta, Complementary Slackness} and \eqref{Eqn: g1 same across theta, theta star} implies that $\boldlambda_k^i \odot \boldg_\urk^{i, \paren{1}}(\xi, \theta) = \boldlambda_k^i \odot \boldg_\urk^{i, \paren{1}}(\xi, \theta^\star) = 0$, while the fact that $\boldlambda_\urk^{(2)} = 0$ implies $\boldlambda_\urk^{(2)} \odot \boldg_\urk^{i, \paren{2}}(\xi, \theta^\star) = 0$. Finally, to verify \eqref{Eqn: KKT, Limitations of Learnability, theta star, Stationarity}, we combine the fact that $\boldlambda_\urk^{(2)} = 0$, \eqref{Eqn: g1 same across theta, theta star}, and \eqref{Eqn: KKT, Limitations of Learnability, theta, Stationarity} to conclude that:
\begin{align*}
    &\nabla_{\xi^i} J^i(\xi) + (\boldlambda_k^i)^\top \nabla_{\xi^i} \boldg_k^i(\xi) + (\boldlambda_\urk^{i, \paren{1}})^\top \nabla_{\xi^i} \boldg_\urk^{i, \paren{1}}(\xi, \theta^\star) \\ \nonumber
    &\hspace{1cm} + (\boldlambda_\urk^{i, \paren{2}})^\top \nabla_{\xi^i} \boldg_\urk^{i, \paren{2}}(\xi, \theta^\star) + (\boldnu^i)^\top \nabla_{\xi^i} \boldh^i(\xi) \\
    = \ &\nabla_{\xi^i} J^i(\xi) + (\boldlambda_k^i)^\top \nabla_{\xi^i} \boldg_k^i(\xi) + (\boldlambda_\urk^{i, \paren{1}})^\top \nabla_{\xi^i} \boldg_\urk^{i, \paren{1}}(\xi, \theta) \\
    &\hspace{1cm} + (\boldnu^i)^\top \nabla_{\xi^i} \boldh^i(\xi) \\
    = \ &\nabla_{\xi^i} J^i(\xi) + (\boldlambda_k^i)^\top \nabla_{\xi^i} \boldg_k^i(\xi) + (\boldlambda_\urk^{i, \paren{1}})^\top \nabla_{\xi^i} \boldg_\urk^{i, \paren{1}}(\xi, \theta) \\
    &\hspace{1cm} + (\boldlambda_\urk^{i, \paren{2}})^\top \nabla_{\xi^i} \boldg_\urk^{i, \paren{2}}(\xi, \theta) + (\boldnu^i)^\top \nabla_{\xi^i} \boldh^i(\xi)  \\
    = \ &0.
\end{align*}
as desired. As a result, \eqref{Eqn: KKT, Limitations of Learnability, theta star} holds, i.e., $(\theta^\star, \boldlambda_k, \boldlambda_\urk^\paren{1}, \boldlambda_\urk^\paren{2}, \boldnu) \in \KKT(\{\xi\})$, so $\xi \in \D(\theta^\star)$. We have thus established that $\D(\theta) \subseteq \D(\theta^\star)$.

To show that $\D(\theta^\star) \subseteq \D(\theta)$, it suffices to reverse the logic of the above proof to establish \eqref{Eqn: KKT, Limitations of Learnability, theta} from \eqref{Eqn: KKT, Limitations of Learnability, theta star}.
\end{proof}

\subsection{Robust Motion Planning via Implicit Constraint Checking}
\label{subsec: App, Robust Motion Planning via Implicit Constraint Checking}

In this appendix, we first discuss how implicit constraint verification can be used within the loop of a Model Predictive Path Integral (MPPI) controller to plan new trajectories that are robustly safe under constraint uncertainty (App. \ref{subsubsec: App, MPPI Control}). Then, we apply this method to plan a safe trajectory for a two-agent double integrator system, with results presented in App. \ref{subsubsec: App, MPPI Simulation}.

\subsubsection{Model Predictive Path Integral (MPPI) Control}
\label{subsubsec: App, MPPI Control}

In this section, we elaborate on the use of our multi-agent constraint learning framework to design robust motion plans, by performing implicit constraint verification via Model Predictive Path Integral (MPPI) control.

First, we note that multiple candidate values of the constraint parameter $\theta$ may in general be consistent with the local Nash stationarity of interaction demonstrations in a given dataset $\D$. Thus, the KKT conditions \eqref{Eqn: KKT, Forward Game} only pose \textit{necessary, but not sufficient} criteria that the true value of $\theta$ must satisfy. To design robust motion plans despite this uncertainty over $\theta$, we perform robust constraint checking to ensure that each designed, candidate trajectory is marked safe by a $\theta$ value consistent with \eqref{Eqn: KKT, Forward Game}. 
To this end, we present an \textit{implicit} approach for constraint checking via Model Predictive Path Integral (MPPI) control \cite{
Williams2016AggressiveDrivingwithMPPI
}. 

Specifically, given a nominal state-control trajectory $\xi_\nom := (x_\nom, u_\nom)$, we enforce constraint satisfaction by iteratively applying Algorithm \ref{Alg: Inverse KKT Guided MPPI Sampling (1 Iterate)} to update the trajectory $\xi_\nom$ until convergence. 
In each iterate of Algorithm \ref{Alg: Inverse KKT Guided MPPI Sampling (1 Iterate)}, we first draw $M$ 
sample controls $\{u^\sample: s \in [M] \}$ i.i.d. from a Gaussian distribution centered on $u_\nom$, which we then unroll using the dynamics $f$ to generate the sample state-control trajectories $\{\xi^\sample: s \in [M] \}$. Next, for each $\xi^\sample$, we wish to determine whether there exists a feasible constraint parameter value $\theta$ which is consistent with the demonstrated interactions $\D$, but not with $\xi^\sample$. In other words, we wish to solve the feasibility problem \eqref{Eqn: KKT, Inverse, Optimal}, augmented with the stipulation that the sample trajectory $\xi^\sample$ \textit{violates} the unknown primal inequality constraints $\bigwedge_{i \in [N], j \in [N_\urk^{\ineq, i}]} \{g_{j,\urk}^i(\xi^\sample, \theta) \leq 0 \}$:
\begin{subequations}
\label{Eqn: App MPPI, KKT, Inverse, Augmented}
\begin{align} \label{Eqn: App MPPI, KKT, Inverse, Augmented, Objective}
    \text{find} \hspace{5mm} &\theta, \boldlambda_k, \boldlambda_\urk, \boldnu, \\ \label{Eqn: App MPPI, KKT, Inverse, Augmented, Constraints, 1}
    \text{s.t.} \hspace{5mm} &(\theta, \boldlambda_k, \boldlambda_\urk, \boldnu) \in \KKT(\D), \\ \label{Eqn: App MPPI, KKT, Inverse, Augmented, Constraints, 2}
    &\bigvee_{i \in [N], j \in [N_\urk^{\ineq, i}]} \big[ g_{j,\urk}^i(\xi^\sample, \theta) > 0 \big].
\end{align}
\end{subequations}
If \eqref{Eqn: App MPPI, KKT, Inverse, Augmented} returns feasible, there exists some constraint parameter value $\theta^\sample$ consistent with the demonstrations $\D$ such that $\xi^\sample$ violates the inequality constraints $\bigwedge_{i,j} \{g_{j,\urk}^i(\xi^\sample, \theta^\sample) \leq 0 \}$. In this case, we compute the total constraint violation for each agent $i$, denoted $c_{\cv}^i$ below, as follows:
\begin{align} \label{Eqn: App MPPI, c cv, Def}
    c_{\cv}^i(\xi^\sample, \theta^\sample) := 
    \sum_{j \in [N_\urk^{\ineq, i}]} \max\{ g_{j, \urk}^i(\xi^\sample, \theta^\sample) , 0\}.
\end{align}
Otherwise, if \eqref{Eqn: App MPPI, KKT, Inverse, Augmented} is infeasible, we set $c_{\cv}^i(\xi^\sample) := 0$ for each $i \in [N]$. We then update the nominal control trajectory for each agent $i \in [N]$ by taking a weighted combination of the sample control trajectories $\{u^\sample: s \in [M] \}$, with weights depending on the corresponding cost value $J^i(\xi^\sample)$ and degree of constraint violation $c_{\cv}(\xi^\sample)$:
\begin{align} \label{Eqn: App MPPI, u nom update}
    \tilde u_{\nom}^{i} := \frac{\sum_{s \in [M]} \exp(- J^i(\xi^\sample) - c_{\cv}(\xi^\sample, \theta^\sample)) u^\sample}{\sum_{s \in [M]} \exp(- J^i(\xi^\sample) - c_{\cv}(\xi^\sample, \theta^\sample ))}.
\end{align}
Finally, we unroll the updated nominal control trajectory $\{\tilde u_{\nom}^i: i \in [N] \}$ to generate the updated nominal state-control trajectory $\tilde \xi^i$.

\begin{algorithm} 
{
\normalsize
\SetAlgoLined
\KwData{Nominal state-control trajectory $\xi_\nom$, dynamics model $f$, demonstrations $\mathcal{D}$, number of samples $M$
} 
 \vspace{2mm}

$\{\xi^\sample: s \in [M] \} \gets$ $N$ sample state-control trajectories generated by perturbing $\xi_\nom$ via a Gaussian distribution, and enforcing the dynamics $f$.

\For{$s \in [M]$}{

    Solve augmented inverse KKT problem \eqref{Eqn: App MPPI, KKT, Inverse, Augmented}
    
    \If{\eqref{Eqn: App MPPI, KKT, Inverse, Augmented} is feasible}{
         $\theta^\sample \gets$ Feasible $\theta$ value from solving \eqref{Eqn: App MPPI, KKT, Inverse, Augmented}

         $c_{\cv}^i(\xi^\sample, \theta^\sample) \gets$ \eqref{Eqn: App MPPI, c cv, Def}, $\ \forall \ i \in [N]$.
    }
    \Else{
            $c_{\cv}^i(\xi^\sample, \theta^\sample) \gets 0$
    }

}

$\{\tilde u_{\nom}^i: i \in [N]\} \gets$ 
\eqref{Eqn: App MPPI, u nom update}, using $c_\cv^i(\xi^\sample, \theta^\sample)$, $J^i(\xi^\sample)$, and $u^{(s)}$.

$\{\tilde \xi_{\nom}^i: i \in [N]\} \gets$ Unroll $\{\tilde u_{\nom}^i: i \in [N]\}$ using dynamics $f$

 \Return{Updated nominal state-control trajectory $\tilde \xi_\nom$}
 
 \caption{Inverse KKT-Guided MPPI control-based Sample Trajectory Update (1 Iterate)}
 \label{Alg: Inverse KKT Guided MPPI Sampling (1 Iterate)}
 }
\end{algorithm}

\subsubsection{Model Predictive Path Integral (MPPI) Simulation}
\label{subsubsec: App, MPPI Simulation}

\begin{figure}[ht]
    \centering
    \includegraphics[width=0.9\linewidth]{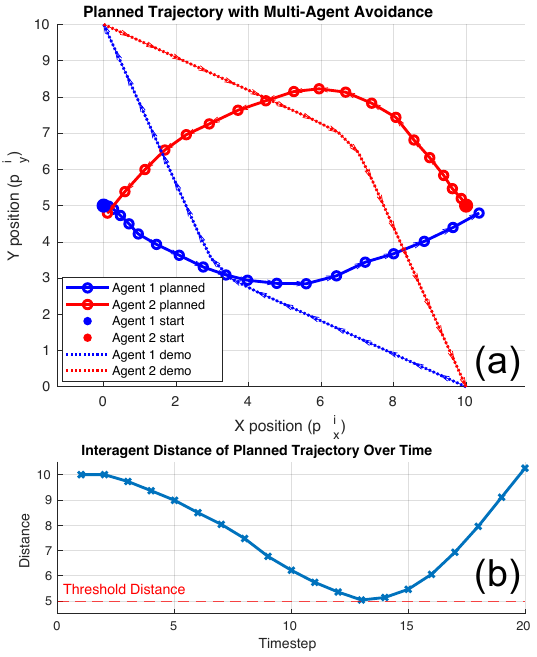}  
    \caption{(a) The multi-agent trajectories \& (b) inter-agent distance checking w.r.t ground truth threshold. The planner can sample trajectories that satisfy \textit{unknown or intangible} constraints embedded in the demonstration, without explicitly solving for the actual constraint parameters.}
    \label{fig:MPPI_planner_combined}
\end{figure}

Below, we present a simulation result illustrating that motion plans designed using our MPPI-based planner satisfy safety guarantees. Consider the setting in which $N = 2$ agents navigate in a shared 2D environment over the time horizon $T = 10$. The state of each agent $i \in [N]$ is given by $x_t^i := (p_{x,t}^i, p_{y,t}^i, v_{x,t}^i, v_{y,t}^i) \in \R^4$ for each $t \in [T]$, and the system state is given by $x_t := (x_t^1, \cdots, x_t^N)$. Each agent follows the double integrator dynamics discretized at intervals of $\Delta t = 1$s. Moreover, each agent $i \in [3]$ aims to optimize the following smoothness cost:
\begin{align} \label{Eqn: App MPPI, Cost, Smoothness}
    J^i = \sum_{t=1}^{T-1} \Big[ \Vert p_{x,t+1}^i - p_{x,t}^i \Vert_2^2 + \Vert p_{y,t+1}^i - p_{y,t}^i \Vert_2^2 \Big]
\end{align}
while ensuring that their trajectory satisfies the following spherical collision avoidance constraints at each time $t \in [T]$, which are a priori unknown to the constraint learner:
\begin{align} \label{Eqn: App MPPI, Spherical Collision-Avoidance Constraints}
    g_{t,\urk}^i(\xi, \theta^i) = - \Vert p_{x,t}^i - p_{x,t}^j \Vert_2^2 + (\theta^i)^2 \leq 0.
\end{align}
Above, $\theta^i$ denotes the radius of the spherical collision avoidance set for each agent $i \in [N]$. Finally, the trajectory of each agent $i \in [N]$ is constrained by a prescribed set of origin and goal positions, given by:
\begin{align} \label{Eqn: App MPPI,Constraints, Known, Spherical}
    h_t^i(\xi) = \begin{bmatrix}
        p_0^i - \bar p_o^i \\
        p_T^i - \bar p_d^i
    \end{bmatrix} = 0,
\end{align}
where $p_t^i := (p_{x,t}^i, p_{y,t}^i)$ for each $i \in [N]$ and $t \in [T]$, while $\bar p_o^i \in \R^2$ and $\bar p_d^i \in \R^2$ respectively denote the origin and destination positions for agent $i$, and are fixed at the following values:
\begin{subequations} \label{Eqn: App MPPI, Origin, Destination Locations, Spherical, Constraint Inference}
    \begin{align}
        \bar p_o^1 = (0, 5), \hspace{3mm} \bar p_o^2 = (10, 5),  \\
        \bar p_d^1 = (10, 5), \hspace{3mm} \bar p_d^2 = (0, 5).
    \end{align}
\end{subequations}
To evaluate our MPPI-based motion planning approach, we first generate a single demonstration trajectory using the costs and constraints given in \eqref{Eqn: App MPPI, Cost, Smoothness}-\eqref{Eqn: App MPPI,Constraints, Known, Spherical}, with origin and destination coordinates:
\begin{subequations} \label{Eqn: App MPPI, Origin, Destination Locations, Spherical, MPPI}
\begin{align}
    \bar p_o^1 = (10, 0), \hspace{3mm} \bar p_o^2 = (0, 10), \\
    \bar p_d^1 = (0, 10), \hspace{3mm} \bar p_d^2 = (10, 0).
\end{align}
\end{subequations}
and ground truth constraint parameters $\theta = (\theta^1, \theta^2) = (5, 5)$. To design robust trajectories, we run Algorithm \ref{Alg: Inverse KKT Guided MPPI Sampling (1 Iterate)} for 70 iterations using $M = 16$ samples, with time horizon $T = 20$ and discretization time $\Delta t = 0.1$s. As illustrated in Figure \ref{fig:MPPI_planner_combined}, our method generates trajectories for each agent which satisfy the spherical inter-agent collision avoidance constraints \eqref{Eqn: App MPPI, Spherical Collision-Avoidance Constraints} with radii $\theta^1 = \theta^2 = 5$. Our Gurobi solve time was 293.36 s. Note that motion planners which use implicit constraint representations, such as MPPI-based methods, are generally much less computationally efficient compared to motion planners which directly encode learned constraints via the volume extraction-based approach. We will illustrate the direct, volume extraction-based motion planning method across numerical simulations and hardware experiments in App. \ref{subsec: App, Experiment Details}.

\subsection{Additional Experiments and Experiment Details}
\label{subsec: App, Experiment Details}

Below, in Sec. \ref{subsubsec: App, Table of Contents for Simulation and Experimental Results}, we provide a table of contents describing the simulation and hardware results that will be presented throughout the remainder of this section. Then, in Sec. \ref{subsubsec: App, Dynamics}, we introduce the dynamics and dynamics transformations used in our simulation and hardware experiments. Sec. \ref{subsubsec: App, Constraint Parameterizations} introduces details concerning the constraint parameterizations used in our work. Finally, in Sec. \ref{subsubsec: App, Double Integrator Simulations}-\ref{subsubsec: App, Hardware Experiments}, we present additional experiments and experiment details that were omitted from the main body of the paper due to space limitations.

\subsubsection{Table of Contents for Simulation and Experimental Results}
\label{subsubsec: App, Table of Contents for Simulation and Experimental Results}

Table \ref{Table: App, Table of Contents for Experiment Results.} lists the simulation and hardware results that will be presented throughout Sec. \ref{subsubsec: App, Double Integrator Simulations}-\ref{subsubsec: App, Hardware Experiments}. Note that we also present the following two sets of results in Sec. \ref{subsubsec: App, Double Integrator Simulations}-\ref{subsubsec: App, Hardware Experiments}, although they are not listed in Table \ref{Table: App, Table of Contents for Experiment Results.}:
\begin{itemize}
    \item A plot of Gurobi solve times for our constraint learning method vs. the number of agents involved in the constraint learning problem (in Sec. \ref{subsec: Double Integrator Simulations} and \ref{subsubsec: App, Double Integrator Simulations}.)

    \item Numerical results comparing our constraint learning-based method to the cost inference-based baseline \cite{Peters2021InferringObjectives} (in Sec. \ref{subsec: Double Integrator Simulations} and \ref{subsubsec: App, Double Integrator Simulations}).
\end{itemize}


\begin{table*}[!ht]
    \setlength{\tabcolsep}{16pt} 
    \renewcommand{\arraystretch}{1.8}
    \normalsize
    \centering
    \begin{tabular}{|c|c|c|} 
    \hline
    \multicolumn{3}{|c|}{\textbf{Numerical Simulations}} \\
    \hline
    \textbf{Dynamics} & \textbf{Constraint type} & \textbf{Sections} \\
    \hline
    \multirow{3}{4em}{\centering Double Integrator} & Polytopic Collision-Avoidance & \ref{subsec: Double Integrator Simulations}, \ref{subsubsec: App, Double Integrator Simulations} \\
    \cline{2-3}
    & Elliptic Collision-Avoidance & \ref{subsec: Double Integrator Simulations}, \ref{subsubsec: App, Double Integrator Simulations} \\
    \cline{2-3}
    & Velocity-Dependent Spherical Collision-Avoidance & \ref{subsec: Double Integrator Simulations}, \ref{subsubsec: App, Double Integrator Simulations} \\
    \hline 
    \multirow{3}{4em}{\centering Unicycle} & Spherical Proximity and Collision-Avoidance & 
    \ref{subsubsec: App, Unicycle Dynamics Simulations} \\
    \cline{2-3}
    & Line-of-Sight & 
    \ref{subsubsec: App, Unicycle Dynamics Simulations} \\
    \hline 
    \multirow{3}{4em}{\centering Quadcopter} & Spherical Collision-Avoidance & \ref{subsec: Quadcopter Simulations}, \ref{subsubsec: App, Quadcopter Simulations} \\
    \cline{2-3}
    & Box-Shaped Collision-Avoidance & \ref{subsec: Quadcopter Simulations}, \ref{subsubsec: App, Quadcopter Simulations} \\
    \hline 
    \multirow{3}{4em}{\centering \vspace{-2pt}Single Integrator} & Nonlinear & 
    \ref{subsubsec: App, Single Integrator Simulations for Nonlinear Constraint Recovery} \\
    \cline{2-3}
    & Nonlinear, with \textit{a priori} Unknown Cost & 
    \ref{subsubsec: App, Single Integrator Simulations for Nonlinear Constraint Recovery} \\
    \hline 
    \hline
    \multicolumn{3}{|c|}{\textbf{Hardware Experiments}} \\
    \hline
    \textbf{Dynamics} & \textbf{Constraint type} & \textbf{Sections} \\
    \hline
    \multirow{3}{4em}{\centering Unicycle} & Spherical Proximity and Collision-Avoidance & \ref{subsec: Hardware Experiments}, \ref{subsubsec: App, Hardware Experiments} \\
    \cline{2-3}
    & Line-of-Sight & \ref{subsec: Hardware Experiments}, \ref{subsubsec: App, Hardware Experiments} \\
    \hline 
    \end{tabular}
    \vspace{2mm}
    \caption{We list the simulation and hardware experiment results presented in Sec. \ref{subsubsec: App, Double Integrator Simulations}-\ref{subsubsec: App, Hardware Experiments}, organized by the dynamics model used and the constraint type, and points to the subsections in the main paper and the appendix where the corresponding results are described.}
    \label{Table: App, Table of Contents for Experiment Results.}
\end{table*}

\subsubsection{Dynamics}
\label{subsubsec: App, Dynamics}


In Sec. \ref{sec: Experiments}, we presented numerical simulations involving systems with single and double integrator dynamics, unicycle dynamics, and quadcopter dynamics, as well as hardware experiments involving unicycle dynamics. Below, we elaborate on implementation details pertaining to the unicycle and quadcopter dynamics. In this subsection, for ease of exposition, we omit all agent and time indices that usually accompany state and control variables in this work.

\paragraph{Unicycle Dynamics}
We use the 4-dimensional unicycle dynamics model for ground robots, which is conventionally formulated using a state vector $x = (p_x, p_y, \phi, v) \in \R^4$ and a control vector $u = (u_1, u_2) \in \R^2$, where $p_x, p_y, \phi, v$ denote the $x$-position, $y$-position, heading, velocity of a 2D agent, respectively. The dynamics equations are given by:
\begin{subequations} \label{Eqn: App, Unicycle Dynamics, original}
\begin{align}
    \dot p_x &= v \cos\phi, \\
    \dot p_y &= v \sin\phi, \\
    \dot \phi &= u_1, \\
    \dot v &= u_2.
\end{align}
\end{subequations}
To formulate the line-of-sight constraints 
considered in this work as affine-parameterized constraints, we introduce the variables $(v_x, v_y) \in \R^2$, which encode the velocity vector and relate to the state $x = (p_x, p_y, \phi, v) \in \R^4$ via the following polar-to-Cartesian coordinate transformation:
\begin{align*}
    v_x &:= v \cos \phi, \\
    v_y &:= v \sin \phi.
\end{align*}
We then introduce the modified state variable $\bar x := (p_x, p_y, v_x, v_y) \in \R^4$, which evolves according to the reformulated unicycle dynamics equations\footnote{When $v_x = v_y = 0$, we set $\dot v_x = \dot v_y = 0$.}:
\begin{subequations} \label{Eqn: App, Unicycle Dynamics, with velocity vector}
\begin{align}
    \dot p_x &= v_x, \\
    \dot p_y &= v_y, \\
    \dot v_x &= - \frac{v_y}{\sqrt{v_x^2 + v_y^2}} u_1 + \frac{v_x}{\sqrt{v_x^2 + v_y^2}} u_2, \\
    \dot v_y &= \frac{v_x}{\sqrt{v_x^2 + v_y^2}} u_1 + \frac{v_y}{\sqrt{v_x^2 + v_y^2}} u_2.
\end{align}
\end{subequations}

\paragraph{Quadcopter Dynamics}
We use a 12-dimensional quadcopter dynamics model, formulated using the state vector $x \in \R^{12}$ and control vector $u \in \R^4$ shown below:
\begin{align}
    x &= (p_x, p_y, p_z, \alpha, \beta, \gamma, \dot p_x, \dot p_y, \dot p_z, \dot \alpha, \dot \beta, \dot \gamma), \\
    u &= (F, \tau_\alpha, \tau_\beta, \tau_\gamma),
\end{align}
where $p_x$, $p_y$, and $p_z$ respectively denote the $x$-, $y$-, and $z$-positions of the quadcopter, while $\alpha$, $\beta$, and $\gamma$ denote the Euler angles of the quadcopter. The dynamics equations are given by:
\begin{subequations} \label{Eqn: App, Quadcopter Dynamics}
\begin{align}
    \ddot{p}_x &= - \frac{F}{m}(\sin\alpha \sin\gamma + \cos\alpha \sin\beta \cos\gamma), \\
    \ddot{p}_y &= - \frac{F}{m}(\cos\alpha \sin\gamma - \sin\alpha \sin\beta \cos\gamma), \\
    \ddot{p}_z &= g - \frac{F}{m} \cos\beta \cos\gamma, \\
    \ddot{\alpha} &= \frac{I_y - I_z}{I_x} \dot \beta \dot \gamma, \\
    \ddot{\beta} &= \frac{I_z - I_x}{I_y} \dot \alpha \dot \gamma, \\
    \ddot{\alpha} &= \frac{I_x - I_y}{I_z} \dot \alpha \dot \beta,
\end{align}
which can be readily rewritten into the form $\dot x = f(x) + g(x,u)$ for suitably defined $f: \R^{12} \ra \R^{12}$ and $g: \R^{12} \times \R^4 \ra \R^{12}$.
\end{subequations}

\subsubsection{Constraint Parameterizations}
\label{subsubsec: App, Constraint Parameterizations}

In this section, we elaborate upon the constraint parameterizations first presented in Sec. \ref{subsec: Constraint Parameterizations and Agent Costs}. In particular, we explicitly formulate constraint parameterizations for elliptical, spherical, and polytope-shaped (\say{polytopic}) collision-avoidance constraints, spherical proximity constraints, and line-of-sight constraints. We will refer to the equations below in future subsections (i.e., Sec. \ref{subsubsec: App, Double Integrator Simulations} to \ref{subsubsec: App, Comparison Against Cost Inference Baseline}) when presenting ground truth constraint parameters.

In our work, the 
line-of-sight constraints can be expressed using the position and velocity vectors of each agent, while all remaining constraints require only the position of each agent to be specified. As stated in Sec. \ref{subsec: Constraint Parameterizations and Agent Costs}, given the state $x_t^i$ of agent $i$ at time $t$, for any $i \in [N]$ and $t \in [T]$, we denote by $p_t^i$ the components of $x_t^i$ corresponding to the position coordinates of agent $i$ at time $t$.

\paragraph{Elliptic Collision-Avoidance Constraints}
By an \textit{elliptical collision-avoidance constraint} on agent $i$, we refer to a constraint that compels agent $i$ to ensure that at each time $t \in [T]$, the relative position of any other agent $j$ with respect to the position of agent $i$, i.e., $p_t^j - p_t^i$, lies outside of an ellipse of fixed shape centered at the origin.
For two-dimensional agents (e.g., ground robots), we encode elliptical collision avoidance constraints for each agent $i$ with positions $p_t^i = (p_{x,t}^i, p_{y,t}^i) \in \R^2$ via the following parameterization of the unknown constraint set:
\begin{align} \label{Eqn: App, Elliptic Collision-Avoidance Constraints, Ground robot, def}
    &\hspace{5mm} C_\urk^{\ineq, i}(\theta) \\ \nonumber
    &= \bigwedge_{t=1}^T \bigwedge_{j \in [N] \backslash \{i\}} \\ \nonumber
    &\hspace{1cm} \Bigg\{- (p_t^j - p_t^i)^\top \begin{bmatrix}
        \theta_2^i & 0 \\
        0 & \theta_3^i
    \end{bmatrix} (p_t^j - p_t^i) + \theta_1^i \leq 0\Bigg\},
\end{align}
where $\theta = (\theta_k^i: i \in [N], k \in [3]) \in \R^{3N}$ denotes the unknown constraint parameter, with $\theta_k^i  \geq 0$ for each $i \in [N]$, $k \in [3]$. 

\paragraph{Spherical Collision Avoidance and Proximity Constraints}
By a \textit{spherical collision-avoidance constraint} on agent $i$, we refer to a constraint that compels agent $i$ to ensure that at each time $t \in [T]$, the relative position of any other agent $j$ with respect to the position of agent $i$, i.e., $p_t^j - p_t^i$, lies outside of a circle of fixed radius centered at the origin.
Spherical collision-avoidance constraints can be viewed as special instances of elliptic collision-avoidance constraints, in which the values $\theta_2^i = \theta_3^i$ are \textit{a priori} known to the constraint learner. 
The unknown constraint set for each agent $i \in [N]$ is then given by:
\begin{align} \label{Eqn: App, Spherical Collision-Avoidance Constraints, def}
    C_\urk^{\ineq, i}(\theta) &= \bigwedge_{t=1}^T \bigwedge_{j \in [N] \backslash \{i\}} \Big\{- \Vert p_t^j - p_t^i \Vert_2^2 + \theta_1^i \leq 0\Big\},
\end{align}
where $\theta = (\theta_1^i \geq 0: i \in [N]) \in \R^N$ denotes the unknown constraint parameter, with $\theta_1^i \geq 0$ for each $i \in [N]$. 

By a \textit{spherical proximity constraint} on agent $i$, we refer to a constraint that compels agent $i$ to ensure that at each time $t \in [T]$, the relative position of any other agent $j$ with respect to the position of agent $i$, i.e., $p_t^j - p_t^i$, lies \textit{inside} of a circle of fixed radius centered at the origin.
When we wish to encode spherical proximity constraints in addition to spherical collision-avoidance constraints, we introduce an \textit{upper} bound on $\Vert p_t^j - p_t^i \Vert_2^2$ in \eqref{Eqn: App, Spherical Collision-Avoidance Constraints, def} encoded by a set of new constraint parameters $(\theta_5^i: i \in [N])$, with $\theta_5^i > \theta_1^i$ for all $i \in [N]$. The unknown constraint set for each agent $i \in [N]$ is then given by:
\begin{align} \label{Eqn: App, Spherical Proximity Constraints, def}
    C_\urk^{\ineq, i}(\theta) &= \bigwedge_{t=1}^T \bigwedge_{j \in [N] \backslash \{i\}} \Big\{\theta_1^i \leq \Vert p_t^j - p_t^i \Vert_2^2 \leq \theta_5^i \Big\},
\end{align}
where $\theta = (\theta_1^i, \theta_5^i: i \in [N]) \in \R^{2N}$ denotes the unknown constraint parameter, with $\theta_5^i > \theta_1^i \geq 0$ for each $i \in [N]$.

\paragraph{Polytopic Collision-Avoidance Constraints}

By a \textit{polytope-shaped (\say{polytopic}) proximity constraint} on agent $i$, we refer to a constraint that compels agent $i$ to ensure that at each time $t \in [T]$, the relative position of any other agent $j$ with respect to the position of agent $i$, i.e., $p_t^j - p_t^i$, lies outside a polytope of fixed shape that contains the origin.
Polytope-shaped collision-avoidance sets for each agent $i \in [N]$ are encoded using unions of half-spaces, 
via the parameterization:\footnote{We have either $a_\beta(\theta) \in \R^2$ or $a_\beta(\theta) \in \R^3$, depending on whether the agents under consideration are two-dimensional, e.g., ground robots, or three-dimensional, e.g., quadcopters, respectively.}
\begin{align} \label{Eqn: App, Polytopic Collision-Avoidance Constraints, def}
    C_\urk^{\ineq, i}(\theta) &= \bigwedge_{t=1}^T \bigwedge_{j \in [N] \backslash \{i\}} \bigvee_{\beta=1}^{N_c} \big\{ a_\beta(\theta)^\top (p_t^j - p_t^i) \leq b_\beta(\theta) \big\}.
\end{align}
where, for each $\beta \in [N_c]$, the terms $a_\beta(\theta)$ and $b_\beta(\theta)$ characterize one face of the polytope. As a special case, box-shaped collision avoidance constraints can formulated as polytopic collision avoidance constraints of the form \eqref{Eqn: App, Polytopic Collision-Avoidance Constraints, def}, with $N_c = 4$ and $\{a_\beta(\theta): \beta \in [4] \}$ chosen so that the resulting constraint sets have boundaries that are pairwise either parallel or orthogonal.

\paragraph{Velocity-Dependent Spherical Collision Avoidance Constraints}

By a \textit{velocity-dependent spherical collision-avoidance constraint} enforced on agent $i$ to prevent collisions with agent $j$, we refer to a spherical collision-avoidance constraint \eqref{Eqn: App, Spherical Collision-Avoidance Constraints, def} whose center is shifted from agent $i$'s position towards agent $j$'s position whenever agent $j$ is moving towards agent $i$.
More specifically, a velocity-dependent spherical collision-avoidance constraint on agent $i$ enforces that, at each time $t \in [T]$, the vector $p_t^j - p_t^i$, with $\theta_6^i > 0$, lies outside of a sphere centered at $- \theta_6^i (v_t^j - v_t^i)$, rather than at the origin.
Concretely, for agents with following the modified unicycle dynamics \eqref{Eqn: App, Unicycle Dynamics, with velocity vector} with states $\bar x = (p_x, p_y, v_x, v_y) = (p, v)$, velocity-dependent collision-avoidance constraints assume the following form:
\begin{align} \label{Eqn: App, Velocity-Dependent Spherical Collision-Avoidance Constraints, def}
    &\hspace{5mm} C_\urk^{\ineq, i} \\ \nonumber
    &= \bigwedge_{t=1}^T \bigwedge_{j \in [N] \backslash \{i\}} \\ \nonumber
    &\hspace{1cm} \Big\{ - \Big\Vert (p_t^j - p_t^i) + \theta_6^i (v_t^j - v_t^i) \Big\Vert_2^2 + \theta_1^i \leq 0 \Big\} \\ \nonumber
    &= \bigwedge_{t=1}^T \bigwedge_{j \in [N] \backslash \{i\}} \left\{ - (\bar x^j - \bar x^i)^\top M_{VDS}
    (\bar x^j - \bar x^i) + \theta_1^i \leq 0 \right\}.
\end{align}
where:
\begin{align}
    M_{VDS} &:= \begin{bmatrix}
        1 & 0 & \theta_6^i & 0 \\
        0 & 1 & 0 & \theta_6^i \\
        \theta_6^i & 0 & (\theta_6^i)^2 & 0 \\
        0 & \theta_6^i & 0 & (\theta_6^i)^2
    \end{bmatrix}
\end{align}
Although the constraints \eqref{Eqn: App, Velocity-Dependent Spherical Collision-Avoidance Constraints, def} are not affine in $\theta_6^i$, we can introduce additional variables appropriately to render them affine in the resulting, expanded set of constraint parameters. Finally, we note that if $\theta_6^i = 0$, the velocity-dependent spherical collision avoidance constraint reduces to the (velocity-independent) spherical collision-avoidance constraint originally defined in \eqref{Eqn: App, Spherical Collision-Avoidance Constraints, def}.

The additional term $\theta_6^i (v_t^j - v_t^i)$ compels each agent $i$ to modulate its sensitivity to the presence of any other agent $j$ based on agent $j$'s speed and direction of movement relative to that of agent $i$. 
Intuitively, if agent $j$ begins to move towards agent $i$, then a velocity-dependent collision avoidance constraint with $\theta_6^i > 0$ would compel agent $i$ to increase the minimum distance it needs to maintain from agent $j$.
In practice, both velocity-dependent and velocity-independent spherical collision-avoidance constraints should be implemented in a multi-agent system, to ensure that each agent accounts for the movements of other agents when reasoning about collision avoidance, while also enforcing a baseline level of collision avoidance against other agents regardless of their relative velocities.

\paragraph{Line-of-Sight Constraints}
By a \textit{line-of-sight constraint on agent $i$ with respect to agent $j$}, we refer to a constraint which compels agent $i$ to ensure that 
the angle between the vectors $v^i$ and $p^j - p^i$ is sufficiently small. Conceptually speaking, agent $i$ must constantly travel in a direction (indicated by $v^i$) that is sufficiently well-aligned with the location of agent $j$ relative to itself (as given by $p^j - p^i$). As noted in Sec. \ref{subsec: Constraint Parameterizations and Agent Costs}, line-of-sight constraints have practical significance in the context of herding, pursuit-evasion games, and other multi-agent interaction scenarios. Below, we concretely formulate line-of-sight constraints for 2-dimensional agents (e.g., ground robots) with modified unicycle dynamics, as given in \eqref{Eqn: App, Unicycle Dynamics, with velocity vector}.

\begin{figure}
    \centering
    \includegraphics[width=0.95\linewidth]{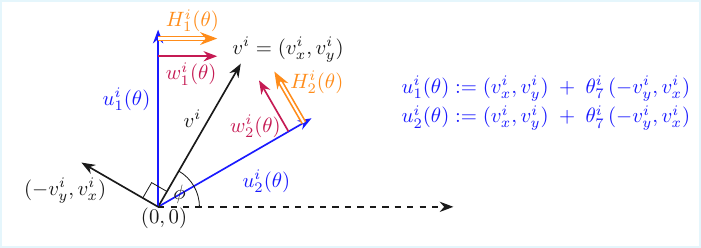}
    \caption{Schematic for explaining the formulation of the line-of-sight constraints. In particular, we begin by defining \textcolor{blue}{$u_1(\theta)$} and \textcolor{blue}{$u_2(\theta)$} as vectors that are displaced from $v^i$, the velocity vector of agent $i$, in the orthogonal direction given by $(-v_y^i, v_x^i)$. Here, $\theta$ captures the degree to which $u_1(\theta)$ and $u_2(\theta)$ differ from $v^i$. Then, we define $w_1^i(\theta)$ and $w_2^i(\theta)$ to be vectors orthogonal to $u_1^i(\theta)$ and $u_2^i(\theta)$, respectively, that both point ``into'' the cone bounded by $u_1^i(\theta)$ and $u_2^i(\theta)$. Finally, we define $H_1^i(\theta)$ and $H_2^i(\theta)$ as the half-spaces with boundaries given by $u_1^i(\theta)$ and $u_2^i(\theta)$, by using $w_1^i(\theta)$ and $w_2^i(\theta)$ as the normal vectors of $H_1^i(\theta)$ and $H_2^i(\theta)$, respectively. Then, our desired cone $\C^i$ can be given as the intersection of $H_1^i(\theta)$ and $H_2^i(\theta)$, i.e., $\C^i(\theta) H_1^i(\theta) \bigcap H_2^i(\theta).$
    }
     \label{fig:line_of_sight_constraint_expl}
\end{figure}

For each agent $i \in [N]$, we wish to constrain the position of any other agent $j$ relative to agent $i$, i.e., $p_t^j - p_t^i \in \R^2$, to lie within a cone-shaped area $\C^i$ whose vertex lies at the origin $(0, 0)$, and whose axis is given by $v^i = (v_x^i, v_y^i) \in \R^2$, the velocity vector of agent $i$ (see Fig. \ref{fig:line_of_sight_constraint_expl}). To characterize the two boundaries of the cone, we displace the velocity vector $v^i = (v_x^i, v_y^i)$ in the direction of $(-v_y^i, v_x^i)$, a vector perpendicular to $v^i$. Specifically, we define the two vectors which form the boundary of the cone, denoted $u_1^i, u_2^i \in \R^2$ below, as follows:
\begin{align}
    u_1^i(\theta) &:= (v_x^i, v_y^i) + \theta_7^i (-v_y^i, v_x^i), \\
    u_2^i(\theta) &:= (v_x^i, v_y^i) - \theta_7^i (-v_y^i, v_x^i),
\end{align}
where the unknown parameters $\theta = (\theta_7^i: i \in [N]) \in \R^N$, with $\theta_7^i > 0$ for each $i \in [N]$, describe the width of the cone $\C^i$, for each $i \in [N]$. Next, to constrain each relative distance vector $p_t^j - p_t^i$ to lie within $\C^i$, we will define halfspaces $H_1(\theta)$ and $H_2(\theta)$ whose boundaries are given by $u_1^i(\theta)$ and $u_2^i(\theta)$, respectively. Let $w_1^i(\theta) \in \R^2$ and $w_2^i(\theta) \in \R^2$ be vectors perpendicular to $u_1^i(\theta)$ and $u_2^i(\theta)$, respectively which point \say{inward} into the cone $\C^i$, as given below:
\begin{align}
    w_1^i(\theta) &= \theta_7^i(v_x^i, v_y^i) - (-v_y^i, v_x^i), \\
    w_2^i(\theta) &= \theta_7^i(v_x^i, v_y^i) + (-v_y^i, v_x^i).
\end{align}
Then the constraint set corresponding to agent $i$'s line-of-sight constraint can be written as:
{
\small
\begin{align} \label{Eqn: App, Line-of-Sight Constraints, def}
    &\bigwedge_{t=1} \bigwedge_{j \in [N] \backslash \{i\}} \Big\{ p_t^j - p_t^i \in H_1(\theta) \bigcap H_2(\theta) \Big\} \\ \nonumber
    = \ &\bigwedge_{t=1} \bigwedge_{j \in [N] \backslash \{i\}} \Big\{ (p_t^j - p_t^i)^\top (v_y + v_x \theta_7^i, -v_x + v_y \theta) \geq 0, \\ \nonumber
    &\hspace{2.5cm} (p_t^j - p_t^i)^\top ( -v_y + v_x \theta, v_x + v_y \theta_7^i) \geq 0 \Big\} \\ \nonumber
    = \ &\bigwedge_{t=1} \bigwedge_{j \in [N] \backslash \{i\}} \\ \nonumber
    &\hspace{5mm} \Big\{ (p_{x,t}^j - p_{x,t}^i) v_{y,t}^i - (p_{y,t}^j - p_{y,t}^i) v_{x,t}^i \\ \nonumber
    &\hspace{1.5cm} + \big[ (p_{x,t}^j - p_{x,t}^i) v_{x,t}^i + (p_{y,t}^j - p_{y,t}^i) v_{y,t}^i \big] \theta_7^i \geq 0, \\ \nonumber
    &\hspace{1cm} 
    -(p_{x,t}^j - p_{x,t}^i) v_{y,t}^i + (p_{y,t}^j - p_{y,t}^i) v_{x,t}^i \\ \nonumber
    &\hspace{1.5cm} + \big[ (p_{x,t}^j - p_{x,t}^i) v_{x,t}^i + (p_{y,t}^j - p_{y,t}^i) v_{y,t}^i \big] \theta_7^i \geq 0.
    \Big\}.
\end{align}
}
Note that the constraints formulated above in \eqref{Eqn: App, Line-of-Sight Constraints, def} are affine in the parameter $\theta$, despite exhibiting nonlinearities (more precisely, bilinearities), in the primal variables describing positions and velocities.

\paragraph{Origin and Goal Constraints}

As stated in Sec. \ref{subsec: Constraint Parameterizations and Agent Costs}, each agent is prescribed a given, \textit{known} set of origin and goal position constraints, given by:
\begin{align} \label{Eqn: App, Start and Goal Constraints}
    \bigwedge_{i \in [N]} \{ p_0^i = \bar p_o^i, p_T^i = \bar p_d^i \},
\end{align}
where $\bar p_o^i$ and $\bar p_d^i$ denote the pre-determined origin and goal positions, respectively.

For experiments which involve line-of-sight constraints or velocity-based collision-avoidance constraints, we may wish to constrain the agents' initial and final velocities in addition to their start and goal positions, in which case \eqref{Eqn: App, Start and Goal Constraints} becomes:
\begin{align} \label{Eqn: App, Start, Goal, and Velocity Constraints}
    \bigwedge_{i \in [N]} \{ p_0^i = \bar p_o^i, v_0^i = \bar v_o^i, p_T^i = \bar p_d^i, v_T^i = \bar v_d^i \}.
\end{align}

\subsubsection{Double Integrator Simulations}
\label{subsubsec: App, Double Integrator Simulations}

As described in Sec. \ref{subsec: Double Integrator Simulations}, we evaluate our method on simulated interactions between agents with double integrator dynamics and either elliptic or polytopic collision-avoidance constraints, as described by \eqref{Eqn: App, Elliptic Collision-Avoidance Constraints, Ground robot, def} and \eqref{Eqn: App, Polytopic Collision-Avoidance Constraints, def}. We also report the Gurobi solve times, for learning spherical collision-avoidance constraints, as a function of the number of interacting agents. Below, we provide the ground truth parameters and other details of the aforementioned experiments that were omitted in the main text.

\paragraph{Elliptic Constraints}

We generate an interaction demonstration from which we learn the elliptic collision-avoidance constraint sets of two agents with double integrator dynamics (Fig. \ref{fig:DI_poly_ell}a). 
Specifically, we discretize the continuous-time double integrator dynamics at intervals of $\Delta t = 1$, and set the time horizon to be $T = 10$. We define the cost of each agent to encode the smoothness objective formulated in Sec. \ref{subsec: Constraint Parameterizations and Agent Costs}, i.e.,
\begin{align} \label{Eqn: App, Smoothness Cost, def}
    J^i &= \sum_{t=1}^{T-1} \Vert p_{t+1}^i - p_t^i \Vert_2^2, \hspace{1cm} \forall \ i \in \{1, 2\}.
\end{align}
Elliptical collision-avoidance constraints of the form \eqref{Eqn: App, Elliptic Collision-Avoidance Constraints, Ground robot, def} with the following ground truth parameters, which are \textit{a priori} unknown to the constraint learner, are enforced on each agent's trajectory:
\begin{align} \nonumber
    (\theta_1^i, \theta_2^i, \theta_3^i) &= (49, 4, 1), \hspace{1cm} \forall \ i \in \{1, 2\}.
\end{align}
The start and goal positions of each agent $i \in [2]$ are constrained via known equality constraints of the form \eqref{Eqn: App, Start and Goal Constraints} to be at $\bar p_o^i$ and $\bar p_d^i$, respectively, as given below:
\begin{align} \nonumber
    &\bar p_o^1 = (0.0, 0.0), \hspace{5mm} \bar p_d^1 = (10.0, 10.0), \\ \nonumber
     &\bar p_o^2 = (10.0, 10.0), \hspace{5mm} \bar p_d^2 = (0.0, 0.0).
\end{align}
We successfully recovered the \textit{a priori} unknown constraint parameters with a Gurobi solve time of 0.2 seconds.

\paragraph{Polytopic Collision-Avoidance Constraints}

We generate two interaction demonstrations from which we learn the polytopic collision-avoidance constraint sets of two agents with double integrator dynamics (Fig. \ref{fig:DI_poly_ell}b, \ref{fig:DI_poly_ell}c). 
The two demonstrations share all of the following parameter settings \textit{except} agents' initial and final goal positions. 
Specifically, we discretize the continuous-time double integrator dynamics at intervals of $\Delta t = 1$, and set the time horizon to be $T = 20$. We define the cost of each agent to encode the smoothness objective formulated in Sec. \ref{subsec: Constraint Parameterizations and Agent Costs}, i.e.,
\begin{align} \nonumber
    J^i &= \sum_{t=1}^{T-1} \Vert p_{t+1}^i - p_t^i \Vert_2^2, \hspace{1cm} \forall \ i \in \{1, 2\}.
\end{align}
Polytopic collision-avoidance constraints of the form \eqref{Eqn: App, Polytopic Collision-Avoidance Constraints, def} with the ground truth parameters $\{a_\beta(\theta) \in \R^2, b_\beta(\theta) \in \R : \beta \in [4] \}$ defined below, which are \textit{a priori} unknown to the constraint learner, are enforced on each agent's trajectory. For each $\beta \in [4]$, we set $a_\beta(\theta) \in \R^2$ to be the transpose of the $\beta$-th row of the matrix $A(\theta)$ defined below, and we set $b_\beta(\theta) \in \R$ to be the $\beta$-th entry of the vector $b(\theta)$ defined below:
\begin{align} \nonumber
    A(\theta) &= 
    \begin{bmatrix}
    -0.2545 & -0.9671 \\
     0.9487 & -0.3162 \\
     0.2169 &  0.9762 \\
    -0.9285 &  0.3714
    \end{bmatrix}
    , \hspace{5mm} b(\theta) = 
    \begin{bmatrix}
    10.0779 \\
     9.4868 \\
    11.6058 \\
     9.4705
    \end{bmatrix}.
\end{align}
The start and goal positions of each agent $i \in [2]$ are constrained via known equality constraints of the form \eqref{Eqn: App, Start and Goal Constraints} to be at $\bar p_o^i$ and $\bar p_d^i$, respectively, as given below:
\begin{align} \nonumber
     \Demo \ 1: \hspace{3mm} &\bar p_o^1 = (-2.0, 0.0), \hspace{3mm} \bar p_d^1 = (9.5, 9.5), \\ \nonumber
     &\bar p_o^2 = (9.5, 9.5), \hspace{3mm} \bar p_d^2 = (-2, 9.5), \\ \nonumber
     \Demo \ 2: \hspace{3mm} &\bar p_o^1 = (15.0, 0.0), \hspace{3mm} \bar p_d^1 = (15.0, 15.0), \\ \nonumber
     &\bar p_o^2 = (15.0, 15.0), \hspace{3mm} \bar p_d^2 = (15.0, 0.0).
\end{align}
We successfully recovered the \textit{a priori} unknown constraint parameters with a Gurobi solve time of 54.86 seconds. 
The learned constraints were then used to generate safe motion plans in simulation, in conjunction with the agent costs, dynamics, time discretization, and time horizon given above, as well as equality constraints of the form \eqref{Eqn: App, Start and Goal Constraints} to enforce the following start positions $\bar p_o^i$ and goal positions $\bar p_d^i$ for each agent $i$:
\begin{align} \nonumber
    &\bar p_o^1 = (15.0, 0.0), \hspace{5mm} \bar p_d^1 = (0.0, 15.0), \\ \nonumber
    &\bar p_o^2 = (15.0, 15.0), \hspace{5mm} \bar p_d^2 = (0.0, 0.0).
\end{align}
The generated safe motion plans are plotted in Fig. \ref{fig:DI_poly_ell}c.

\paragraph{Velocity-Dependent Spherical Collision Avoidance Constraints}

We generate an interaction demonstration from which we learn the \textit{velocity-dependent} spherical collision-avoidance constraint sets, of the form \eqref{Eqn: App, Velocity-Dependent Spherical Collision-Avoidance Constraints, def}, of two agents with double integrator dynamics (Fig. \ref{fig:DI_poly_ell}e, \ref{fig:DI_poly_ell}f, and \ref{fig:DI_poly_ell}g). 
Specifically, we discretize the continuous-time double integrator dynamics at intervals of $\Delta t = 0.5$, and set the time horizon to be $T = 20$. We define the cost of each agent as follows:
\begin{align} \nonumber
    J^i = \sum_{t=1}^T \sum_{j=1}^2 \Big( \Vert p_{t+1}^j - p_t^j \Vert_2^2 + \Vert u_t^j \Vert_2^2 \Big).
\end{align}
\textit{Velocity-dependent} spherical collision-avoidance constraints of the form \eqref{Eqn: App, Velocity-Dependent Spherical Collision-Avoidance Constraints, def}, with the following \textit{a priori} unknown ground truth parameters, are enforced on the agents' trajectories:
\begin{align} \nonumber
    \theta_1^i = 2.0, \hspace{5mm} \theta_6^i = 1.0, \hspace{5mm} \forall \ i \in \{1, 2\}.
\end{align}
The start position, initial velocity, goal position, and final velocity of each agent $i \in [3]$ are constrained via known equality constraints of the form \eqref{Eqn: App, Start, Goal, and Velocity Constraints} to be at $\bar p_o^i$, $v_i^i$, $\bar p_d^i$, and $\bar v_d^i$, respectively, as given below:
\begin{alignat}{2} \nonumber
     &\bar p_o^1 = (-3.0, 0.0), \hspace{3mm} &&\bar v_o^1 = (0.715, 0.145), \\ \nonumber
     &\bar p_d^1 = (3.0, 3.0), &&\bar v_d^1 = (0.546, 0.483), \\ \nonumber
     &\bar p_o^2 = (3.0, 3.0), &&\bar v_o^2 = (-0.715, -0.145), \\ \nonumber
     &\bar p_d^2 = (-3.0, 0.0), &&\bar v_d^2 = (-0.546, -0.483).
\end{alignat}
We successfully recovered the \textit{a priori} unknown constraint parameters with a Gurobi solve time of 0.12 seconds.
The learned constraints were then used to generate two safe motion plans in simulation, in conjunction with the agent costs, dynamics, time discretization, and time horizon given above, but different start positions $\bar p_o^i$, initial velocities $\bar v_o^i$, goal positions $\bar p_d^i$, and final velocities $\bar v_d^i$ for each agent $i$, as enforced by equality constraints of the form \eqref{Eqn: App, Start, Goal, and Velocity Constraints}. In particular, for the first motion plan, as plotted in Fig. \ref{fig:unicycle}f, we set:
\begin{alignat}{2} \nonumber
    &\bar p_o^1 = (2.0, 0.0), \hspace{5mm} &&\bar v_o^1 = (-0.622, 0.179), \\ \nonumber
    &\bar p_d^1 = (-3.0, 3.0), &&\bar v_d^1 = (-0.439, 0.439), \\ \nonumber
    &\bar p_o^2 = (-3.0, 3.0), &&\bar v_o^2 = (0.727, -0.179), \\ \nonumber
    &\bar p_d^2 = (3.0, 0.0), &&\bar v_d^2 = (0.544, -0.439).
\end{alignat}
For the second motion plan, as plotted in Fig. \ref{fig:unicycle}g, we set:
\begin{alignat}{2} \nonumber
    &\bar p_o^1 = (2.0, 0.0), \hspace{5mm} &&\bar v_o^1 = (-0.526, 0.316), \\ \nonumber
    &\bar p_d^1 = (-3.0, 3.0), &&\bar v_d^1 = (-0.526, 0.316), \\ \nonumber
    &\bar p_o^2 = (2.0, 3.0), &&\bar v_o^2 = (-0.526, 0.316), \\ \nonumber
    &\bar p_d^2 = (-3.0, 6.0), &&\bar v_d^2 = (-0.526, 0.316).
\end{alignat}

\paragraph{Robustness of the Volume Extraction Approach Relative to Inferred Uncertainty}

\begin{figure}
    \centering
    \includegraphics[width=0.90\linewidth]{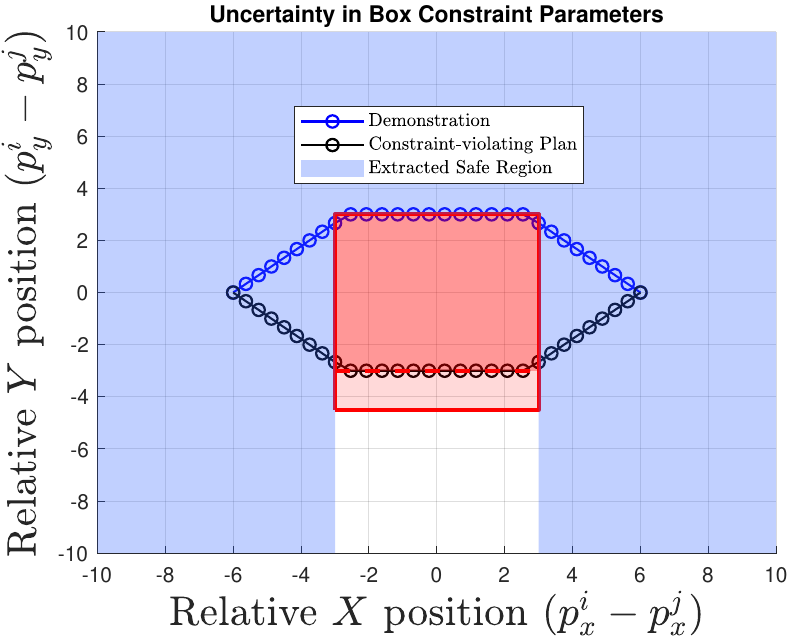}
    \caption{An example in which a single demonstration (blue, with circles) allows the recovery of the left, top, and right edges of the true, unknown 2D box collision-avoidance constraint set (shaded dark and light red, bordered by solid red lines), but not its bottom edge, $y = -4.5$. Using volume extraction, we compute the guaranteed safe region (shaded blue), which is then used to design guaranteed safe motion plans which, in this instance, coincide with the blue demonstration.
    However, a naive point estimate of the constraint's bottom edge returns $y = -3$, which underestimates the extent to which the collision avoidance set extends downwards. This results in an under-approximation (shaded dark red) of the true collision avoidance set. The motion plan generated as a result (black) is constraint-violating.
    }
    \label{fig:rebuttal___volume_extraction_robustness}
\end{figure}

Below, we present an experiment which illustrates that, in terms of robustly guaranteeing safety despite incomplete recovery of the true constraints, volume extraction-informed motion plans outperform motion plans designed using a single point estimate of constraint parameter. Specifically, Fig. \ref{fig:rebuttal___volume_extraction_robustness} illustrates a 2-agent interaction demonstration (blue) which activates all but the bottom edge of a box-shaped constraint. We generated (1) motion plans naively generated using an incorrect, single point estimate of the position of the bottom edge of the constraint (black), as well as (2) motion plans generated by leveraging conservative estimates of the bottom edge using volume extraction (coincides with the blue interaction demonstration). Whereas our volume-extraction based approach informed the successful design of interactive trajectories that obeyed all underlying, true constraints, the naive motion plan, generated using the incorrect point estimate of the location of the box constraint's bottom edge, violated the box constraint.

\paragraph{Gurobi Solve Times vs. Number of Interacting Agents}

\begin{figure}[ht]
    \centering
    \includegraphics[width=0.90\linewidth]{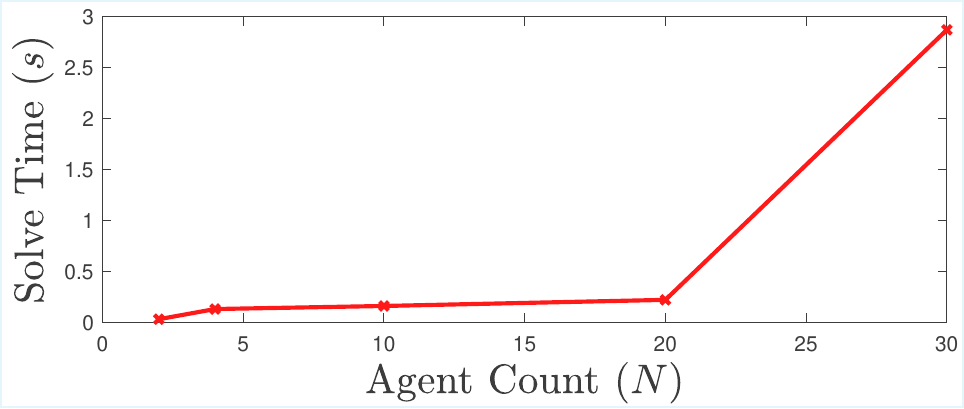}
    \caption{Gurobi solve times for our constraint learning method, when applied to simulations involving $N = 2, 4, 10, 20, 30$ agents with double integrator dynamics and spherical collision-avoidance constraint sets. When $N \leq 20$, our method easily recovers the constraints of all agents in less than $0.25$ s. When $N = 30$, our method can still recover the constraints of all agents in roughly 3s.}
    \label{fig:scaling_computation_time}
\end{figure}

To illustrate the computational tractability of our constraint learning method, we record Gurobi solve times for multi-agent constraint learning problems involving different numbers of interacting agents, namely $N = 2, 4, 10, 20, 30$ (Fig. \ref{fig:scaling_computation_time}). In each simulation, we assume all agents have double integrator dynamics discretized at intervals of $\Delta t = 1$, and we set the time horizon to be $T = 20$. We define the cost of each agent to encode the smoothness objective formulated in Sec. \ref{subsec: Constraint Parameterizations and Agent Costs}, i.e.,
\begin{align} \nonumber
    J^i &= \sum_{t=1}^{T-1} \Vert p_{t+1}^i - p_t^i \Vert_2^2, \hspace{1cm} \forall \ i \in \{1, 2\}.
\end{align}
Moreover, when inferring \textit{box} collision avoidance constraints from agent demonstrations, the 2-, 4-, and 10-agent problem instances required Gurobi solve times of 0.06 s, 0.59 s, and 5.37 s, respectively.

Spherical collision-avoidance constraints of the form \eqref{Eqn: App, Spherical Collision-Avoidance Constraints, def} with the following ground truth parameters are enforced on the trajectories of all agents:
\begin{align}
    \theta_1^i = i, \hspace{5mm} \forall \ i \in [N]
\end{align}
The start and goal positions $\{\bar p_o^i, \bar p_d^i: i \in [N] \}$ of the interacting agents are identically and independently drawn from a circle of radius 200 centered at the origin, and subsequently encoded into known equality constraints of the form \eqref{Eqn: App, Start and Goal Constraints}.


\subsubsection{Unicycle Simulations}
\label{subsubsec: App, Unicycle Dynamics Simulations}

We also evaluate our method on simulated interactions between agents with unicycle dynamics and either spherical collision-avoidance and proximity constraints, as described by \eqref{Eqn: App, Spherical Proximity Constraints, def}, or line-of-sight constraints, as described by \eqref{Eqn: App, Line-of-Sight Constraints, def}. 

\paragraph{Spherical Proximity and Collision-Avoidance Constraints}

We generate an interaction demonstration from which we learn the spherical proximity and collision-avoidance constraint sets, as given by \eqref{Eqn: App, Spherical Proximity Constraints, def}, of two agents with double integrator dynamics (Fig. \ref{fig:unicycle}a, \ref{fig:unicycle}b). 
Specifically, we discretize the continuous-time double integrator dynamics at intervals of $\Delta t = 0.5$, and set the time horizon to be $T = 10$. We define the cost of each agent as follows:
\begin{align} \nonumber
    J^i = \sum_{t=1}^T \sum_{j=1}^2 \Big( \Vert p_{t+1}^j - p_t^j \Vert_2^2 + \Vert u_t^j \Vert_2^2 \Big) + 5 \cdot \sum_{t=1}^T p_{y,t}^2.
\end{align}


Spherical proximity and collision-avoidance constraints of the form \eqref{Eqn: App, Spherical Proximity Constraints, def}, with the following \textit{a priori unknown} ground truth parameters, are enforced on each agent's trajectory:
\begin{align} \nonumber
    (\theta_1^i, \theta_5^i) = (1.0, 4.0), \hspace{5mm} \forall \ i \in \{1, 2\}.
\end{align}
The start position, initial velocity, goal position, and final velocity of each agent $i \in [N]$ are constrained via known equality constraints of the form \eqref{Eqn: App, Start, Goal, and Velocity Constraints} to be at $\bar p_o^i$, $v_o^i$, $\bar p_d^i$, and $\bar v_d^i$ respectively, as given below:
\begin{alignat}{2} \nonumber
     &\bar p_o^1 = (-3.0, 0.0), \hspace{1cm} &&\bar v_o^1 = (0.0, 0.1), \\ \nonumber
     &\bar p_d^1 = (3.0, 3.0), &&\bar v_o^1 = (0.0, 0.1), \\ \nonumber
     &\bar p_o^2 = (-1.95, -0.05), &&\bar v_o^2 = (2.35, 0.1), \\ \nonumber
     &\bar p_d^2 = (2.95, 1.0), &&\bar v_o^2 = (0.77, 1.77).
\end{alignat}
We successfully recovered the \textit{a priori} unknown constraint parameters with a Gurobi solve time of 0.02 seconds.
The learned constraints were then used to generate safe motion plans in simulation, in conjunction with the agent costs, dynamics, time discretization, and time horizon given above, as well as equality constraints of the form \eqref{Eqn: App, Start and Goal Constraints} to enforce the following start positions $\bar p_o^i$ and goal positions $\bar p_d^i$ for each agent $i$:
\begin{alignat}{2} \nonumber
    &\bar p_o^1 = (-3.0, 0.0), \hspace{1cm} &&\bar v_o^1 = (0.0, 0.1), \\ \nonumber
    &\bar p_d^1 = (1.0, 2.0), &&\bar v_d^1 = (2.356, 0.1), \\ \nonumber
    &\bar p_o^2 = (-1.95, -0.55), &&\bar v_o^2 = (2.356, 0.1), \\ \nonumber
    &\bar p_d^2 = (1.129, 0.004), &&\bar v_d^2 = (0.685, 1.349).
\end{alignat}
The generated safe motion plans are plotted in Fig. \ref{fig:unicycle}b.

\paragraph{Line-of-Sight Constraints}

We generate an interaction demonstration from which we learn the line-of-sight constraints of two agents with unicycle dynamics (Fig. \ref{fig:unicycle}c, \ref{fig:unicycle}d). 
Specifically, we discretize the (modified) continuous-time unicycle dynamics \eqref{Eqn: App, Unicycle Dynamics, with velocity vector} at intervals of $\Delta t = 0.3$, and set the time horizon to be $T = 20$. We define the cost of each agent to be as follows: 
\begin{align} \nonumber
    J^i = \sum_{t=1}^T \sum_{j=1}^2 \Big( \Vert p_{t+1}^j - p_t^j \Vert_2^2 + \Vert u_t^j \Vert_2^2 \Big).
\end{align}
The following constraints, which are \textit{a priori} unknown to the constraint learner, are enforced on each agent's trajectory---Spherical proximity constraints of the form \eqref{Eqn: App, Spherical Proximity Constraints, def}, with ground truth parameters:
\begin{align} \nonumber
    (\theta_1^i, \theta_5^i) &= (1.0, 16.0), \hspace{5mm} \forall \ i \in \{1, 2\}, 
\end{align}
and line-of-sight constraints of the form \eqref{Eqn: App, Line-of-Sight Constraints, def}, with ground truth parameters: 
\begin{align} \nonumber
    \theta_7^i = 0.1, \hspace{5mm} \forall \ i \in \{1, 2\}, 
\end{align}
The start position, initial velocity, goal position, and final velocity of each agent $i \in [N]$ are constrained via known equality constraints of the form \eqref{Eqn: App, Start, Goal, and Velocity Constraints} to be at $\bar p_o^i$, $\bar v_o^i$, $\bar p_d^i$, and $\bar v_d^i$, respectively, as given below:
\begin{alignat}{2} \nonumber
    &\bar p_o^1 = (-3.0, 0.0), \hspace{1cm} &&\bar v_o^1 = (0.270, 0.0), \\ \nonumber
    &\bar p_d^1 = (3.0, 3.0), &&\bar v_d^1 = (0.855, 0.0), \\ \nonumber
    &\bar p_o^2 = (-4.1, 0.0), &&\bar v_o^2 = (0.438, 0.0), \\ \nonumber
    &\bar p_d^2 = (1.413, 2.245), &&\bar v_d^2 = (1.056, 0.576).
\end{alignat}
The Gurobi solve time was 0.09 seconds.
The learned constraints were then used to generate safe motion plans in simulation, in conjunction with the agent costs, dynamics, time discretization, and time horizon given above, as well as equality constraints of the form \eqref{Eqn: App, Start and Goal Constraints} to enforce the following start positions $\bar p_o^i$ and goal positions $\bar p_d^i$ for each agent $i$:
\begin{alignat}{2} \nonumber
    &\bar p_o^1 = (0.0, 3.0), \hspace{1cm} &&\bar v_o^1 = (0.293, 0.0), \\ \nonumber
    &\bar p_d^1 = (2.001, 0.013), &&\bar v_d^1 = (0.007, -0.565), \\ \nonumber
    &\bar p_o^2 = (-1.2, 3.0), &&\bar v_o^2 = (0.210, 0.0) \\ \nonumber &\bar p_d^2 = (0.985, 1.735), &&\bar v_d^2 = (-0.492, 0.238).
\end{alignat}
The generated safe motion plans are plotted in Fig. \ref{fig:unicycle}d.

\subsubsection{Quadcopter Simulations}
\label{subsubsec: App, Quadcopter Simulations}

As described in Sec. \ref{subsec: Quadcopter Simulations}, we evaluate our method on simulated interactions between agents with 12-D quadcopter dynamics and either elliptic or polytopic (in particular, box-shaped) collision-avoidance constraints, as described by \eqref{Eqn: App, Elliptic Collision-Avoidance Constraints, Ground robot, def} and \eqref{Eqn: App, Polytopic Collision-Avoidance Constraints, def}. We also report the Gurobi solve times, for learning spherical collision-avoidance constraints, as a function of the number of interacting agents. Below, we provide the ground truth parameters and other details of the aforementioned experiments that were omitted in the main text.

\paragraph{Spherical Collision-Avoidance Constraints}

We generate an interaction demonstration from which we learn the spherical collision-avoidance constraint sets of three agents with quadcopter dynamics (Fig. \ref{fig:quad_sphere}a and \ref{fig:quad_sphere}b). 
Specifically, we discretize the continuous-time quadcopter dynamics \eqref{Eqn: App, Quadcopter Dynamics} at intervals of $\Delta t = 1$, and set the time horizon to be $T = 20$. We define the cost of each agent to encode the smoothness objective formulated in Sec. \ref{subsec: Constraint Parameterizations and Agent Costs}, i.e., 
\begin{align} \nonumber
    J^i &= \sum_{t=1}^{T-1} \Vert p_{t+1}^i - p_t^i \Vert_2^2, \hspace{1cm} \forall \ i \in \{1, 2\}.
\end{align}
Spherical collision-avoidance constraints of the form \eqref{Eqn: App, Spherical Collision-Avoidance Constraints, def}, with the following \textit{a priori} unknown ground truth parameters, are enforced on the agents' trajectories:
\begin{align} \nonumber
    \theta_1^1 = 6, \hspace{5mm} \theta_1^2 = 8, \hspace{5mm} \theta_1^3 = 9.
\end{align}
The start and goal positions of each agent $i \in [3]$ are constrained via known equality constraints of the form \eqref{Eqn: App, Start and Goal Constraints} to be at $\bar p_o^i$ and $\bar p_d^i$, respectively, as given below:
\begin{alignat}{2} \nonumber
     &\bar p_o^1 = (0.0, 0.0, 0.0), \hspace{1cm} &&\bar p_d^1 = (20.0, 20.0, 20.0), \\ \nonumber
     &\bar p_o^2 = (20.0, 20.0, 20.0), &&\bar p_d^2 = (0.0, 0.0, 0.0), \\ \nonumber
     &\bar p_o^3 = (0.0, 20.0, 0.0), &&\bar p_d^3 = (20.0, 0.0, 20.0).
\end{alignat}
We successfully recovered the \textit{a priori} unknown constraint parameters with a Gurobi solve time of 0.2 seconds.
The learned constraints were then used to generate safe motion plans in simulation, in conjunction with the agent costs, dynamics, time discretization, and time horizon given above, as well as equality constraints of the form \eqref{Eqn: App, Start and Goal Constraints} to enforce the following start positions $\bar p_o^i$ and goal positions $\bar p_d^i$ for each agent $i$:
\begin{alignat}{2} \nonumber
    &\bar p_o^1 = (0.0, 0.0. 0.0), \hspace{1cm} &&\bar p_d^1 = (10.0, 10.0, 20.0), \\ \nonumber
    &\bar p_o^2 = (10.0, 10.0, 10.0), &&\bar p_d^2 = (0.0, 0.0, 0.0), \\ \nonumber
    &\bar p_o^3 = (0.0, 10.0, 0.0), &&\bar p_d^3 = (20.0, 0.0, 20.0).
\end{alignat}
The generated safe motion plans are plotted in Fig. \ref{fig:quad_sphere}b.


\paragraph{Box-Shaped Collision-Avoidance Constraints}

\begin{figure}[ht]
    \centering
    \includegraphics[width=\linewidth]{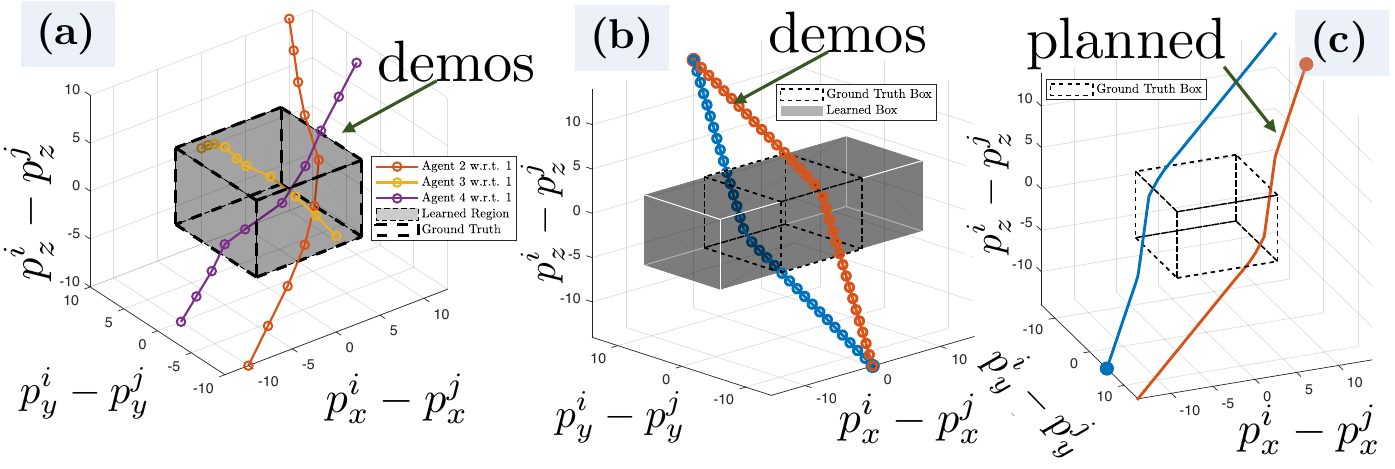}
    \caption{Constraint learning for quadcopter agents with box constraints, from a demonstration with (a) 4 agents, from which the constraint was recovered, or (b) 2 agents, which fails to activate some constraints that were thus not learned. (c) Regardless, our method generated safe plans via volume extraction over trajectory space. 
    }
    \label{fig:quad_box}
\end{figure}

We generate an interaction demonstration from which we learn the polytopic collision-avoidance constraint sets of four agents with quadcopter dynamics (Fig. \ref{fig:quad_box}b, \ref{fig:quad_box}c). 
Specifically, we discretize the continuous-time quadcopter dynamics at intervals of $\Delta t = 1$, and set the time horizon to be $T = 10$. We define the cost of each agent to encode the smoothness objective formulated in Sec. \ref{subsec: Constraint Parameterizations and Agent Costs}, i.e.,
\begin{align} \nonumber
    J^i &= \sum_{t=1}^{T-1} \Vert p_{t+1}^i - p_t^i \Vert_2^2, \hspace{1cm} \forall \ i \in \{1, 2\}.
\end{align}
Polytopic collision-avoidance constraints of the form \eqref{Eqn: App, Polytopic Collision-Avoidance Constraints, def} with the \textit{a priori} unknown ground truth parameters $\{a_\beta(\theta) \in \R^2, b_\beta(\theta) \in \R : \beta \in [4] \}$ defined below, are enforced on each agent's trajectory. For each $\beta \in [6]$, we set $a_\beta(\theta) \in \R^2$ to be the transpose of the $\beta$-th row of the matrix $A(\theta)$ defined below, and we set $b_\beta(\theta) \in \R$ to be the $\beta$-th entry of the vector $b(\theta)$ defined below:
\begin{align} \nonumber
    A(\theta) &= 
    \begin{bmatrix}
        1.0 & 0.0 & 0.0 \\
        -1.0 & 0.0 & 0.0 \\
        0.0 & 1.0 & 0.0 \\
        0.0 & -1.0 & 0.0 \\
        0.0 & 0.0 & 1.0 \\
        0.0 & 0.0 & -1.0
    \end{bmatrix}
    , \hspace{5mm} b(\theta) = 
    \begin{bmatrix}
        -6.0 \\ -6.0 \\ -6.0 \\ -6.0 \\ -4.0 \\ -4.0 
    \end{bmatrix}.
\end{align}
The start and goal positions of each agent $i \in [2]$ are constrained via known equality constraints of the form \eqref{Eqn: App, Start and Goal Constraints} to be at $\bar p_o^i$ and $\bar p_d^i$, respectively, as given below:
\begin{alignat}{2} \nonumber
     &\bar p_o^1 = (0.0, 0.0, 0.0), \hspace{1cm} &&\bar p_d^1 = (10.0, 10.0, 10.0), \\ \nonumber
     &\bar p_o^2 = (10.0, 10.0, 10.0), &&\bar p_d^2 = (0.0, 0.0, 0.0), \\ \nonumber
     &\bar p_o^3 = (0.0, 10.0, 0.0), &&\bar p_d^3 = (10.0, 0.0, 10.0), \\ \nonumber
     &\bar p_o^4 = (10.0, 0.0, 10.0), &&\bar p_d^4 = (0.0, 10.0, 0.0).
\end{alignat}
We successfully recovered the \textit{a priori} unknown constraint parameters with a Gurobi solve time of 1.67 seconds.
The learned constraints were then used to generate safe motion plans in simulation, in conjunction with the agent costs, dynamics, time discretization, and time horizon given above, as well as equality constraints of the form \eqref{Eqn: App, Start and Goal Constraints} to enforce the following start positions $\bar p_o^i$ and goal positions $\bar p_d^i$ for each agent $i$:
\begin{alignat}{2} \nonumber
    &\bar p_o^1 = (7.0, -7.0, -7.0), \hspace{5mm} &&\bar p_d^1 = (-7.0, 7.0, 7.0), \\ \nonumber
    &\bar p_o^2 = (0.0, 7.0, 7.0), &&\bar p_d^2 = (7.0, -7.0, -7.0).
\end{alignat}
The generated safe motion plans are plotted in Fig. \ref{fig:quad_box}c.

\paragraph{Comparison Against the Single-Agent Constraint Learning Approach in \cite{Chou2020LearningConstraintsFromLocallyOptimalDemonstrationsUnderCostFunctionUncertainty}}

As shown in the 3-quadcopter interaction demonstrations illustrated in Fig. \ref{fig:quad_sphere} (Sec. \ref{sec: Experiments}), our game-theoretic approach correctly deduces that Agent 2’s constraint avoidance radius is 8, and that the larger gap of 9 between the trajectories of Agents 2 and 3 should be attributed to Agent 3 instead. As a baseline comparison, a single-agent constraint inference method applied to the same problem failed to accurately recover the constraint radius of Agent 2 with zero stationarity error (specifically, with stationarity error equal to 1.8).
The reason for this failure is that the single-agent constraint learning method for recovering Agent 2's constraints treats the other two agents as moving obstacles without intents and constraints of their own.
As a result,
assigning Agent 2 a constraint radius of 9 would lead to the conclusion that Agent 2 behaved suboptimally in Agent 1's vicinity, while assigning Agent 2 a constraint radius of 8 would lead to the conclusion that Agent 2 violated their constraint in Agent 3's vicinity.

\begin{figure}[ht]
    \centering
    \includegraphics[width=0.9\linewidth]{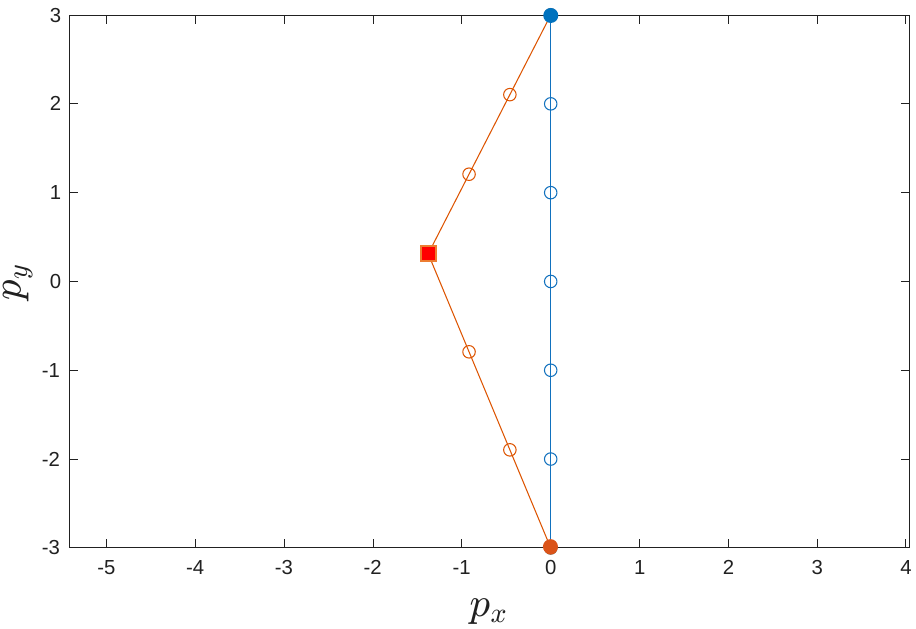}
    \caption{A demonstration of two interacting dynamic agents with spherical collision-avoidance constraints of different radii: 1 for Agent 1 (blue) and $\sqrt{2}$ for Agent 2 (red).
    Constraint learning via the centralized, single-agent approach characterized by \eqref{Eqn: Single-Agent Baseline, KKT} produced a stationarity error of 2.5.
    }
    \label{fig:nash_cexp}
\end{figure}

\begin{remark}
Alternatively, one can consider a single-agent inverse optimal control framework, in which one sums across all agent costs to define a system-level cost $J(\xi) := \sum_{i=1}^N J^i(\xi)$, and applies the single-agent constraint inference method presented in \cite{Chou2020LearningConstraintsFromLocallyOptimalDemonstrationsUnderCostFunctionUncertainty} to attempt to learn all agents' constraints at once. Concretely, consider the following feasibility problem of searching for valid parameter values $\theta$ that are consistent with the following KKT conditions, across all agent indices $i \in [N]$. The following equations can essentially be regarded as a centralized, single-agent version of the KKT conditions for our multi-agent constraints, as presented in \eqref{Eqn: KKT, Forward Game}:
\begin{subequations} \label{Eqn: Single-Agent Baseline, KKT}
\begin{align} 
\label{Eqn: Single-Agent Baseline, KKT, Primal Feasibility}
    &\boldh^i(\xi) = 0, \hspace{5mm} \boldg_k^i(\xi) \leq 0, \hspace{5mm} \boldg_\urk^i(\xi, \theta) \leq 0, \\ 
    \label{Eqn: Single-Agent Baseline, KKT, Lagrange Multiplier non-negativity}
    &\boldlambda_{d, k}^i, \boldlambda_{d, \urk}^i \geq 0, \\ 
    \label{Eqn: Single-Agent Baseline, KKT, Forward, Complementary Slackness}
    &\boldlambda_{d, k}^i \odot \boldg_k^i(\xi) = 0, \hspace{5mm} \boldlambda_{d, \urk}^i \odot \boldg_\urk^i(\xi, \theta) = 0, \\ 
    \label{Eqn: Single-Agent Baseline, KKT, Stationarity}
    &\nabla_\xi J(\xi) + (\boldlambda_{d, k}^i)^\top \nabla_\xi \boldg_k^i(\xi) \\ \nonumber
    & \hspace{2mm} + (\boldlambda_{d, \urk}^i)^\top \nabla_\xi \boldg_\urk^i(\xi, \theta) 
    + (\boldnu_d^i)^\top \nabla_\xi \boldh^i(\xi) = 0.
\end{align}
\end{subequations}
However, we empirically verified that for a 2-agent interaction setting, the inverse optimization problem characterized by \eqref{Eqn: Single-Agent Baseline, KKT} produced a stationarity error value of 2.5, thus likewise incorrectly concluding that the supplied demonstrations were not at Nash stationarity. These results were obtained using single-integrator dynamics, with $T = 7$, and with the following start and goal positions for each agent:
\begin{alignat}{2} \nonumber
    &\bar p_o^1 = (0.0, 3.0), \hspace{5mm} &&\bar p_d^1 = (0.0, -3.0), \\ \nonumber
    &\bar p_o^2 = (0.0, -3.0), &&\bar p_d^2 = (0.0, 3.0).
\end{alignat}
Agent 1's constraint was defined to ensure that $\Vert x_t^1 - x_t^2\Vert_2 \ge 1$ for all $t \in [T]$, while Agent 2's constraint was defined to ensure that $\Vert x_t^1 - x_t^2\Vert_2 \ge \sqrt{2}$ for all $t \in [T]$. The demonstrations are plotted in Fig. \ref{fig:nash_cexp}.
\end{remark}

\subsubsection{Nonlinear Constraint Recovery}
\label{subsubsec: App, Single Integrator Simulations for Nonlinear Constraint Recovery}

We also evaluate our method on simulated interactions between agents with single-integrator dynamics subject to \textit{a priori} unknown constraints that are nonlinear in the state and control variables (but affine in the unknown constraint parameter $\theta$). We also report the Gurobi solve times, for learning spherical collision-avoidance constraints, as a function of the number of interacting agents. Below, we provide the ground truth parameters and other details of the aforementioned experiments that were omitted in the main text.

\paragraph{Nonlinear Constraint Recovery under Cost Certainty}

\begin{figure}[ht]
    \centering
    \includegraphics[width=0.9\linewidth]{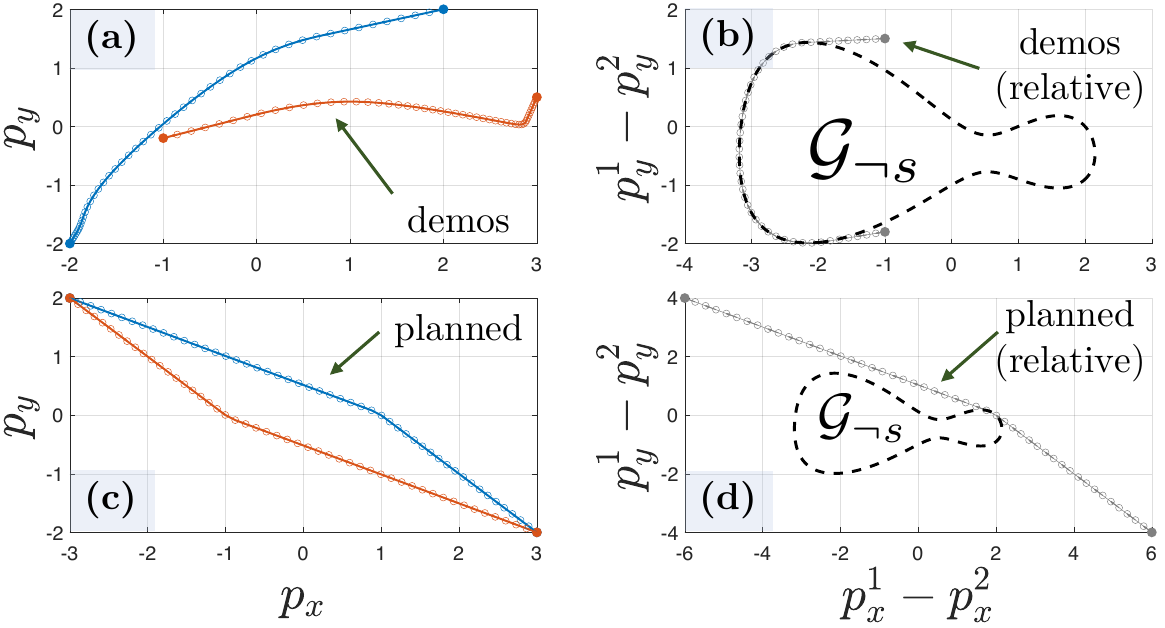}
    \caption{
    Constraint learning and safe planning for Agents 1 (blue) and 2 (red) with single integrator dynamics satisfying nonlinear constraints characterized by \textit{a priori} unknown parameters. (a) A demonstration of Agents 1 and 2, in relative coordinates, operating in a shared environment while satisfying their nonlinear constraints. (b) Our method exactly recovers the \textit{a priori} unknown constraint parameters of each agent. We have overlaid the true unsafe set $\Avoid(\theta^\star)$ which coincides with the guaranteed learned unsafe set $G_\urs(\D)$. (c) Using our learned constraints, we generate safe motion plans via volume extraction over the trajectory space. (d) We plot our generated safe motion plans overlaid with $\Avoid(\theta^\star)$ and $G_\urs(\D)$.
    }
    \label{fig:nonlinear}
\end{figure}

We generate, in simulation, an interaction demonstration of two agents whose motion satisfies single integrator dynamics discretized at intervals of $\Delta t = 0.5$, as well as the following parameterized constraints:
\begin{align} \label{Eqn: App, Nonlinear Constraints, def}
    \bigwedge_{t=1}^T \bigwedge_{j \in [N] \backslash \{i\}} &\Bigg\{  2(p_{x,t}^j - p_{x,t}^i)^4 + 2(p_{y,t}^j - p_{y,t}^i)^4 \\ \nonumber
    &\hspace{6mm} - 5 (p_{x,t}^j - p_{x,t}^i)^3 - 5 (p_{y,t}^j - p_{y,t}^i)^3 \\ \nonumber
    &\hspace{6mm} + 5(p_{x,t}^j - p_{x,t}^i -1)^3 \\ \nonumber
    &\hspace{6mm} + 5(p_{y,t}^j - p_{y,t}^i + 1)^3 - \theta_8^i \leq 0 \Bigg\}.
\end{align}
We set the following ground truth parameter values, which are \textit{a priori} unknown to the constraint learner, as follows:
\begin{align}
    \theta_8^i &= 2, \hspace{5mm} \forall \ i \in \{1, 2\}.
\end{align}
To generate the interaction demonstration, we set the time horizon to be $T = 50$ and define the cost function $J^i$ as follows: 
\begin{align} \label{Eqn: App, Nonlinear Constraints, Known Cost def}
    J^i = \sum_{t=1}^T \sum_{j=1}^2 \Big( \Vert p_{t+1}^j - p_t^j \Vert_2^2 + \Vert u_t^j \Vert_2^2 \Big).
\end{align}
The start and goal positions of each agent $i \in [2]$ are constrained via known equality constraints of the form \eqref{Eqn: App, Start and Goal Constraints} to be at $\bar p_o^i$ and $\bar p_d^i$, respectively, as given below:
\begin{alignat}{2} \nonumber
     &\bar p_o^1 = (-2.0, -2.0), \hspace{5mm} &&\bar p_d^1 = (-1.0, -0.2), \\ \nonumber
     &\bar p_o^2 = (2.0, 2.0), &&\bar p_d^2 = (3.0, 0.5).
\end{alignat}
We successfully recovered the \textit{a priori} unknown constraint parameters with a Gurobi solve time of 0.02 seconds. The generated demonstration interactions are provided in Fig. \ref{fig:nonlinear}a and \ref{fig:nonlinear}b.
The learned constraints were then used to generate safe motion plans in simulation, in conjunction with the agent costs, dynamics, time discretization, and time horizon given above, as well as equality constraints of the form \eqref{Eqn: App, Start and Goal Constraints} to enforce the following start positions $\bar p_o^i$ and goal positions $\bar p_d^i$ for each agent $i$:
\begin{align} \nonumber
    &\bar p_o^1 = (3.0, -2.0), \hspace{5mm} \bar p_d^1 = (-3.0, 2.0), \\ \nonumber
    &\bar p_o^2 = (-3.0, 2.0), \hspace{5mm} \bar p_d^2 = (3.0, -2.0).
\end{align}
The generated safe motion plans are plotted in Figs. \ref{fig:nonlinear}c and \ref{fig:nonlinear}d.

\paragraph{Nonlinear Constraint Recovery under Cost \textit{Uncertainty}}

We repeat the above nonlinear constraint learning and motion planning simulations, with all parameters (including start and goal positions) held fixed, for the setting where the cost objectives of the agents, denoted $\bar J^1$ and $\bar J^2$ below, are also characterized by parameters $\bar \theta_1^1$ and $\bar \theta^2$ whose ground truth values are \textit{a priori} unknown by the constraint learners. Specifically, we assume here that the agents' cost functions are given by:
\begin{align} \label{Eqn: App, Nonlinear Constraints, Unknown Cost def}
    \bar J^i(\bar \theta^i) &= \sum_{t=1}^T \Big( \Vert p_{t+1}^1 - p_t^1 \Vert_2^2 + \bar \theta^i \Vert p_{t+1}^2 - p_t^2 \Vert_2^2 \Big) \\ \nonumber
    &\hspace{5mm} + \sum_{t=1}^T \sum_{j=1}^2 \Vert u_t^j \Vert_2^2, \hspace{5mm} \forall \ i \in \{1, 2\},
\end{align}
with ground truth parameter values $\bar \theta^i = 0.73$ for each $i \in \{1, 2\}$. We successfully recovered the \textit{a priori} unknown constraint parameters with a Gurobi solve time of 0.03 seconds. The generated demonstration interactions are provided in Fig. \ref{fig:nonlinear}a and \ref{fig:nonlinear}b.
The learned constraints were then used to generate safe motion plans in simulation, in conjunction with the agent costs, dynamics, time discretization, and time horizon given above, as well as equality constraints of the form \eqref{Eqn: App, Start and Goal Constraints} to enforce the following start positions $\bar p_o^i$ and goal positions $\bar p_d^i$ for each agent $i$:
\begin{align} \nonumber
    &\bar p_o^1 = (3.0, -2.0), \hspace{5mm} \bar p_d^1 = (-3.0, 2.0), \\ \nonumber
    &\bar p_o^2 = (-3.0, 2.0), \hspace{5mm} \bar p_d^2 = (3.0, -2.0).
\end{align}
The generated safe motion plans are plotted in Figs. \ref{fig:nonlinear}c and \ref{fig:nonlinear}d.


\subsubsection{Hardware Experiments}
\label{subsubsec: App, Hardware Experiments}

As described in Sec. \ref{subsec: Hardware Experiments}, we evaluate our method on hardware experiments between ground robots with unicycle dynamics and either spherical or polytopic collision-avoidance constraints, as described by \eqref{Eqn: App, Spherical Collision-Avoidance Constraints, def} and \eqref{Eqn: App, Polytopic Collision-Avoidance Constraints, def}. Below, we provide the ground truth parameters and other details of the aforementioned experiments that were omitted in the main text.

\begin{figure*}[ht]
    \centering
    \includegraphics[width=\linewidth]{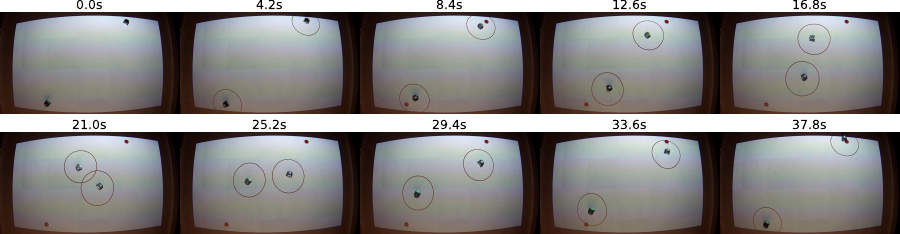}
    \caption{
    A time-lapse montage of an interaction demonstration between two ground robots in our hardware setup. Here, both robots follow unicycle dynamics and obey spherical collision-avoidance constraints.
    }
    \label{fig:hardware_demo}
\end{figure*}

\paragraph{Spherical Collision-Avoidance Constraints}

We generate three interaction demonstrations of two ground robots, which we then use to learn their spherical collision-avoidance constraint sets (Fig. \ref{fig:hardware}a). 
The three interaction demonstrations share all of the following parameter settings presented below, including the two robots' initial and final goal positions.
Specifically, we discretize the continuous-time unicycle dynamics \eqref{Eqn: App, Unicycle Dynamics, original} at intervals of $\Delta t = 1$, and set the time horizon to be $T = 33$. We define the cost of each agent to encode the smoothness objective formulated in Sec. \ref{subsec: Constraint Parameterizations and Agent Costs}, i.e., 
\begin{align} \nonumber
    J^i &= \sum_{t=1}^{T-1} \Vert p_{t+1}^i - p_t^i \Vert_2^2, \hspace{1cm} \forall \ i \in \{1, 2\}.
\end{align}
Spherical collision-avoidance constraints of the form \eqref{Eqn: App, Spherical Collision-Avoidance Constraints, def}, with the following \textit{a priori} unknown ground truth parameters, are enforced on the agents' trajectories:
\begin{align} \nonumber
    \theta_1^1 = 0.55, \hspace{5mm} \theta_1^2 = 0.55.
\end{align}
The start and goal positions of each agent $i \in [2]$ are constrained via known equality constraints of the form \eqref{Eqn: App, Start and Goal Constraints} to be at $\bar p_o^i$ and $\bar p_d^i$, respectively, as given below:
\begin{align} \nonumber
     &\bar p_o^1 = (0.8, -0.8), \hspace{5mm} \bar p_d^1 = (-0.8, 0.8), \\ \nonumber
     &\bar p_o^2 = (-0.8, 0.8), \hspace{5mm} \bar p_d^2 = (0.8, -0.8).
\end{align}
We successfully recovered the \textit{a priori} unknown constraint parameters with a Gurobi solve time of 0.03 seconds.
The learned constraints were then used to generate two safe motion plans on our hardware setup, in conjunction with the agent costs, dynamics, time discretization, and time horizon given above, but different start positions $\bar p_o^i$ and goal positions $\bar p_d^i$ for each agent $i$, as enforced by equality constraints of the form \eqref{Eqn: App, Start and Goal Constraints}. In particular, for the first motion plan, we set:
\begin{alignat}{2} \nonumber
    &\bar p_o^1 = (-0.7, -0.7), \hspace{5mm} &&\bar p_d^1 = (0.7, 0.7), \\ \nonumber
    &\bar p_o^2 = (-0.7, 0.7), &&\bar p_d^2 = (0.7, -0.7).
\end{alignat}
For the second motion plan, we set:
\begin{align} \nonumber
    &\bar p_o^1 = (0.7, -0.7), \hspace{5mm} \bar p_d^1 = (-0.7, 0.7), \\ \nonumber
    &\bar p_o^2 = (-0.7, 0.7), \hspace{5mm} \bar p_d^2 = (0.7, -0.7).
\end{align}
Both motion plans are plotted in Fig. \ref{fig:hardware}b.

\paragraph{Box-Shaped Collision-Avoidance Constraints}

We generate two interaction demonstrations of two ground robots, which we then use to learn their polytopic collision-avoidance constraint sets (Fig. \ref{fig:hardware}c). 
The two interaction demonstrations share all of the following parameter settings presented below, including the two robots' initial and final goal positions.
Specifically, we discretize the continuous-time quadcopter dynamics at intervals of $\Delta t = 1$, and set the time horizon to be $T = 33$. We define the cost of each agent to encode the smoothness objective formulated in Sec. \ref{subsec: Constraint Parameterizations and Agent Costs}, i.e.,
\begin{align} \nonumber
    J^i &= \sum_{t=1}^{T-1} \Vert p_{t+1}^i - p_t^i \Vert_2^2, \hspace{1cm} \forall \ i \in \{1, 2\}.
\end{align}
Polytopic collision-avoidance constraints of the form \eqref{Eqn: App, Polytopic Collision-Avoidance Constraints, def} with the \textit{a priori} unknown ground truth parameters $\{a_\beta(\theta) \in \R^2, b_\beta(\theta) \in \R : \beta \in [4] \}$ defined below, are enforced on each agent's trajectory. For each $\beta \in [6]$, we set $a_\beta(\theta) \in \R^2$ to be the transpose of the $\beta$-th row of the matrix $A(\theta)$ defined below, and we set $b_\beta(\theta) \in \R$ to be the $\beta$-th entry of the vector $b(\theta)$ defined below:
\begin{align} \nonumber
    A(\theta) &= 
    \begin{bmatrix}
        1.0 & 0.0 \\
        -1.0 & 0.0 \\
        0.0 & 1.0 \\
        0.0 & -1.0
    \end{bmatrix}
    , \hspace{5mm} b(\theta) = 
    \begin{bmatrix}
        -0.4 \\ -0.4 \\ -0.2 \\ -0.2
    \end{bmatrix}.
\end{align}
The start and goal positions of each agent $i \in [2]$ are constrained via known equality constraints of the form \eqref{Eqn: App, Start and Goal Constraints} to be at $\bar p_o^i$ and $\bar p_d^i$, respectively, as given below:
\begin{align} \nonumber
     &\bar p_o^1 = (0.8, -0.8), \hspace{5mm} \bar p_d^1 = (-0.8, 0.8), \\ \nonumber
     &\bar p_o^2 = (-0.8, 0.8), \hspace{5mm} \bar p_d^2 = (0.8, -0.8).
\end{align}
We successfully recovered the \textit{a priori} unknown constraint parameters with a Gurobi solve time of 0.23 seconds.
The learned constraints were then used to generate two safe motion plans on our hardware setup, in conjunction with the agent costs, dynamics, time discretization, and time horizon given above, but different start positions $\bar p_o^i$ and goal positions $\bar p_d^i$ for each agent $i$, as enforced by equality constraints of the form \eqref{Eqn: App, Start and Goal Constraints}. In particular, for the first motion plan, we set:
\begin{align} \nonumber
    &\bar p_o^1 = (-0.7, -0.7), \hspace{5mm} \bar p_d^1 = (0.7, 0.7), \\ \nonumber
    &\bar p_o^2 = (-0.7, 0.7), \hspace{5mm} \bar p_d^2 = (0.7, -0.7).
\end{align}
For the second motion plan, we set:
\begin{align} \nonumber
    &\bar p_o^1 = (0.7, -0.7), \hspace{5mm} \bar p_d^1 = (-0.7, 0.7), \\ \nonumber
    &\bar p_o^2 = (-0.7, 0.7), \hspace{5mm} \bar p_d^2 = (0.7, -0.7).
\end{align}
Both motion plans are plotted in Fig. \ref{fig:hardware}d.

\subsubsection{Comparison Against the Cost Inference Baseline}
\label{subsubsec: App, Comparison Against Cost Inference Baseline}

As described in Sec. \ref{subsec: Comparison Against Cost Inference Baseline}, we compared our constraint-learning based approach for safe motion planning against the cost inference-based baseline approach in \cite{Peters2021InferringObjectives}. 
Concretely, we first generated an interaction demonstration of two agents with double integrator dynamics discretized at intervals of $\Delta t = 1$, who interact over a time horizon of $T = 10$. When generating the interaction demonstration, we define the cost of each agent to encode the smoothness objective formulated in Sec. \ref{subsec: Constraint Parameterizations and Agent Costs}, i.e.,
\begin{align} \nonumber
    J^i &= \sum_{t=1}^{T-1} \Vert p_{t+1}^i - p_t^i \Vert_2^2, \hspace{1cm} \forall \ i \in \{1, 2\}.
\end{align}
Spherical collision-avoidance constraints of the form \eqref{Eqn: App, Spherical Collision-Avoidance Constraints, def} with \textit{a priori} unknown ground truth parameters $\theta_1^1$ and $\theta_1^2$ as given below, are enforced on each agent's demonstration trajectory:
\begin{align} \label{Eqn: App, Baseline, true parameter}
    \theta_1^i = 49, \hspace{5mm} \forall \ i \in \{1, 2\}.
\end{align}
The start and goal positions of each agent $i \in [2]$ are constrained via known equality constraints of the form \eqref{Eqn: App, Start and Goal Constraints} to be at $\bar p_o^i$ and $\bar p_d^i$, respectively, as given below:
\begin{alignat}{2} \nonumber
     &\bar p_o^1 = (0.0, 0.0), \hspace{6mm} &&\bar p_d^1 = (10.0, 10.0), \\ \nonumber
     &\bar p_o^2 = (10.0, 10.0), &&\bar p_d^2 = (0.0, 0.0).
\end{alignat}
Our constraint learning method successfully recovered the \textit{a priori} unknown constraint parameters with a Gurobi solve time of 0.04 seconds. Meanwhile, the baseline method extracts each agents' collision-avoidance intent by attempting to learn the parameter $\tilde \theta$ in the following cost function (first presented in Sec. \ref{subsec: Comparison Against Cost Inference Baseline}), which encodes collision-avoidance as a log-barrier function. Note that the baseline method does not assume that hard collision-avoidance constraints are enforced on either agent's trajectory:
\begin{align} \label{Eqn: App, Baseline Cost, tilde J}
    \tilde J^i(\tilde \theta) &:= J^i - \tilde \theta \cdot \sum_{t=1}^T \log(\Vert p_t^i - p_t^{2-i} \Vert_2^2) \\ \nonumber
    &= \sum_{t=1}^{T-1} \Vert p_{t+1}^i - p_t^i \Vert_2^2 - \tilde \theta \cdot \sum_{t=1}^T \log(\Vert p_t^i - p_t^{2-i} \Vert_2^2).
\end{align}
Using the baseline method, we recover the parameter $\tilde \theta = 2.51$.

We compare the safety guarantees of motion plans generated using our learned constraints against motion plans generated using the baseline method. Specifically, we first generate motion plans via our method, using (1) the smoothness cost \eqref{Eqn: App, Smoothness Cost, def}, (2) the spherical collision-avoidance constraints \eqref{Eqn: App, Spherical Collision-Avoidance Constraints, def} with the recovered constraint parameters $\theta_1^1 = \theta_1^2 = 49$, i.e., \eqref{Eqn: App, Baseline, true parameter}, and (3) equality constraints of the form \eqref{Eqn: App, Start and Goal Constraints} to encode the start positions $\bar p_o^i$ and goal positions $\bar p_d^i$ of each agent $i \in \{1, 2\}$:\footnote{Note that Fig. \ref{fig:baseline}a shows the forward motion plans generated in \textit{relative} coordinates, whereas the start and goal positions reported here are in \textit{absolute} coordinates.}
\begin{alignat}{2} \nonumber
    &\bar p_o^1 = (0.0, 5.0), \hspace{6mm} &&\bar p_d^1 = (10.0, 5.0), \\ \nonumber
    &\bar p_o^2 = (10.0, 5.0), &&\bar p_d^2 = (0.0, 5.0).
\end{alignat}
Next, we generate motion plans via the baseline approach, using the smoothness plus log barrier cost \eqref{Eqn: App, Baseline Cost, tilde J} with inferred parameter $\tilde \theta = 2.51$, \textit{without any hard collision-avoidance constraints}. As noted in Sec. \ref{subsec: Comparison Against Cost Inference Baseline} and Fig. \ref{fig:baseline}, the motion plan generated using our method (and our learned constraints) satisfies the ground truth spherical collision-avoidance constraint, while the motion plan generated using the baseline approach violates the ground truth spherical collision-avoidance constraint.

\end{document}